\documentclass[journal]{IEEEtran}

\usepackage{mathpazo,amsmath,amssymb,epsfig,enumerate,bbm,calc,ifthen,capt-of} 
\usepackage{algorithm}
\usepackage{relsize}

\usepackage[noend]{algorithmic}
\usepackage[center]{subfigure}
\usepackage{color,graphicx}
\usepackage[margin=25mm,textheight=247mm,textwidth=145mm]{geometry}
\usepackage{multicol}
\usepackage{multirow}
\usepackage{braket}
\usepackage{array}
\usepackage{tabu}
\usepackage{makecell}
\usepackage{booktabs}
\usepackage{textcomp}
\usepackage{bm}
\newsavebox{\mytable}
\usepackage{tikz}
\usetikzlibrary{calc}

\usepackage{tabularx}
\usepackage[utf8]{inputenc} 
\usepackage[T1]{fontenc}    
\usepackage{hyperref}       
\usepackage{url}            
\usepackage{booktabs}       
\usepackage{amsfonts}       
\usepackage{nicefrac}       
\usepackage{microtype}      
\usepackage{xcolor}         
\usepackage{comment}
\usepackage{titletoc}
\usepackage{mathtools} 
\usepackage{amsthm}
\usepackage[scaled]{helvet} 
\usepackage{courier} 
\normalfont
\usepackage[T1]{fontenc}

\newcommand\subsubsubsection[1]{\hspace{4mm}\textit{#1}}

\newcommand{\tab}[1]{\hspace{3mm}}

\DeclareMathOperator*{\argmin}{arg\,min}

\theoremstyle{plain}
\newtheorem{theorem}{Theorem}[section]

\theoremstyle{definition}
\newtheorem{definition}[theorem]{Definition}

\theoremstyle{remark}

\begin{document}
%
\title{A Survey on Heterogeneous \\ Federated Learning}

%

\author{Dashan~Gao,
        Xin~Yao,~\IEEEmembership{Fellow,~IEEE}
        and~Qiang~Yang,~\IEEEmembership{Fellow,~IEEE}
\thanks{Dashan Gao is with the Department
of Computer Science and Engineering, HKUST, Hong Kong,
China and the Department
of Computer Science and Engineering, Southern University of Science and Technology, Shenzhen, China, e-mail: dgaoaa@connect.ust.hk}
\thanks{Xin Yao is with the Department
of Computer Science and Engineering, Southern University of Science and Technology, Shenzhen, China, e-mail: xiny@sustech.edu.cn}
\thanks{Qiang Yang is with the Department
of Computer Science and Engineering, HKUST, Hong Kong, China, e-mail: qyang@cse.ust.hk}
}

%
%

\markboth{Journal of \LaTeX\ Class Files, September~2022}%
{Shell \MakeLowercase{\textit{et al.}}: Bare Demo of IEEEtran.cls for IEEE Journals}
%



\maketitle

\begin{abstract}

Federated learning (FL) has been proposed to protect data privacy and virtually assemble the isolated data silos by cooperatively training models among organizations without breaching privacy and security. 
However, FL faces heterogeneity from various aspects, including data space, statistical, and system heterogeneity. 
For example, collaborative organizations without conflict of interest often come from different areas and have heterogeneous data from different feature spaces. Participants may also want to train heterogeneous personalized local models due to non-IID and imbalanced data distribution and various resource-constrained devices. Therefore, heterogeneous FL is proposed to address the problem of heterogeneity in FL.
In this survey, we comprehensively investigate the domain of heterogeneous FL in terms of data space, statistical, system, and model heterogeneity. 
We first give an overview of FL, including its definition and categorization. 
Then, We propose a precise taxonomy of heterogeneous FL settings for each type of heterogeneity according to the problem setting and learning objective. 
We also investigate the transfer learning methodologies to tackle the heterogeneity in FL. 
We further present the applications of heterogeneous FL. 
Finally, we highlight the challenges and opportunities and envision promising future research directions toward new framework design and trustworthy approaches.

\end{abstract}

\begin{IEEEkeywords}
Heterogeneous Federated Learning,
Vertical Federated Learning
\end{IEEEkeywords}

\ifCLASSOPTIONcompsoc
\IEEEraisesectionheading{\section{Introduction}\label{sec-introduction}}
\else
\section{Introduction}
\label{sec-introduction}
\fi

%

\IEEEPARstart{I}{n}
the last decade, machine learning (ML) has achieved tremendous success and is widely applied to numerous areas, thanks to the surge of big data. 
Despite the great success of ML, the broad application of ML is heavily restricted by the lack of large-scale datasets. 
Moreover, the data collection and labeling can be expensive, making it infeasible for a single organization to collect enough data.
In other cases, although companies may already have sufficient data regarding the number of recordings. 
They may want to enrich the feature space of their data by incorporating the auxiliary features distributed among other institutions. 
However, sharing data between institutions or aggregating data from user devices leads to the risk of privacy leakage.

In recent years, concerns about data privacy protection have been emerging due to continuous privacy violations. 
In response, information privacy policies and laws such as GDPR~\cite{GDPRhtmlFull}, and CCPA~\cite{ghosh2018you} stipulate on sharing of data between companies to protect personal data from being abused.
Moreover, data also has enormous commercial values for companies. It is infeasible to transmit data out of privacy, security, and business concerns. 
Therefore, it is challenging to efficiently leverage data distributed in multiple institutions for ML.

Federated learning (FL) was proposed to train ML models based on data distributed among clients~\cite{McMahan2016ModelAvg,HBrendanMcMahan2016}. In FL, intermediate results such as gradients are exchanged and aggregated to update models without raw data transmission. 
Many efforts have been devoted to designing federated learning algorithms from various aspects, including privacy preservation, robustness, efficiency, security, scalability, and performance~\cite{Kamp2018MA,Pillutla2019Robust,HBrendanMcMahan2016,JeongDistillation2018,FelixSattler2019NonIID}.

However, FL faces a problem of heterogeneity. Heterogeneity in FL comes from various aspects, including imbalanced data distribution, heterogeneous feature spaces, unstable network connectivity, and limited device resources. 
We categorize the heterogeneity in FL into three types: \textit{data space, statistical, and system heterogeneity}. 
All types of heterogeneity have attracted tremendous research interest in recent years. 
To tackle the heterogeneity as mentioned above, various approaches that leverage \textit{model heterogeneity} are also widely explored, as heterogeneous models can help train tailored model architectures according to the feature space and device resources of parties. 
Some surveys have focused on statistical heterogeneous FL~\cite{tan2022personalizedflsurvey,TianLi2019Survey} and system heterogeneous FL~\cite{chen2021towards,xu2021asynchronous}. Some works investigate the system design of FL~\cite{BingshengHe2019Survey}
However, there is still a lack of comprehensive survey on all the above-mentioned heterogeneity in FL.

To tackle statistical heterogeneity, some works train a robust global model against the non-IID and unbalanced distributions in different participants~\cite{YueZhao2018NonIID,LierIID,li2019convergence,JeongDistillation2018}. Another approach is to train personalized local models for all parties by formulating the FL as multi-task learning problems~\cite{Ghosh2020AnEF,jiang2019improving,wang2019federated}.
Similarly, to tackle system heterogeneity, some approaches propose to efficiently build a single global model via client selection~\cite{nishio2018client,yoshida2019hybrid,chen2021towards}, asynchronous aggregation~\cite{sprague2018asynchronous,Wu2021safa}, and model compression~\cite{yao2021fedhm,Caldas2018exband}. When different resource-constraint devices want to deploy a personalized local model, it is possible to train heterogeneous local models by model splitting and clustered training~\cite{hong2022efficient,munir2021fedprune,diao2021heterofl}.

Although most studies focus on statistical and system heterogeneity, the data spaces of different participants may differ as well.
In many real-world applications, the collaborating participants are organizations (e.g., banks or hospitals) from different industries without conflict of interest. Therefore, the collected data of each participant are from heterogeneous feature spaces. 
For example, in advertisement scenarios, an advertiser holds users' browsing history and clicking records, while a businesses hold the purchase history. 
To leverage data from both feature spaces, the two parties must conduct data space heterogeneous FL to build a model on the joint feature space. 
Such heterogeneity in data space brings new challenges to FL in various aspects. 

According to data space distribution, data space heterogeneous FL can be categorized into \textit{vertical federated learning} (VFL) and \textit{heterogeneous federated transfer learning} (Hetero-FTL). 
When data are partitioned by features, VFL aims to train a model on the aligned instances.
However, participants may share limited or no aligned instances and some common features in many cases. 
Such overlap on feature space or instance space provides opportunities to transfer knowledge among parties. 
Fortunately, federated transfer learning is proposed to adopt transfer learning to share knowledge among the participants~\cite{QiangYang2019}. According to the shared data space among participants, we categorize Hetero-FTL into three cases: feature-sharing, instance-sharing, and label-sharing. 
Compared to VFL, Hetero-FTL relaxes the requirement on instance alignment and improves the model availability. 
We find that the data space heterogeneous FL, especially the Hetero-FTL is still under-explored and is a promising research direction that attracts increasing research interests.

We investigate the approaches to tackle different types of heterogeneity in FL. There are mainly five approaches to tackle heterogeneity: knowledge distillation, data augmentation, parameter sharing, domain adaptation, and matrix factorization. 
For each approach, we discuss and review how to use them to address each type of heterogeneity.

We also investigate the applications of heterogeneous FL in different areas, seeing it being actively promoted for industrial applications.
Furthermore, we conclude the major challenges and promising future directions in heterogeneous FL algorithm designing. 

Overall, our contributions are threefold: 
\begin{enumerate}
	\item We make the first survey on the existing works about data space, statistical, system, and model heterogeneity. We propose a taxonomy for each type of heterogeneity based on the problem setting and learning approach. 
	\item We identify five transfer learning strategies to address data space, statistical, and system heterogeneity in FL and review the literature from data-based, architecture-based, and model-based perspectives. 
	\item We envision promising future research directions toward new framework design and trustworthy approaches toward building heterogeneous FL systems. 
\end{enumerate}

\begin{figure*}[t!] 
\includegraphics[width=0.9\linewidth]{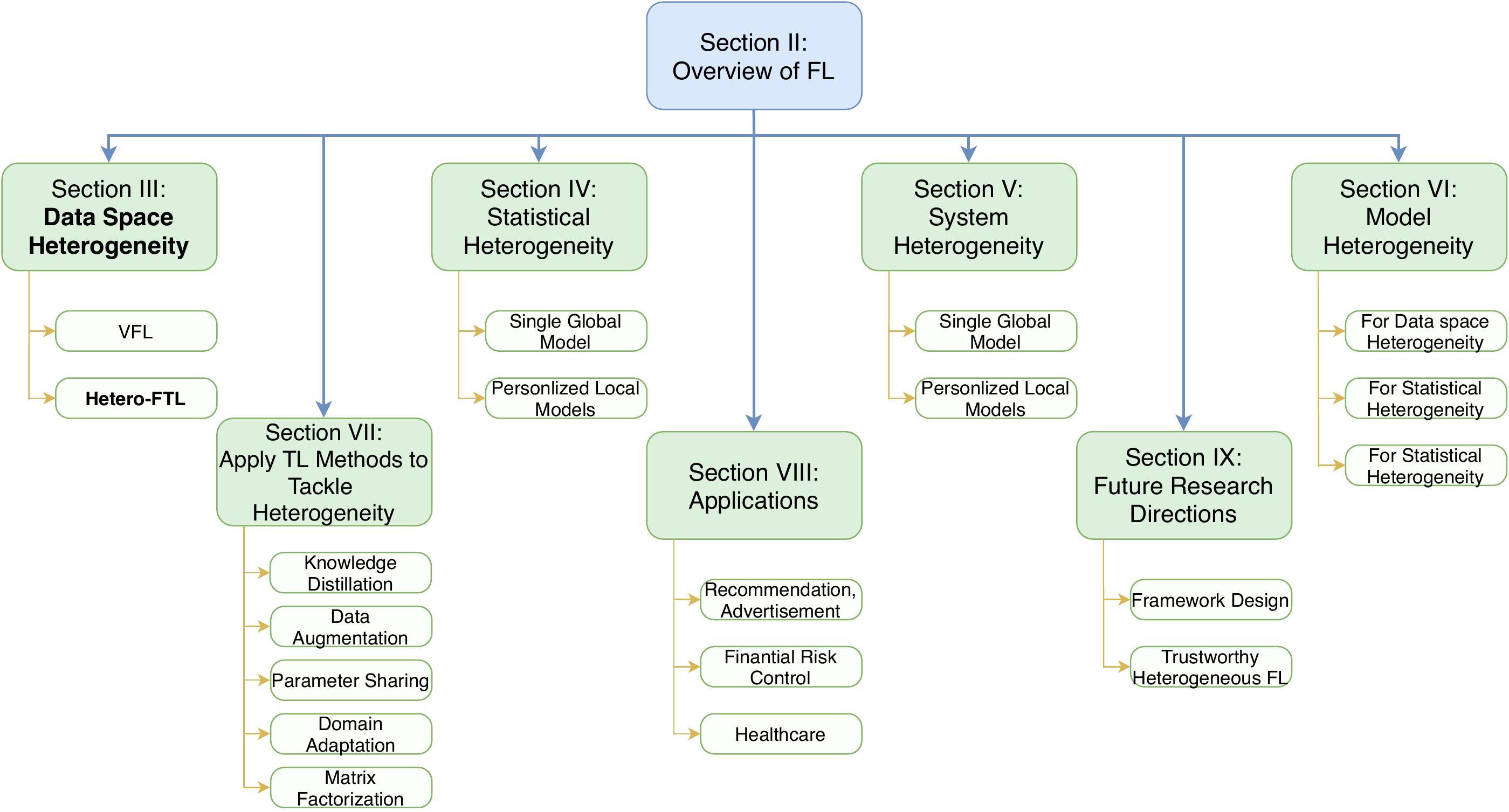}
\centering
\caption{ Organization of this survey.  }
\label{fig_survey_structure}
\end{figure*}

\textit{Organization.} The remainder of the survey is organized as shown in Figure~\ref{fig_survey_structure}. 
Section~\ref{sec-overview} gives an overview of FL and introduces the heterogeneity, trade-off, and categorization of FL. 
Section~\ref{sec-dataspace_FL} proposes the notion of data space heterogeneous FL and provide a categorization method according to the data distribution of the heterogeneous FL.
Section~\ref{sec-statistical_FL} introduces the statistical heterogeneity. 
Section~\ref{sec-system_FL} discusses the system heterogeneity. 
Section~\ref{sec-model_FL} introduces how to train heterogeneous models in FL and how model heterogeneity can help to tackle data space, statistical, and system heterogeneity. 
Section~\ref{sec-TL_HeteroFL} discusses how to use different transfer learning approaches to tackle different types of heterogeneity in FL.
Section~\ref{sec-applications} presents the applications of heterogeneous FL.
Section~\ref{sec-challenges} discusses the challenges in heterogeneous FL. 
Finally, Section~\ref{sec-conclusion} draws the conclusion. Table~\ref{table_acronyms} lists the key acronyms and abbreviations used throughout the paper.

\begin{table}[ht]
\centering
\begin{tabular}{c | c }
\toprule
Acronyms & Definitions \\
\midrule
FL & Federated Learning \\
HFL & Horizontal Federated Learning \\
VFL & Vertical Federated Learning \\
Hetero & Heterogeneous \\
Homo & Homogeneous \\
TL & Transfer Learning \\
FTL & Federated Transfer Learning \\
ML & Machine Learning \\
HE & Homomorphic Encryption \\
DP & Differential Privacy \\
MPC & Secure Multiparty Computation \\
\bottomrule
\end{tabular}
\vspace{2mm}
\caption{List of Key Acronyms}
\vspace{-2mm}
\label{table_acronyms}
\end{table}

\section{An Overview of Federated Learning}\label{sec-overview}

In this section, we introduce the basic idea of FL. We first give the definition of FL. 
We categorize the FL according to the data distribution among parties. From the view of data distribution, we follow the categorization from~\cite{QiangYang2019}.

\subsection{Definition of Federated Learning}
We provide a formal definition of FL with the reference of ~\cite{QiangYang2019,BingshengHe2019Survey,Kairouz2019Survey105}. We assume there are $N$ parties denoted as $\{P_1, P_2, ..., P_{N}\}$ with their private isolated datasets $\{\mathcal{D}_1, \mathcal{D}_2, ..., \mathcal{D}_N \}$, respectively. 
There can be three components in a complete dataset: the feature space $\mathcal{X}$, the label space $\mathcal{Y}$ and the ID space $\mathcal{I}$. The feature space $\mathcal{X}$ and ID space $\mathcal{I}$ of the parties may not be identical.

Different FL problems differ in the distribution of data among parties. 
We denote a party $P_i$ with a complete dataset $\mathcal{D}_i = \{X_i, Y_i, I_i\}$ as an active party, and a party $P_j$ with data $\mathcal{D}_j = \{X_j, I_j\}$ as a passive party. An active party can train a model locally to predict the label. In contrast, a passive party can only provide auxiliary features to other parties via ID alignment and cannot train a valid model on its own.
In the canonical FL problem, a.k.a. HFL, all parties are active parties and hold homogeneous labeled data where $\mathcal{X}_i = \mathcal{X}_j$ and $\mathcal{Y}_i = \mathcal{Y}_j$ for $\forall i, j \in [1, N]$.
In heterogeneous FL, different parties may hold heterogeneous data. 
For example, in vertical FL, $\mathcal{X}_i \cap \mathcal{X}_j = \emptyset$ for $\forall i, j \in [1, N]$, and $\mathcal{Y}_k \neq \emptyset, \mathcal{Y}_i = \emptyset$ for $\forall i \in [1, ..., k-1, k+1, ..., N]$. We demonstrate the detailed categorization of FL problems in Section~\ref{FLcategorization}.

For the local setting, a.k.a. distributed on-site learning\cite{rahman2020survey}, each active party $P_i$ uses its own local data $\mathcal{D}_i$ to train a machine learning model $\mathcal{M}_i$. The predictive accuracy of $\mathcal{M}_i$ is denoted as $\mathcal{V}_i$.
For the centralized setting, all parties aggregate and align their data together to get $\mathcal{D}_{agg} = \bigcup\limits_{i=1}^{N} \mathcal{D}_i$. A model $\mathcal{M}_{cen}$ is trained over the aggregated data $\mathcal{D}_{agg}$ in a centralized fashion. 
In contrast, FL allows the parties to train a model $\mathcal{M}_{fed}$ collaboratively on the whole dataset $\mathcal{D}_{agg}$, while each party $P_i$ does not disclose its raw data $\mathcal{D}_i$ to any other party. 
The predictive accuracy of $\mathcal{M}_{agg}$ and $\mathcal{M}_{fed}$ are denoted as $\mathcal{V}_{agg}$ and $\mathcal{V}_{fed}$, respectively. 
Then, an FL system should satisfy the following properties:
\begin{enumerate}
\item{For a valid FL system, for local models trained by active parties $P_{act}$, the performance of the federated model should be better than the performance of any local model, i.e. }
\begin{equation*}
\mathcal{V}_{fed} \geq \mathcal{V}_i , \; \forall i \in P_{act}.
\end{equation*}
\item{The difference between the performance of federated model $\mathcal{M}_{fed}$ and the centralized model $\mathcal{M}_{agg}$ should be bounded by $\delta > 0$, i.e.}
\begin{equation*}
 | \mathcal{V}_{agg} - \mathcal{V}_{fed} | \leq \delta,  \; \exists \delta > 0.
\end{equation*}
\end{enumerate}

\subsection{Heterogeneity in Federated Learning}
In FL systems, \textit{heterogeneity} is one of the most significant challenges and is studied from multiple aspects, including 1) data space heterogeneity, 2) statistical heterogeneity, and 3) system heterogeneity.

\begin{itemize}
    \item{\textit{Data Space Heterogeneity}}: The data space, including feature space $\mathcal{X}$, label space $\mathcal{Y}$ and ID space $\mathcal{I}$ can differ among the parties in FL due to data availability, domain difference, and device heterogeneity issues. 
    Parties with different feature spaces or label spaces can not train a shared model like traditional HFL. 
    Parties who share common instance IDs can leverage auxiliary features from other parties in training and inference processes. 
    In another case, parties without common instance IDs can explore to transfer knowledge from other feature spaces to boost the model performance in the training process. Heterogeneous FL techniques investigated in this survey aim to tackle data space heterogeneity via vertical and heterogeneous federated transfer learning.
    (Section~\ref{sec-dataspace_FL}).
    
    \item{\textit{Statistical Heterogeneity}}: Data collected from the same data space can be non-IID (independent and identically) distributed in the FL system. For example, autonomous vehicles run in environments varying from rural areas to urban areas. This leads to the marginal distributions $P(\mathcal{X})$ of collected data differ. Moreover, different drivers in the same environments show various driving style preferences, leading to varying conditional distributions $P(\mathcal{Y}|\mathcal{X})$ of human driving behaviors. 
    In addition, class imbalance and data size imbalance across parties may vary significantly as well.     (Section~\ref{sec-statistical_FL}).

    \item{\textit{System Heterogeneity}}: In FL systems, heterogeneity that happens at the system level can also exert a significant influence on the FL training process. Devices may differ in hardware (CPU, GPU, memory), network connectivity (WiFi, 4G, 5G), and power supply (battery mode), which leads to various computation, communication, and storage capabilities. 
    (Section~\ref{sec-system_FL}).
\end{itemize}

\subsection{Utility-Privacy-Efficiency Trade-off}
The design of FL algorithms faces three major objectives: \textit{model performance (utility)}, \textit{data privacy}, and \textit{system efficiency}. 
Improving one objective usually leads to the decline of other objectives. 
Multiple factors can affect such a trade-off. 
Here we identify six major factors in FL that affect the trade-off: model complexity (e.g., linear model, decision tree, neural networks), cryptographic privacy (e.g., HE, MPC), and obfuscation-based privacy (DP), sharing labels, asynchronization, and coordinator. We summarize the impact of each factor on the utility-privacy-efficiency trade-off in Table~\ref{trade_off}. 
Generally speaking, model performance can be improved by increasing model complexity. Privacy can be improved by cryptographic or obfuscation-based approaches. Efficiency can be improved via sharing labels, asynchronization, and introducing a coordinator. We elaborate on each factor as follows. 
\begin{table}[ht]
\centering
\begin{tabular}{c | c c c }
\toprule
Factor & Utility & Privacy & Efficiency \\
\midrule
model complexity & \textcolor{red}{$\bm{\uparrow}$} & $\downarrow$ & $\downarrow$ \\
cryptographic privacy & - & \textcolor{red}{$\bm{\uparrow}$} & $\downarrow$ \\
obfuscation-based privacy & $\downarrow$ & \textcolor{red}{$\bm{\uparrow}$} & $\downarrow$ \\
asynchronization & $\downarrow$ & $\downarrow$ & \textcolor{red}{$\bm{\uparrow}$} \\
coordinator & - & $\downarrow$ & \textcolor{red}{$\bm{\uparrow}$} \\
sharing label & - & $\downarrow$ & \textcolor{red}{$\bm{\uparrow}$} \\
\bottomrule
\end{tabular}
\vspace{2mm}
\caption{Impact of different factors on \textit{utility-privacy-efficiency} trade-off. 
$\uparrow$, -, and $\downarrow$ denote increase, no change, and decrease, respectively.}
\vspace{-2mm}
\label{trade_off}
\end{table}

\subsubsection{Utility.} As the model complexity increases from linear models to collaborative filtering models, decision trees, and neural networks. The gradients and model parameters contain more information about private training data. Therefore, the risk of privacy leakage increases. Moreover, more complex models can achieve higher utility with more significant computation and communication costs. 

\subsubsection{Privacy.} Cryptographic approaches protect data privacy by encrypting data and intermediate results during FL. As no numerical error is introduced, cryptographic approaches only trade efficiency for privacy. As one exception, CKKS scheme~\cite{cheon2017homomorphic} allows approximated decryption with little numerical error, and it achieves much more efficient homomorphic operations on encrypted data than previous fully HE schemes.
Obfuscation-based approaches such as DP protects the membership privacy of data by adding noise. The added noise not only decreases model performance but also requires more iterations for convergence. Therefore, they trade utility and efficiency for data privacy. 

\begin{figure}[t!] 
\includegraphics[width=0.95\linewidth]{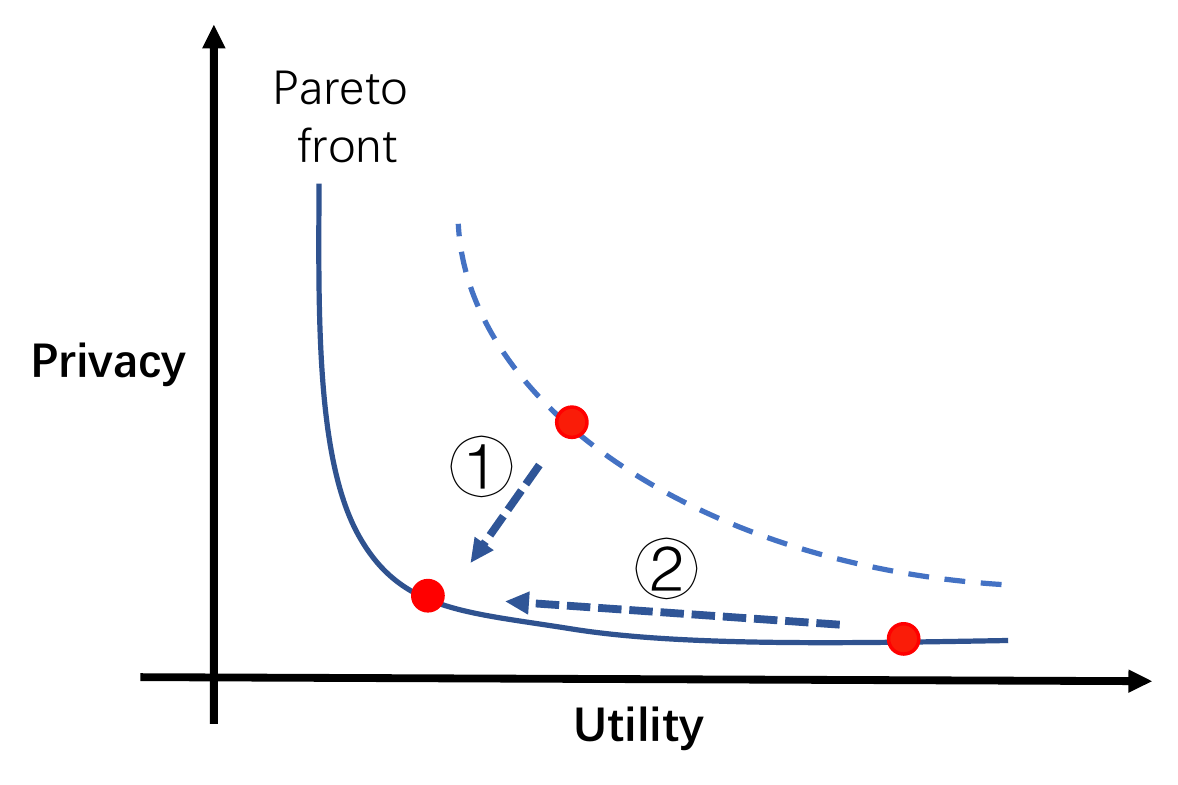}
\centering
\caption{ Two strategies to improve FL algorithms in the trade-off space: 1) Reduce system redundancy. 2) Trade one objective with little decline for another with significant benefit. Smaller values are preferred. Each red dot indicates an FL algorithm.  }
\label{img_pareto}
\end{figure}

\subsubsection{Efficiency.} 
Some works explore asynchronous updates to improve training efficiency. However, the plain-text asynchronous update may lead to utility decline and privacy leakage to the server if the gradients are not protected by cryptographic approaches. 
Some attempts~\cite{fu2021vf2boost} on asynchronization identified and reduced system redundancy to improve efficiency without sacrificing model performance and data privacy. 
A few works on VFL~\cite{wang2006classification,zhang2021secure} assume the share labels of the active party are shared among all passive parties. Such an assumption trades data privacy for training efficiency. 
Introducing a central coordinator for model aggregation can avoid high peer-to-peer communication costs when there are multiple participants in FL. However, the coordinator might infer private information from participants. Therefore, introducing a coordinator trades privacy for efficiency.

From the view of multi-objective optimization, the design of FL algorithms is to search the \textit{Pareto front} in the utility-privacy-efficiency space. 
Although some research claims that there is \textit{no free lunch} for the three objectives in FL~\cite{zhang2022nofreelunch} and optimizing one object leads to decline of other objectives, we can still optimize the objectives in two aspects. 
1) First, we can identify the redundancy in the FL algorithms and reduce the redundant operations and communication delays as in~\cite{fu2021vf2boost}. Therefore, a newly proposed solution can \textit{dominate} previous FL algorithms in the utility-privacy-efficiency space. 
2) Second, when the Pareto front is highly convex, we can find the \textit{keen points} by proposing \textit{non-dominated} approaches to trade one objective with a little decline for another with significant benefit, thus achieve higher overall benefit. 
For example, the CKKS scheme~\cite{cheon2017homomorphic} achieves higher computational efficiency with little or negligible utility decline via approximated decryption. 
Figure~\ref{img_pareto} takes utility and privacy as an example to demonstrate the two strategies of FL algorithm design. Each red dot indicates an FL algorithm.

\subsection{A Categorization of Federated Learning}\label{FLcategorization}
\begin{figure*}[t!] 
\includegraphics[width=16cm]{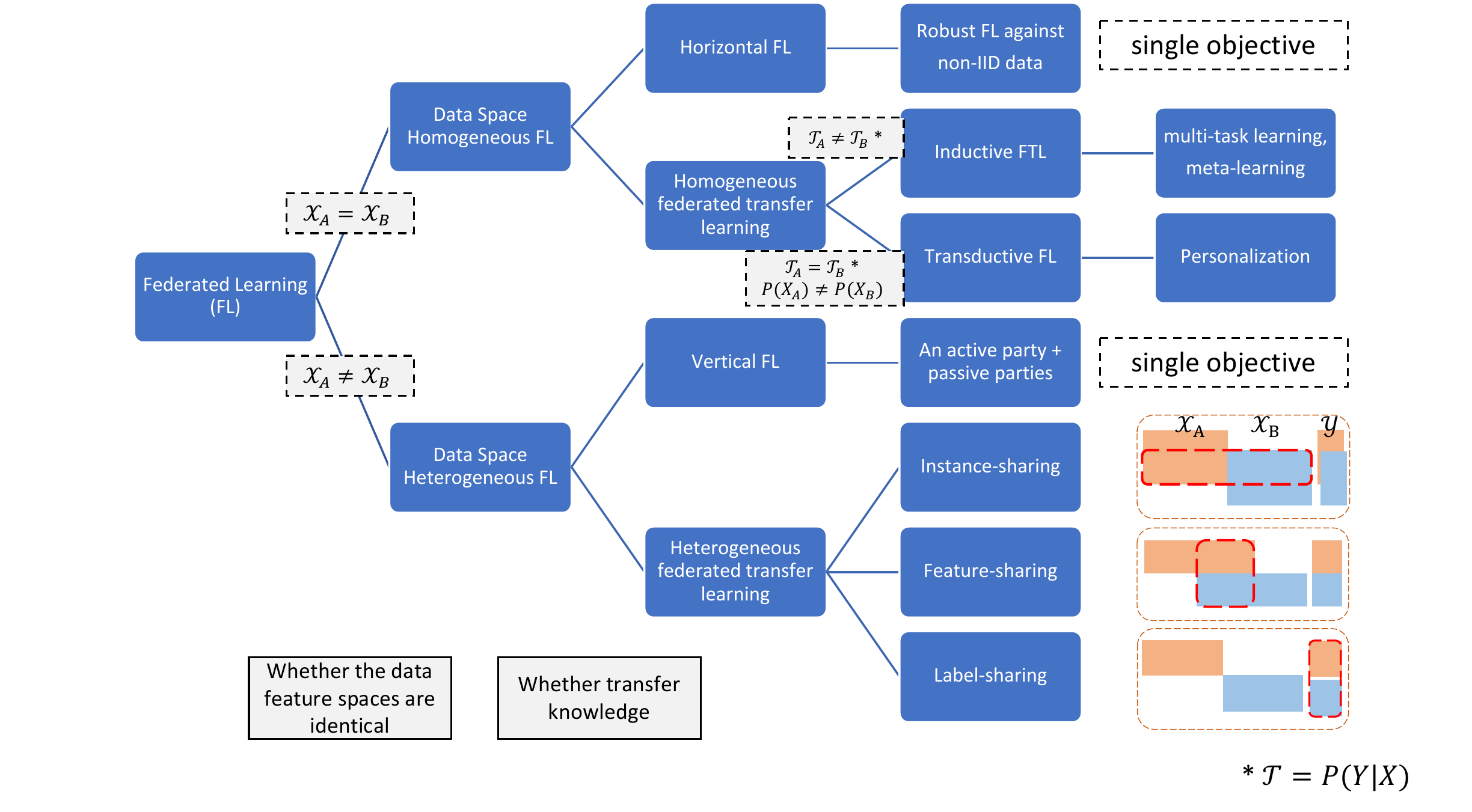}
\centering
\caption{Categorization of FL in terms of data space heterogeneity.}
\label{fl_categorization}
\end{figure*}

We categorize the typical scenarios of FL according to the data space heterogeneity. 
According to the feature space in each party, FL can be divided into \textit{Data Space-Homogeneous FL} and \textit{Data Space-Heterogeneous FL}. 
In this section, we demonstrate the details of each category. We demonstrate the categorization of FL in Figure~\ref{fl_categorization}

\subsubsection{Data Space Homogeneous Federated Learning} 
In data space-homogeneous FL, the data spaces $\mathcal{D}_i$ in each party $P_i$ are homogeneous in terms of both feature space $\mathcal{X}$ and label space $\mathcal{Y}$. 
Homogeneous FL is widely applied in cross-device scenarios. The devices (e.g., IoT devices and mobile phones) independently collect private data sharing the same feature space and label space. 
Since it was proposed, FL is meant to study the case where multiple clients holding homogeneous data train a single global model collaboratively under specific privacy constraints. Therefore, most existing researches focus on homogeneous FL.
Research on homogeneous FL can be categorized into two cases: \textit{horizontal federated learning} and \textit{homogeneous federated transfer learning}. The major difference is that, in HFL, there is one learning objective, and parties collaboratively train a global model on all data. However, in homogeneous federated transfer learning, each party optimizes its objective function and builds its model. 
Data distribution in homogeneous FL can be summarized as follows:
\begin{equation*}
\mathcal{X}_i = \mathcal{X}_j, \mathcal{Y}_i = \mathcal{Y}_j, \mathcal{I}_i \neq \mathcal{I}_j, \; \; \forall i, j \in [1, N], i \neq j.
\end{equation*}

We discuss the two cases in the following. 
Interested readers may refer to~\cite{Kairouz2019Survey105,9084352} for more details on data space homogeneous FL.

\subsubsubsection{a) Horizontal federated learning (HFL).}
HFL handles multi-client data that share the same set of features but are different in samples and is also referred to as \textit{federated optimization}. 
The pioneering works~\cite{McMahan2016ModelAvg,HBrendanMcMahan2016,DBLP:journals/corr/KonecnyMRR16} in FL mainly focus on this setting, where participants collaboratively train a global model based on their data with the help of a coordinator for model aggregation. 
The goal of HFL is to train a high-quality centralized global model. The objective can be formulated as:
\begin{equation*}
\min_{w\in \mathbb{R}^d} \left\{ f(w) := \frac{1}{n}\sum_{i=1}^{N}f_i(w) \right\} ,
\end{equation*}
where $w$ is the parameters of the federated model, and $f_i : \mathbb{R}^d \to \mathbb{R}$ is the training objective function that maps the model parameter set $w \in \mathbb{R}^d$ to a real valued training loss with respect to the training data $\mathcal{D}_i$ of party $P_i$.

\subsubsubsection{b) Homogeneous federated transfer learning (Homo-FTL).}
Different from HFL, which learns a single model for all parties even if the data is non-IID across different parties, homogeneous federated transfer learning aims to transfer knowledge from the other parties to some or each party. 
Homogeneous federated transfer learning optimizes multiple objectives to achieve the best performance for each client with non-IID data. Therefore, homogeneous federated transfer learning is highly related to multi-task learning, personalization, and meta-learning. A variety of multi-model approaches are proposed to tackle the non-IID data problem. 
We formulate the homogeneous federated transfer learning as:
\begin{equation*} 
\min_{W \in \mathbb{R}^{d \times N} }  \left\{ \mathcal{G}(W) := \sum_{i=1}^{N} f_i(w_i) \right\},
\end{equation*} 
where $W = [w_1, ..., w_N]$ is a $d \times N$-dimensional matrix that collects $w_1, ..., w_N$ as its columns. 



\subsubsection{Data Space Heterogeneous Federated Learning}
Here we briefly introduce the data distribution, categorization, and major application scenarios of heterogeneous FL to compare with homogeneous FL. In Section~\ref{sec-dataspace_FL}, we will discuss heterogeneous FL in depth. 
The data spaces $\mathcal{D}_i$ held by parties $P_i$ are non-identical in heterogeneous FL. 
Heterogeneous FL is common in \textit{cross-silo settings} where organizations collaborate to leverage the heterogeneous features held by other parties for accurate prediction. Unlike cross-device settings where numerous devices, each with limited samples, collaborate to boost data size, parties in cross-silo settings generally hold a large dataset but are from different domains. Therefore, tackling the lack of features is of the key importance in cross-silo settings. Heterogeneous FL provides a promising approach to leveraging different features in different parties. 
According to learning objectives and the distribution characteristics of the data spaces between parties, heterogeneous FL can be categorized into \textit{Vertical Federated Learning} and \textit{Heterogeneous Federated Transfer Learning}.

\subsubsubsection{a) Vertical federated learning (VFL).}
Vertical FL enables different parties holding multi-view data partitioned by feature to collaboratively train machine learning models without exposing one's own data~\cite{QiangYang2019}. It addresses the data silos and privacy problems together. 
Data distribution in vertical FL can be summarized as:$\mathcal{X}_i \cup \mathcal{X}_j = \emptyset, \; \mathcal{Y}_1 \neq \emptyset, \; \mathcal{Y}_k = \emptyset, \; \mathcal{I}_i = \mathcal{I}_j, \; \forall  i \neq j \in [1, N], \; \forall k \in [2, N].$
Here, party $P_1$ is the active party that owns label space, and the other parties $P_2, ... , P_N$ are passive parties that only own feature spaces. We denote that, in some cases, the active party may only hold the label space $\mathcal{Y}_1$ and has no feature $\mathcal{X}_1 = \emptyset$. 

The active party first sends the user ID to all parties in the inference process. Then each party inputs their own features to the model and collaboratively predicts the result. 
It is worth noting that if the user ID does not exist in all parties, the model inference can not be directly conducted without imputation. Therefore, the viability of the vertical federated model dramatically decreases as the number of parties increases due to the lack of sufficient linked training data and the hardness of alignment.

\subsubsubsection{b) Heterogeneous federated transfer learning (Hetero-FTL).}
Instead of collecting different features from separate parties to train a more accurate model, heterogeneous federated transfer learning aims to transfer knowledge from the heterogeneous data spaces in other parties to boost the performance of the local model. 
Therefore, heterogeneous federated transfer learning releases the ID alignment requirement in vertical FL. When the inference data is not aligned between parties, the parties in heterogeneous FL can leverage the trained local model for inference. Since the model is trained by transferring knowledge from other parties, the model can achieve higher performance than that trained solely on local data. 
In heterogeneous federated transfer learning, data can be linked in three ways. 
1) parties have an overlap of ID space but no intersection of feature spaces. 
2) parties have an intersection of feature space but no overlap of ID spaces.
3) parties have neither feature space overlap nor ID space overlap. 
In section~\ref{sec-dataspace_FL}, we will discuss heterogeneous FL in depth.

\subsubsection{Summary}

We summarize the main objective of each category of FL in Table~\ref{FL_category_comparison}. 
There has been many surveys investigating papers that tackle non-IID data in FL, which is statistical heterogeneity~\cite{tan2022personalizedflsurvey,BingshengHe2019Survey,zhang2018survey11,Kairouz2019Survey105}. 
To our best knowledge, there is no survey investigate on data space heterogeneous FL, especially data space heterogeneous federated transfer learning. 
To mitigate this gap, we deliver a comprehensive investigation on data space heterogeneity in Section~\ref{sec-dataspace_FL}. 
We summarize the transfer learning techniques used to tackle data space heterogeneity in Section~\ref{sec-TL_HeteroFL}.

\begin{table*}[ht]
\vspace{8mm}
\centering
\footnotesize
\begin{tabular}{c c | c  c  } 
 \toprule
   \multicolumn{2}{c}{Categorization} & Data & Objective \\
   \midrule
   \multirow{2}{*}{\begin{tabular}{@{}c@{}} Data Space \\ Homogeneous FL \end{tabular}} & \multirow{1}{*}{\begin{tabular}{@{}c@{}} HFL \end{tabular}} & \multirow{1}{*}{\begin{tabular}{@{}c@{}} IID/ Non-IID, homo-feature \end{tabular}} &  One single global model  \\
   \cmidrule{2-4} 
   & Homo-FTL & Non-IID, homo-feature & Personalized local models for all parties \\
   \midrule
   \multirow{2}{*}{\begin{tabular}{@{}c@{}} Data Space \\ Heterogeneous FL \end{tabular}} & VFL & hetero-feature &  One single global model \\ 
    \cmidrule{2-4} 
   & Hetero-FTL & hetero-feature & Personalized local models for all parties  \\
 \bottomrule
\end{tabular}
\vspace{2mm}
\caption{Comparison of different FL settings.}
\vspace{-3mm}
\label{FL_category_comparison}
\end{table*}







\subsection{Security Definition and Privacy Threats}

\subsubsection{Security Definition}
In the cryptographic study, based on the extent to deviate from the protocol, the adversaries in FL can be categorized into two types: \textit{semi-honest} (a.k.a \textit{honest-but-curious}) adversaries and \textit{malicious} adversaries~\cite{lindell2017simulate}. The definitions of security levels corresponding to each type of adversary are listed below: 
\begin{itemize}
    \item \emph{Semi-honest security}: The adversaries strictly follow the prespecified protocol without deviation. However, they will collect all received intermediate data and try to derive knowledge from it. Protocols achieving semi-honest security prevent inadvertent information leakage between parties. 
    \item \emph{Malicious Security}: The adversaries can arbitrarily deviate from the protocol in their attempt to cheat. The adversaries can only cause the honest parties to abort. Furthermore, if the honest parties obtain output, they are guaranteed that it is correct. Their privacy is always preserved.
\end{itemize}

\subsubsection{Privacy Threats}
The idea of information privacy is to have control over the collection and processing of one's information.
In ML tasks, the participants usually take up three different roles: 1) as the input party, e.g., the data owner, 2) as the computation party (e.g., the model builder and inference service provider), and 3) as the result party (e.g., the model querier and user)~\cite{bogdanov2014privacy}. 

Attacks on ML may happen at any stage, including data publishing, model training, and model inference. Adversaries can attack each stage to breach data privacy and infer private knowledge such as model parameters, raw data, and membership. 
We introduce three types of attacks on data privacy: \textit{model-inversion attack}, \textit{data-reconstruction attack}, and \textit{membership-inference attack}. We will discuss each attack in the following. 

\begin{itemize}
    \item  \emph{Model-inversion attack} can happen in the model training stage and model inference stage. The adversary may attempt to reconstruct the model parameters from intermediate gradients or model predictions or train a mimic model for malevolent purposes.
    \item \emph{Data-reconstruction attack} can happen in the data publishing stage and model training stage. The computation party aims to reconstruct the data providers' raw data or learn more information about the data providers than what the model builders intend to reveal. 
    \item \emph{Membership-inference attack} can happen in the data publishing stage, model training stage, and model inference stage. The goal is to know whether an instance is inside the training set of the model. The adversary is assumed to have black-box access or white-box access to the model and an instance. 
\end{itemize}

\subsection{Privacy Preservation Techniques}\label{ch03:sec_ppt}
There are four types of privacy preservation approaches widely used for privacy preserving machine learning, namely 1) secure multi-party computation (MPC), 2) homomorphic encryption (HE), 3) differential privacy (DP), and 4) trusted execution environment (TEE).
We elaborate these techniques in Appendix~\ref{appendix_priv_pres_tech}.

\section{Data Space Heterogeneity}\label{sec-dataspace_FL}

In data space heterogeneous FL, the feature spaces $\mathcal{X}_i$ held by parties $P_i$ are non-identical. 
Such a setting is meaningful and practical in various cross-silo industrial applications, such as recommendation, advertisement, fraud detection, and credit extension. Unlike HFL, it allows non-competitive participants from different areas to cooperate with their unique features. 
For example, a loan company with demographic features, an e-commercial company with customers' purchase history, and an online video company with users' movie browsing history can cooperate and leverage the heterogeneous features via vertical FL to boost the performance of credit extension business of the loan company. 
According to the learning objectives and the distribution characteristics of the data spaces between parties, we categorize heterogeneous FL into \textit{Vertical Federated Learning} (VFL) and \textit{Heterogeneous Federated Transfer Learning} (Hetero-FTL), as demonstrated in Figure~\ref{FLcategorization}. We will elaborate on them in the following.


\subsection{Vertical Federated Learning}
\begin{figure}[!htb] 
\includegraphics[width=0.95\linewidth]{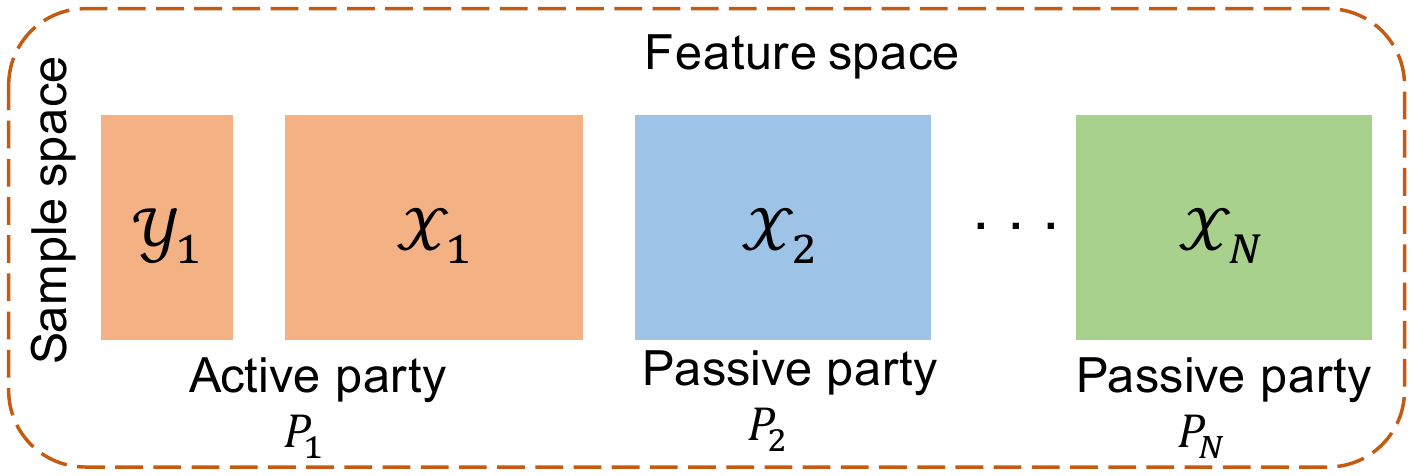}
\centering
\caption{Data distribution in VFL}
\label{vfldatadistribution}
\end{figure}

VFL enables multiple parties with multi-view data partitioned by feature to collaboratively train machine learning models without violating data privacy and confidentiality~\cite{QiangYang2019}. 
It addresses the data silos and privacy problems together. 
The collaborating parties own data of the same set of users/instances but with disjoint features, and only one party holds the labels. 
Taking the collaboration between a book recommender and an auxiliary data provider as an example, the intersection of their ID spaces is large. The book recommender records the consumer's rating, browsing history, and item features, including book titles, authors, descriptions, and publishers. At the same time, an auxiliary data provider owns the demographic information such as age, phone number, location, and gender. 
Their feature spaces are very different. Incorporating auxiliary user information helps the book recommender predict more precise recommendations. Therefore, the book recommender can collaborate with the auxiliary data provider to conduct VFL to train a model on the combination of the feature spaces of the two parties. 
Figure~\ref{vfldatadistribution} demonstrates the data distribution in VFL. 

\begin{definition}\textbf{(Vertical Federated Learning).}
Given $N$ parties and each party $P_i$ contains a set of features and labels of some instances, i.e., 
$\mathcal{D}_i = \{\bm{X}_i, \bm{I}_i, \bm{Y}_i \}$. $P_i$ has features $\bm{X}_i \in \mathcal{X}_i$, instances $\bm{I}_i \in \mathcal{I}_i$, and labels $\bm{Y}_i \in \mathcal{Y}_i$. 
$P_1$ is an active party that holds labels. 
The data distribution is:
\begin{align*}
    & \mathcal{X}_i \cup \mathcal{X}_j = \emptyset, \; \mathcal{Y}_1 \neq \emptyset, \; \mathcal{Y}_k = \emptyset, \; \mathcal{I}_i = \mathcal{I}_j, \\
    & \forall  i \neq j \in [1, N], \; \forall k \in [2, N].
\end{align*}

VFL aims to train a model without revealing private data of each party:
\begin{equation*}
\argmin_{\bm{W}\in \mathbb{R}^{d\times N}}  \; L \left( \bm{Y}_1, f_{\bm{W}}\left(\bm{X}_1, z(\{\bm{X}_{i=2}^N\}) 
\right) \right) ,
\end{equation*}
where $\bm{W} = [\bm{w}_1, ..., \bm{w}_N]$ is a $d \times N$-dimensional matrix that collects sub-models $\bm{w}_1, ..., \bm{w}_N$ of all parties as its columns, 
$f_{\bm{W}}(\cdot , \cdot)$ is the prediction model parameterized by $\bm{W}$ 
is the training objective function, $z(\cdot)$ stands for the data processing function that exchanges intermediate values in training process. 
And $L(\cdot , \cdot)$ is a loss function that maps the model parameter set $\bm{W}$ to a real valued training loss with respect to the training data $\{\bm{X}_i\}_{i=1}^N$ of all parties $\{P_i\}_{i=1}^N$.
\end{definition}

We denote that, in some cases, the active party $P_1$ may only hold the label space $\mathcal{Y}_1$ and has no feature $\mathcal{X}_1 = \emptyset$. 
In practice, parties need to find the instance IDs that co-occur in all parties as training instances. Then, the shared instances are linked to form the \textit{training instances} of each party. 
Exchanging instance IDs with other parties can leak membership information, which is not acceptable in VFL. 
Privacy-preserving instance alignment allows different parties to find shared instances. Meanwhile, it protects the membership privacy of records or users. 
We will introduce instance alignment in Section~\ref{instanceAlignment}. 

The active party first sends the user ID to all parties in the inference process. Then each party inputs its features to the model and collaboratively predicts the result. 
It is worth noting that if the user ID does not exist in all parties, the model inference can not be directly conducted without imputation. Therefore, the availability of the vertical federated model dramatically decreases as the number of parties increases due to the lack of sufficient aligned training data. 

\subsubsection{Private Instance Alignment}\label{instanceAlignment}
Recall that instance alignment is essential in the data preprocessing phase of vertical federated model training. 
If the instance IDs of training data are exchanged for instance alignment, one party can easily learn all membership information from the other, leading to severe privacy violations. 
Delegating the instance alignment to an external party is also risky and infeasible in reality. 
Privacy-preserving instance alignment techniques, such as Private Set Interaction (PSI)~\cite{dachman2009efficient,de2010practical}, Privacy-Preserving Record Linkage (PPRL)~\cite{vatsalan2013taxonomy,hall2010privacy}, and Private Entity Resolution (PER)~\cite{christendata} can be applied to tackle this issue. 
In PSI, separate parties compare encrypted versions of their sets to compute the intersection. Neither party reveals anything to the others except for the elements in the intersection. 
For example, Bellare et al.~\cite{bellare1996keying} introduce a keyed-hash message authentication code to encrypt the IDs for comparison. 
PSI requires exact matching of identifiers among parties and is suitable for identifiers like phone numbers, national identification numbers, or social security numbers. 
Nevertheless, instances can be linked by combining features with errors, such as name, birthday, and location. 
PPRL and PER allow fuzzy matching that tolerates errors in the identifiers. 
Schnell et al. ~\cite{schnell2009privacy} adopt Bloom filters to perform fuzzy private record linkage on identifiers by computing string similarities. 
Angelou et al.~\cite{angelou2020asymmetric} implement a cross-platform, open-source library for asymmetric PSI and PSI-cardinality, which combines DDH-based protocols with compression based on Bloom filters to reduce communication in preprocessing.  
Harty et al.~\cite{hardy2017private} utilize Cryptographic Longterm Key (CLK) to compute encrypted identifiers and use Dice coefficient to match instances. 
After alignment, neither party knows the aligned instances except for an encrypted mask. 
The training process with the encrypted ID mask is costly as all un-aligned instances should be used for training. 
Since the instance alignment process is independent of the training process, participants can still dynamically align new data and add the aligned new data for training in the training process.

\subsubsection{Existing Studies}
In this section, we categorize the model-building approaches based on their base models. Then, we also discuss the privacy leakage in VFL. 
Finally, we demonstrate and compare the existing studies in VFL.

\subsubsubsection{a) Linear Models.}
The study of training a regression model in privacy-preserving distributed learning over vertically partitioned data can trace back to Sanil~\cite{sanil2004privacy}, where the authors use secret sharing in a secure linear regression model to ensure privacy in their approach. 
Liu et al.~\cite{liu2019communication} apply stochastic block coordinate descent to VFL. Before exchanging the intermediate results, each party updates its local parameters for multiple rounds. 
Harty et al.~\cite{hardy2017private} leverage entity resolution and Paillier's homomorphic encryption to conduct vertical logistic regression between two parties with privacy against semi-honest adversaries. 
Feng et al.~\cite{feng2020multiparticipant} propose MMVFL, where each party locally predicts labels based on its own features. It also studies the feature selection problem in the vertical federated linear model.

Traditional VFL requires synchronous computation, which is inefficient. To tackle this problem, Zhang et al.~\cite{zhang2021secure} integrate with backward updating mechanism and bi-level asynchronous parallel architecture to train a linear model. The convergence rates of the proposed algorithms are theoretically analyzed under strongly convex and non-convex conditions. The security against a semi-honest adversary is proved as well.  
Xu et al.~\cite{xu2021fedv} proposed FedV for secure gradient computation for linear models, logistic regression, and support vector machines. It introduces an aggregator and uses functional encryption schemes to reduce communication among parties.
Li et al.~\cite{wang2020hybrid} introduce hybrid differential privacy to protect membership differential privacy for conducting vertical federated logistic regression. 
Xu et al.~\cite{Xu2019AchievingDP} also applied differential privacy to vertical federated linear regression. 
Gu et al.~\cite{gu2020privacy} study the asynchronous multi-party vertical federated linear model by proposing a tree-structured communication protocol. They also provide the convergence rate of the proposed AFSGD-VP approach. 
Chen et al.~\cite{chen2020practical} and Gao et al.~\cite{gao2020FedFM} apply factorization machine with second-order feature interaction into vertical federated recommender systems.

\subsubsubsection{b) Neural Networks.}
Hu et al.~\cite{hu2019fdml} consider a multi-view learning setting, where all parties hold the labels. Instead of exchanging the intermediate results, it aggregates each participating party's plain-text local prediction results. The intermediate local prediction results for each training or testing sample of each local party are sent to a server for aggregation. To protect private local features against malicious servers or clients, DP is applied by perturbing updated local predictions. 
Vepakomma et al.~\cite{vepakomma2018split} study the VFL on neural networks by proposing SplitNN, where a neural network is divided into two parts. Each party trains several layers and shares the output with the server in every iteration. 
A framework for multi-headed SplitNN is implemented~\cite{romanini2021pyvertical}. 
Chen et al.~\cite{chen2020vafl} study vertical asynchronous FL with differential privacy based on neural networks. However, the labels of the active party are transmitted to an honest server in~\cite{romanini2021pyvertical,chen2020vafl}. 
Zhou et al.~\cite{zhou2021vertically} investigate graph neural networks for privacy-preserving node classification. 
Recently, Liu et al.~\cite{Liu2022FedBCD} introduce the stochastic block coordinate descent algorithm to VFL to achieve communication efficient training process. The proposed FedBCD can prevent an adversary from inferring exact raw training data and has superior convergence than standard VFL algorithms. 

\subsubsubsection{c) Decision Trees.}
Recently, many studies used decision trees to design algorithms for VFL. 
This is because tree-based models are feasible to be adapted to VFL by nature. 
Du et al.~\cite{du2002building} first propose a decision tree for vertically partitioned data by using a scalar product protocol to compute information gained at a node. The labels are shared with all parties for training. 
Wang et al.~\cite{wang2006classification} also assume the labels in the training data could be shared with all parties, which is counterfactual in most real-world cases. They propose a secure decision tree construction trained in plain text by applying PSI. 
Li et al.~\cite{li2020practical} leverage locality-sensitive hashing into vertical federated GBDT to protect the privacy of each party with relaxed privacy constraints.
Additive homomorphic encryption is adapted for privacy protection in several frameworks based on GBDT~\cite{KeweiCheng2019,feng2019securegbm,fu2021vf2boost}.
The work in ~\cite{wu2020privacy} studies privacy-preserving federated vertical decision trees by proposing Pivot. Pivot provides protection against an honest-but-curious adversary without relying on any trusted third party based on threshold Paillier encryption and secret sharing. It identifies two privacy leakages of the plain-text trained model and proposes to use differential privacy and message authentication code to enhance the protocol. Pivot can also be extended to ensemble models such as random forest and gradient boost decision trees. 
Liu et al.~\cite{Liu_2020fedforest} propose Federated Forest based on the CART tree and bagging. Each party stores the split information with respect to its own features. The proposed approach achieves comparable performance to the centralized version. 
Fu et al.~\cite{fu2021vf2boost} designed an efficient vertical federated GBDT algorithm named VF2Boost, by proposing a concurrent training protocol and customized cryptographic operations to speed up homomorphic encryption. 
Federated gradient boosting models sequentially build a boosting model with the base learner, which can be communication inefficient and time-consuming. Therefore, Han et al.~\cite{han2022fedgbf} propose to integrate boosting and bagging to build multiple decision trees in parallel. 
Chen et al.~\cite{Fed-EINI9671749} explore the interpretability in the inference phase for decision tree ensembles in VFL by canceling decision paths and adapting an efficient, secure inference method.

\begin{table*}[ht]
\centering
\footnotesize
\begin{tabular}{ c | c c c c c } 
 \toprule
 & Utility & \multicolumn{1}{c}{Privacy}  & \multicolumn{3}{c}{Efficiency} \\
 \cmidrule{4-6} 
 VFL Studies & model type & privacy mechanism & share labels & asynchronous &  aggregator \\
 \midrule
	Sanil et al.~\cite{sanil2004privacy} & LM & MPC & \texttimes & \texttimes & \texttimes \\
	Zhang et al.~\cite{zhang2021secure} & LM & MPC & \checkmark & \checkmark & \checkmark \\
	Harty et al.~\cite{hardy2017private} & LM & HE & \texttimes & \texttimes & \checkmark \\
	Xu et al.~\cite{xu2021fedv} & LM & HE & \texttimes & \texttimes &  \checkmark \\
	Li et al.~\cite{wang2020hybrid} & LM & DP & \texttimes & \texttimes & \checkmark \\
	Xu et al.~\cite{Xu2019AchievingDP} & LM & DP & \texttimes & \texttimes & \checkmark \\
	Liu et al.~\cite{liu2019communication} & LM & PT & \texttimes & \checkmark & \texttimes \\
	Gu et al.~\cite{gu2020privacy} & LM & PT & \texttimes & \checkmark & \texttimes \\
	Feng et al.~\cite{feng2020multiparticipant} & LM & PT & \texttimes & \texttimes & \checkmark \\
	Chen et al.~\cite{chen2020practical} & FM & MPC & \texttimes & \texttimes & \checkmark \\
	Gao et al.~\cite{gao2020FedFM} & FM & HE, MPC & \texttimes & \texttimes & \texttimes \\
	Liu et al.~\cite{Liu2022FedBCD} & NN, LM & PT & \texttimes & \checkmark & \texttimes \\
	Hu et al.~\cite{hu2019fdml} & NN, LM & DP & \texttimes & \texttimes & \checkmark \\
	Vepakomma et al.~\cite{vepakomma2018split} & NN & PT & \texttimes & \checkmark & \texttimes \\
	Romanini et al.~\cite{romanini2021pyvertical} & NN & PT & \texttimes & \checkmark & \texttimes \\
	Chen et al.~\cite{chen2020vafl} & NN & DP & \texttimes & \checkmark & \checkmark \\
	Du et al.~\cite{du2002building} & DT & PT & \texttimes & \texttimes & \checkmark \\
	Wang et al.~\cite{wang2006classification} & DT & PT & \checkmark & \texttimes & \checkmark \\
	Cheng et al.~\cite{KeweiCheng2019} & DT & PT & \texttimes & \texttimes & \texttimes \\
	Feng et al.~\cite{feng2019securegbm} & DT & PT & \texttimes & \texttimes & \texttimes \\
	Wu et al.~\cite{wu2020privacy}  & DT & HE, MPC & \texttimes & \texttimes & \texttimes \\
	Li et al.~\cite{li2020practical} & DT & LSH & \texttimes & \texttimes & \checkmark \\
	Liu et al.~\cite{Liu_2020fedforest} & DT & HE & \texttimes & \texttimes & \checkmark \\
	Fu et al.~\cite{fu2021vf2boost} & DT & HE & \texttimes & \checkmark & \checkmark \\
 \bottomrule
\end{tabular}
\vspace{2mm}
\caption{Comparison among existing works of VFL. LM denotes Linear Models. FM denotes Factorization Machines. NN denotes Neural Networks. DT denotes Decision Trees. HE denotes Homomorphic Encryption. MPC denotes Secure Multiparty Computation. DP denotes Differential Privacy. PT denotes Plaintext. LSH denotes locality-sensitive hashing.}
\label{vfl_summary}
\vspace{-3mm}
\end{table*}

\subsubsubsection{d) Privacy Leakage in VFL.}
The privacy leakage in VFL is studied~\cite{weng2020privacy,gao2020privacy,gao2020FedFM}. 
Gao et al.~\cite{gao2020FedFM} demonstrate that in a plain-text vertical federated factorization machine, participants can infer model parameters as well as raw training data according to intermediate results. 
The privacy threats in vertical matrix factorization are investigated as well~\cite{gao2020privacy,chai2019secure}. 
Weng et al.~\cite{weng2020privacy} design reverse sum attack and reverse multiplication attack to steal the private training data in linear models without affecting the model accuracy. 

Several works explored the label inference attack in VFL~\cite{liu2021defendlabelvfl,fu2022labelinferencevfl}. 
Fu et al.~\cite{fu2022labelinferencevfl} proposed three label inference attack methods: passive model completion, active model completion, and direct label inference attack. In model completion attacks, the attacker requires extra auxiliary labeled samples or sample-level gradients for fine-tuning its local model. 
Liu et al.~\cite{liu2021defendlabelvfl} consider a more challenging scenario where the communication between participants in VFL is black-boxed (e.g., encrypted via HE). In the proposed gradient inversion label inference attack and defense, the attacker only has access to the gradients of the final local model.

\subsubsection{Summary}
We summarize and compare the reviewed studies in VFL in Table~\ref{vfl_summary}.


\subsection{Heterogeneous Federated Transfer Learning}

Although VFL boosts model performance by aligning the features from isolated parties, it is hard to collect sufficient aligned data for training in the real world. On the other hand, VFL can not conduct inference on un-aligned test data. 
Therefore, it is desirable to leverage the numerous un-aligned heterogeneous data for FL and inference on un-aligned test data. 

Heterogeneous federated transfer learning (Hetero-FTL) aims to transfer knowledge from the heterogeneous feature spaces of other parties to boost the performance of the local model. 
Hetero-FTL relaxes the ID alignment requirement in VFL, and transfers the knowledge of the un-aligned training data to other parties to improve the performance of their local models.
When the test data are not aligned among all parties, the parties holding features can leverage the local model trained in Hetero-FTL for inference. 
Since the model is trained by transferring knowledge from other parties, the model can achieve higher performance than that trained solely on local data. 

In Hetero-FTL, we identify three settings according to the way data are associated:
\begin{itemize}
\item \textbf{Instance-sharing Hetero-FTL.} Parties share an overlap of ID space for a part of instances, while there is no intersection of their feature spaces. In contrast to VFL, parties can utilize the un-aligned local data to train local models here. Moreover, the models learned in instance-sharing Hetero-FTL can conduct inference on un-aligned test data with solely local features. 
\item \textbf{Feature-sharing Hetero-FTL.} Parties have no overlap of ID spaces, while there is an intersection of their feature spaces. The shared feature space is leveraged to transfer knowledge between heterogeneous feature spaces of different parties. 
\item \textbf{Label-sharing Hetero-FTL.} Parties may have an overlap of neither feature space nor ID space. However, the label space in their tasks is the same. Therefore, the parties can each train a model to map their heterogeneous feature spaces into a common latent space. Then, they collaboratively build a model to project the common latent space to the label space via HFL. 
\end{itemize}

In each setting, the association between parties is limited and weak. It brings new challenges to transferring knowledge among privacy-sensitive parties in heterogeneous FL. 
We identify four transfer learning methods adopted by existing Hetero-FTL research, including representation learning, knowledge distillation, collaborative filtering, and auxiliary public dataset. 
We will elaborate on each setting in the following. For each setting, we categorize and review the existing works according to the transfer learning methods.


\subsubsection{Instance-sharing Hetero-FTL}
\begin{figure}[!htb] 
\subfigure[single-active party]{
\label{instanceSharing1}
\includegraphics[width=0.45\linewidth]{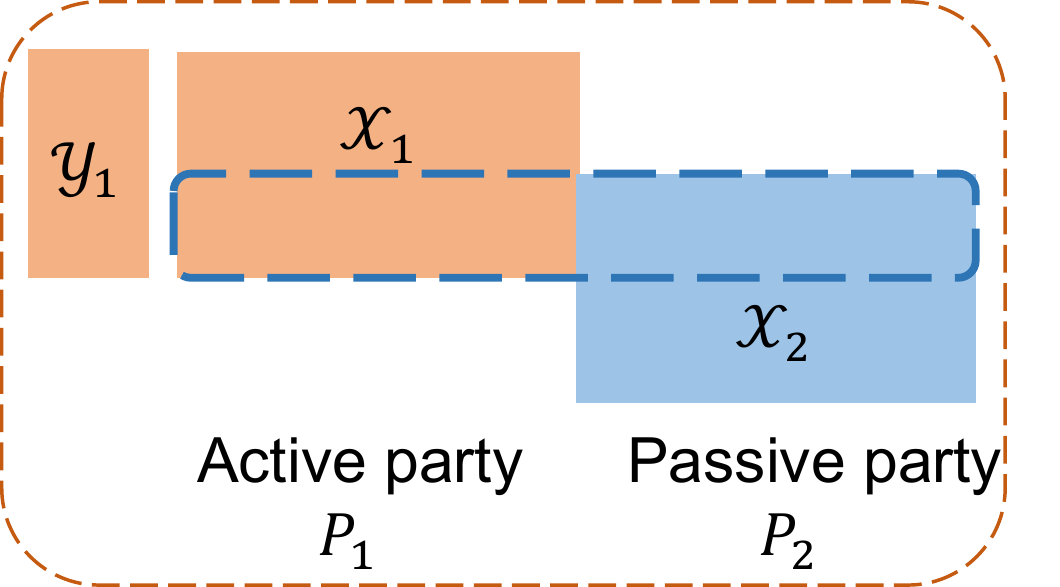} }
\subfigure[multi-active party]{
\label{instanceSharing2}
\includegraphics[width=0.45\linewidth]{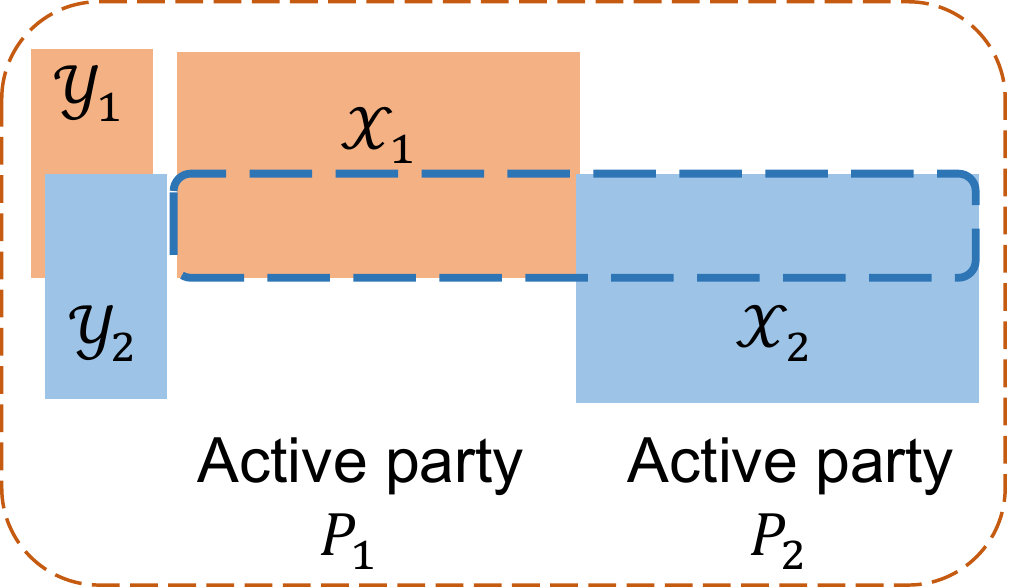} }
\centering
\caption{Data distribution in instance-sharing Hetero-FTL}
\label{instanceSharing}
\end{figure}

In instance-sharing Hetero-FTL, there are some aligned data samples between parties, making it possible to transfer knowledge across parties. However, different parties have no shared feature. The data distribution in instance-sharing Hetero-FTL is demonstrated in Figure~\ref{instanceSharing}. 

\begin{definition}\textbf{(Instance-sharing Hetero-FTL).}
Given $N$ parties and each party $P_i$ contains a set of features and labels of some instances, i.e., 
$\mathcal{D}_i = \{\bm{X}_i, \bm{I}_i, \bm{Y}_i \}$. $P_i$ has features $\bm{X}_i \in \mathcal{X}_i$, instances $\bm{I}_i \in \mathcal{I}_i$, and labels $\bm{Y}_i \in \mathcal{Y}_i$. 
There are $k$ active parties $\exists k \in P_{act}=\{k | k\in[1, N]\}\neq \emptyset$, 
where $P_{act}$ is a non-empty set that contains the index of active parties.
The data distribution is:
\begin{align*}
	& \mathcal{X}_i \cap \mathcal{X}_j = \emptyset, \; \mathcal{I}_i \cap \mathcal{I}_j \neq \emptyset, \mathcal{Y}_k \neq \emptyset, \; \\
	& \forall  i \neq j \in [1, N], \; \exists k \in P_{act}=\{k | k\in[1, N]\}\neq \emptyset.
\end{align*}
Instance-sharing Hetero-FTL aims to train a model for each active party without revealing private data of each party:
\begin{align*}
	\argmin_{\bm{W}\in \bm{R}^{d\times N}} \sum_{i\in P_{act} } \lambda_{i} L \Bigl( \bm{Y}_i, f_{\bm{W}} \bigl( 
	& I_{\cap}, \bm{X}_i, z(\{\bm{X}_j\}_{j\in [1, N] \setminus \{i\}} )  \\
	& | \{\mathcal{D}_{j} \}_{j\in [1, N] \setminus \{i\}}  \bigr) \Bigr),
\end{align*}
where $\lambda_{i}$ is the weight for balancing the performance on each party, $I_{\cap}$ is the index of overlapped instances. $\bm{W} = [\bm{w}_1, ..., \bm{w}_N]$ contains local models $\{\bm{w}_i\}_{i\in [1, N]}$ corresponding to each party $\{P_{i}\}_{i\in [1,N]}$. 
\end{definition}

After federated transfer learning, the trained model $f_{\bm{W}}(\cdot)$ predicts both aligned and un-aligned target data.
For aligned instances, model $f_{\bm{W}}(\cdot)$ takes features from both active and passive parties as input for prediction, which is similar to VFL. 
However, for un-aligned instances, model $f_{\bm{W}}(\cdot)$ leverages local model $\bm{w}_i$ to predict on local features $\bm{X}_i$. 
In contrast, VFL can only predict on aligned test data, which greatly restricts its application to real-world problems where most test data are unaligned. 
Moreover, the prediction performance of $f_{\bm{W}}(\cdot)$ over local un-aligned data outperforms the naive model trained locally, as instance-sharing Hetero-FTL transfers knowledge from other parties to local models $\bm{w}_i$ in active party $P_i$.

The instance-sharing Hetero-FTL problems differ according to the number of active parties. Specifically, there are two representative cases in instance-sharing Hetero-FTL: 

\begin{itemize}
\item For the single-active party case, we have $\mathcal{Y}_1 \neq \emptyset, \mathcal{Y}_k = \emptyset, \;\; \forall k \in [2, N]$. All $N$ parties collaboratively train parameters to improve the performance on $\mathcal{D}_1$. Figure~\ref{instanceSharing1} demonstrate the single-active party case. 
\item For the multi-active party case, we have $\mathcal{Y}_i \neq \emptyset, \forall i \in [1, N]$. Each active party has labeled data and the objective is to train a model that improves performance for each active party by federated knowledge transfer. Figure~\ref{instanceSharing2} demonstrates the multi-active party case.
\end{itemize}

\begin{figure}[!htb] 
\includegraphics[width=\linewidth]{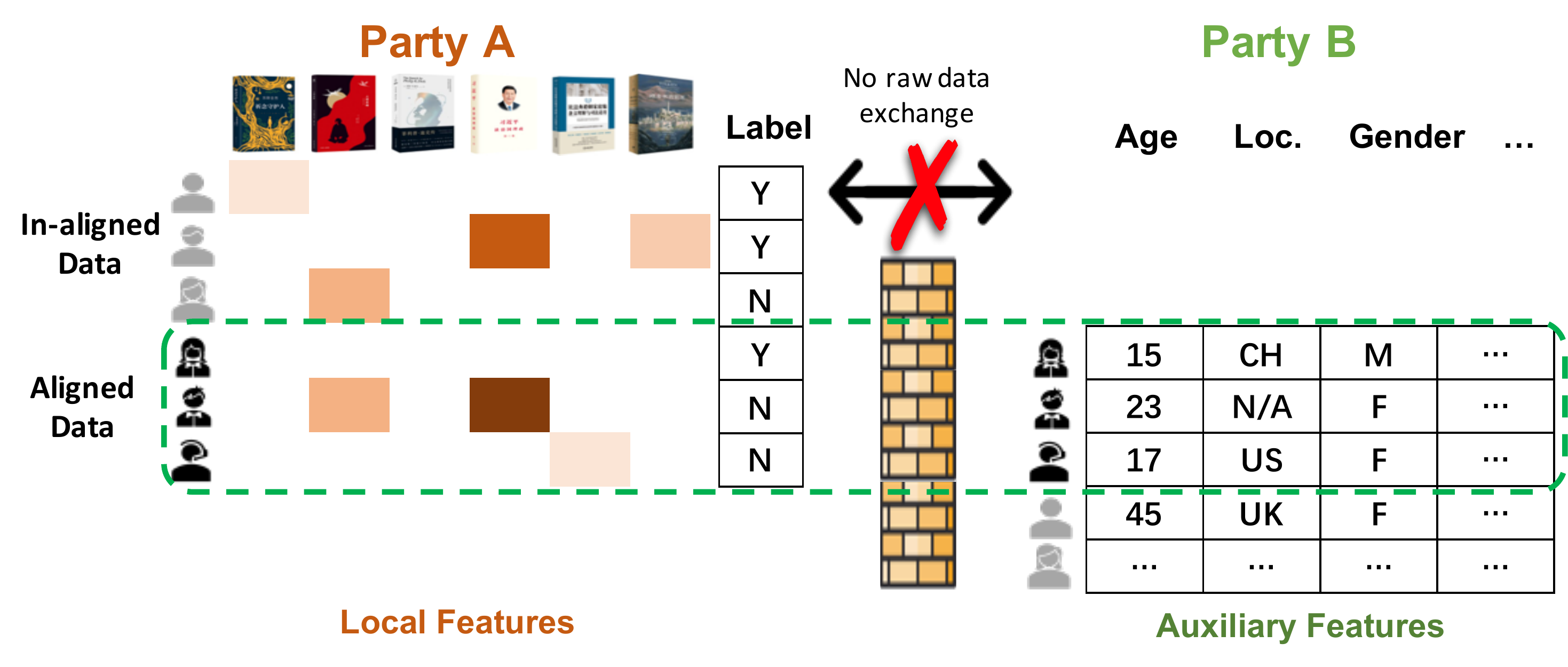}
\centering
\caption{Overview of a two-party instance-sharing Hetero-FTL problem. In this example, party A has a labeled dataset, and party B has some auxiliary features. Some of the data co-occur in both parties' datasets and can be aligned. The two parties aim to build a recommender system collaboratively with data from both parties. However, they cannot exchange their private data due to privacy and security concerns.}
\label{instanceSharingDemo}
\end{figure}

\subsubsubsection{a) Application examples.}
Faced with the fact that most instances can hardly be aligned across multiple parties in the real world, it is desirable to collaboratively train ML models with limited aligned data and no common features. 
Instance-sharing Hetero-FTL is a promising relief from the stringent restriction on data alignment in VFL. 
As illustrated in Figure~\ref{instanceSharingDemo}, a recommender A wants to build a recommendation system with the help of an auxiliary data provider B. The recommender A holds the user's rating, browsing history, and item features such as price, category, color, etc. At the same time, the auxiliary data provider B holds the demographic information, including occupation, education, gender, etc. 
The recommender and the data provider have a part of users in common. These common users can be aligned by privacy-preserving instance alignment~\ref{instanceAlignment}. 

The recommender aims to boost the performance of the recommender system by leveraging the data from the auxiliary data provider in a privacy-preserving way. 
Since only limited aligned users are in training data, VFL cannot achieve high accuracy and cannot be applied to un-aligned test data. 
The instance-sharing Hetero-FTL helps build a better recommender system by making full use of all data. 

\subsubsubsection{b) Existing studies.}
Various transfer learning approaches have been explored to tackle instance-sharing Hetero-FTL problems, including parameter sharing and knowledge distillation. We elaborate on these approaches in Section~\ref{sec-TL_HeteroFL}.

\subsubsection{Feature-sharing Hetero-FTL}
\begin{figure}[!htb] 
\includegraphics[width=0.7\linewidth]{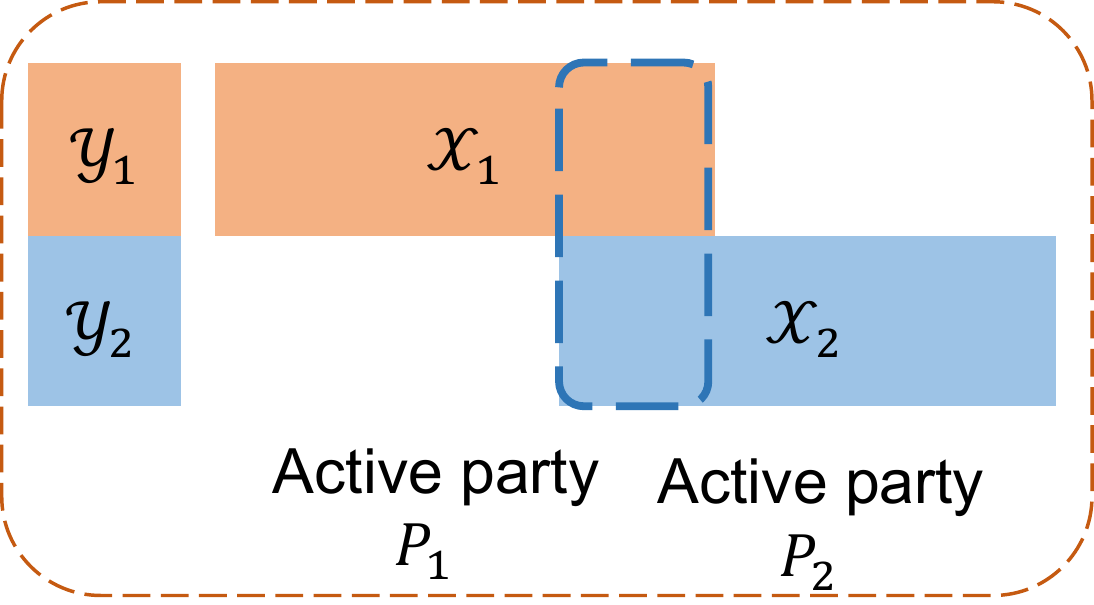}
\centering
\caption{Data distribution in feature-sharing Hetero-FTL}
\label{featureSharing}
\end{figure}

In feature-sharing Hetero-FTL, different parties have some features in common. However, the ID spaces are disjoint. Thus the data samples cannot be aligned across parties. We show the data distribution in feature-sharing Hetero-FTL in Figure~\ref{featureSharing}. 
To transfer knowledge to the active parties, the shared common features can be used as a pivot to bridge knowledge from different domains.

\begin{definition}\textbf{(Feature-sharing Hetero-FTL).}
Given $N$ parties and each party $P_i$ contains a set of features and labels of some instances, i.e., 
$\mathcal{D}_i = \{\bm{X}_i, \bm{I}_i, \bm{Y}_i \}$. $P_i$ has features $\bm{X}_i \in \mathcal{X}_i$, instances $\bm{I}_i \in \mathcal{I}_i$, and labels $\bm{Y}_i \in \mathcal{Y}_i$. 
There are $k$ active parties $\exists k \in P_{act}=\{k | k\in[1, N]\}\neq \emptyset$, 
where $P_{act}$ is a non-empty set that contains the index of active parties.
The data distribution is: 
\begin{align*}
	& \mathcal{X}_i \cap \mathcal{X}_j \neq \emptyset, \; \mathcal{I}_i \cap \mathcal{I}_j = \emptyset, \mathcal{Y}_k \neq \emptyset, \;\\
	& \forall  i \neq j \in [1, N], \; \exists k \in P_{act}=\{k | k\in[1, N]\}\neq \emptyset.
\end{align*}
Feature-sharing Hetero-FTL aims to train a model for each active party without revealing private data of each party:
\begin{align*}
	\argmin_{\bm{W}\in \bm{R}^{d\times N}} \sum_{i\in P_{act} } \lambda_{i} L \Bigl( \bm{Y}_i, f_{\bm{W}}\bigl( 
	& X_{\cap}, \bm{X}_i, z(\{\bm{X}_j\}_{j\in [1, N] \setminus \{i\}} ) \\
	& | \{\mathcal{D}_{j} \}_{j\in [1, N] \setminus \{i\}} 
 \bigr) \Bigr),
\end{align*}
where $X_{\cap}$ is the index of overlapped features, $\lambda_{i}$ is the weight for balancing the performance on each party, and $\bm{W} = [\bm{w}_1, ..., \bm{w}_N]$ contains local models $\{\bm{w}_i\}_{i\in [1, N]}$ corresponding to each party $\{P_{i}\}_{i\in [1,N]}$. 

\end{definition}

Just as instance-sharing Hetero-FTL is an extension of vertical FL by leveraging un-aligned private data for model training, 
feature-sharing Hetero-FTL can also be considered as an extension of HFL by taking advantage of private local features for model training. 
Before federated training, all parties first find co-existing features. 
During model training, each party locally trains a compact model with all local features as input. 
In FL, different parties aggregate the model parameters of the common features and compute an aggregated model. 
Therefore, the knowledge is transferred among parties via the aggregation of the parameters of common features. 

\begin{figure}[!htb] 
\subfigure[Data distribution]{
\label{featureSharing_demo}
\includegraphics[width=0.95\linewidth]{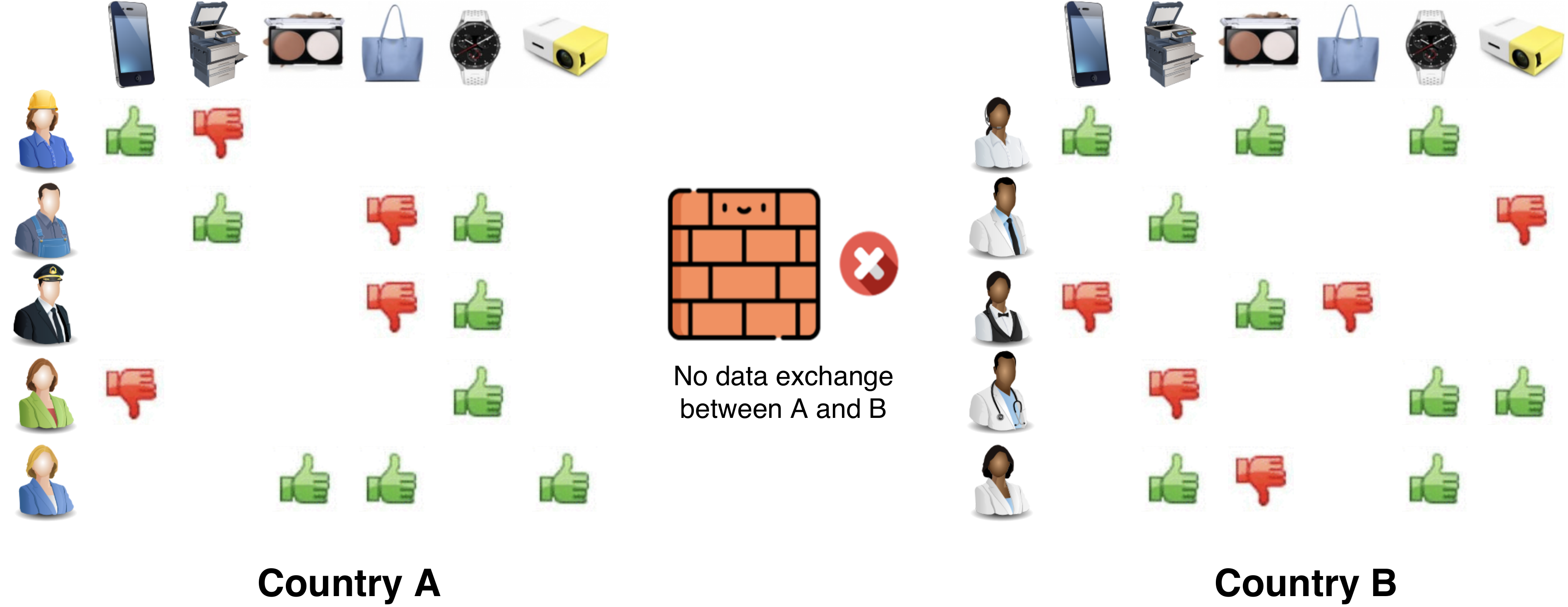} }
\subfigure[Feature space]{
\label{featureSharing_featurespace}
\includegraphics[width=0.95\linewidth]{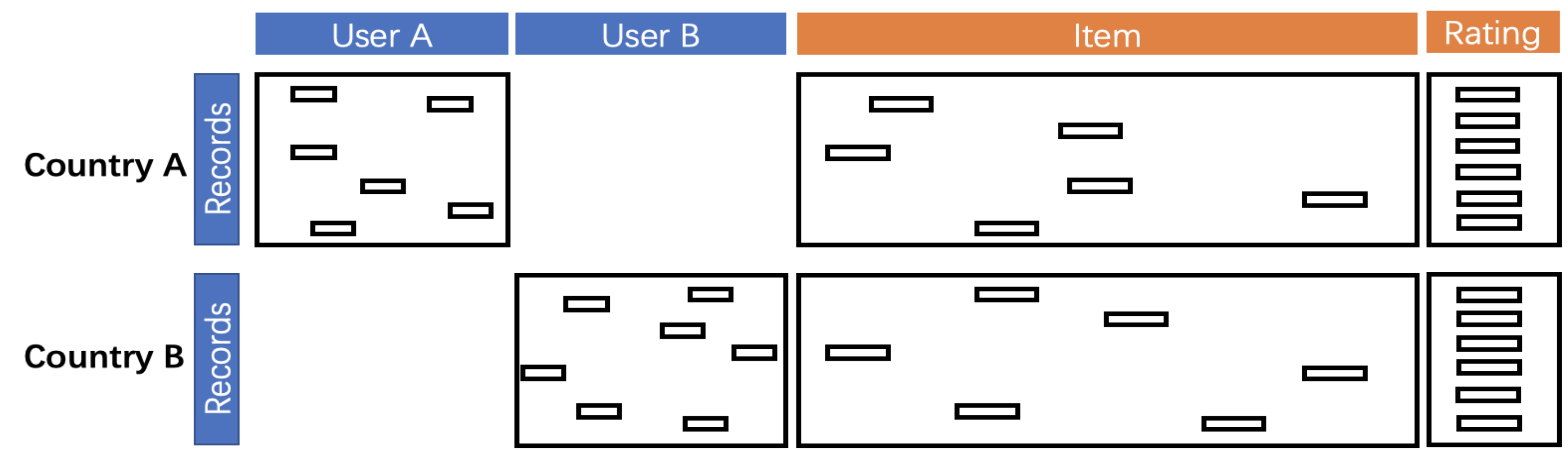} }
\centering
\caption{A typical scenario of feature-sharing Hetero-FTL. A transnational e-commerce company builds a federated recommender system with data from two countries. 
In this example, the company sells the same set of products (items) in two countries with different users. 
When using the matrix factorization algorithm for FL, the two parties share the same item features and different user features. }
\label{featureSharing_demos}
\end{figure}

\subsubsubsection{a) Application examples.}
The federated recommendation system is a typical application scenario of feature-sharing Hetero-FTL. 
As shown in Figure~\ref{featureSharing_demos}, a transnational e-commercial company aims to build a recommendation system with user-item interaction data collected from two countries. It is infeasible to exchange raw data among parties subject to legal constraints. 
The datasets from the two countries share the same set of products and different users, as shown in Figure~\ref{featureSharing_demo}. 
The feature space shown in Figure~\ref{featureSharing_featurespace} demonstrates that each data record contains a user feature and an item feature. 
Since the two parties share the same item features, the model weights of item features can be aggregated when training a federated recommendation system. 

Similarly, in another scenario, two recommenders may share the same set of users but own different items. 
For example, a movie recommender and a book recommender in the same country collaboratively build recommendation systems with common users. 
The user features are shared and thus can be used to transfer knowledge between recommenders. 
Some existing work~\cite{yang2020federated} categorize the first user-sharing federated recommendation as HFL and the second item-sharing federated recommendation as VFL. 
However, Gao et al.~\cite{Gao2020PrivacyTA} denote that both federated recommendation scenarios are the same in feature space and are typical feature-sharing Hetero-FTL problems. 

\subsubsubsection{b) Existing studies.}
There are mainly three types of transfer learning approaches used to tackle feature-sharing Hetero-FTL problems, including data augmentation~\cite{9005992}, parameter sharing~\cite{zhu2011heterogeneous}, and collaborative filtering~\cite{chai2019secure,ammad2019federated,chen2020practical,gao2020FedFM,chai2022federated}. 
Linear models and collaborative filtering approaches are widely explored to tackle the feature-sharing Hetero-FTL problem. 
Most practical algorithms are designed and implemented for federated recommendation, which is the major application scenario in the feature-sharing Hetero-FTL problem. However, deep models and three-based models are still under-explored in feature-sharing Hetero-FTL. 
We elaborate on these approaches in Section~\ref{sec-TL_HeteroFL}.

\subsubsection{Label-sharing Hetero-FTL}

\begin{figure}[!htb] 
\includegraphics[width=0.7\linewidth]{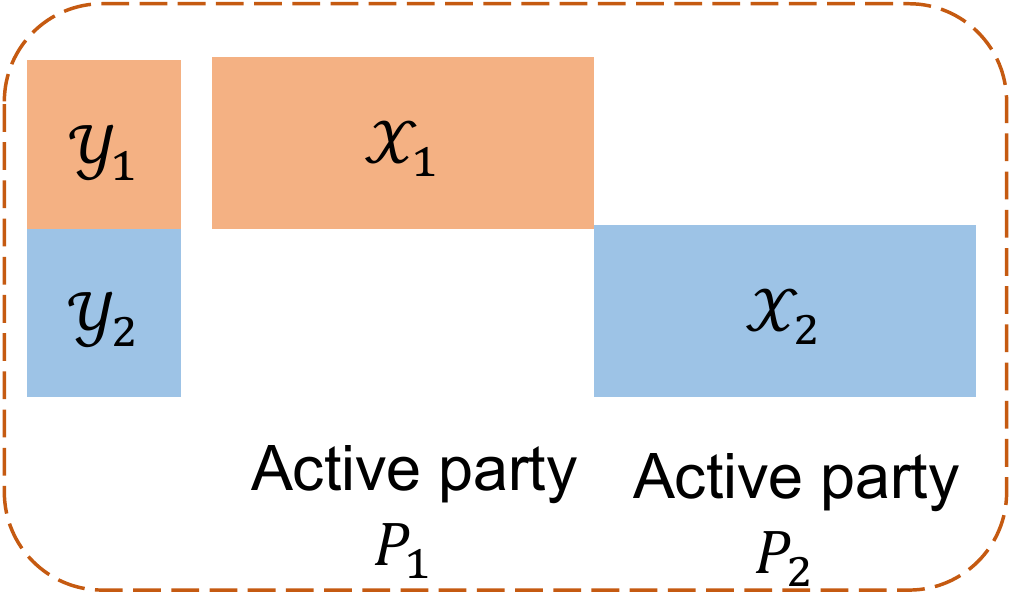}
\centering
\caption{Data distribution in label-sharing Hetero-FTL}
\label{labelSharing}
\end{figure}

Label-sharing Hetero-FTL considers a more challenging scenario in terms of data distribution, where all parties have (almost) neither co-occurring instances/IDs nor co-existing features, and they only share the same label space $\mathcal{Y}$. 
We demonstrate the data distribution in label-sharing Hetero-FTL in Figure~\ref{labelSharing}. To transfer knowledge between parties, the shared label space could be used as a pivot to bridge different domains during collaborative training. 

\begin{definition}\textbf{(Label-sharing Hetero-FTL).}
Given $N$ parties and each party $P_i$ contains a set of features and labels of some instances, i.e., 
$\mathcal{D}_i = \{\bm{X}_i, \bm{I}_i, \bm{Y}_i \}$. $P_i$ has features $\bm{X}_i \in \mathcal{X}_i$, instances $\bm{I}_i \in \mathcal{I}_i$, and labels $\bm{Y}_i \in \mathcal{Y}_i$. 
The data distribution of label-sharing Hetero-FTL is:
\begin{equation*}
	\mathcal{X}_i \cap \mathcal{X}_j = \emptyset, \; \mathcal{I}_i \cap \mathcal{I}_j = \emptyset, \mathcal{Y}_i = \mathcal{Y}_j, \; \forall  i \neq j \in [1, N],
\end{equation*}
Label-sharing Hetero-FTL aims to train a model for each active party without revealing private data of each party:
\begin{equation*}
	\argmin_{\bm{W}\in \bm{R}^{d\times N}} \sum_{i\in P_{act} } \lambda_{i} L \left( \bm{Y}_i, f_{\bm{W}}\left(\bm{X}_i | z(\{\mathcal{D}_{j} \}_{j\in [1, N] \setminus \{i\})} 
 \right) \right),
\end{equation*}
where $\lambda_{i}$ is the weight for balancing the performance on each party, and $\bm{W} = [\bm{w}_1, ..., \bm{w}_N]$ contains local models $\{\bm{w}_i\}_{i\in [1, N]}$ corresponding to each party $\{P_{i}\}_{i\in [1,N]}$. 

\end{definition}

\begin{figure*}[!htb] 
\includegraphics[width=13cm]{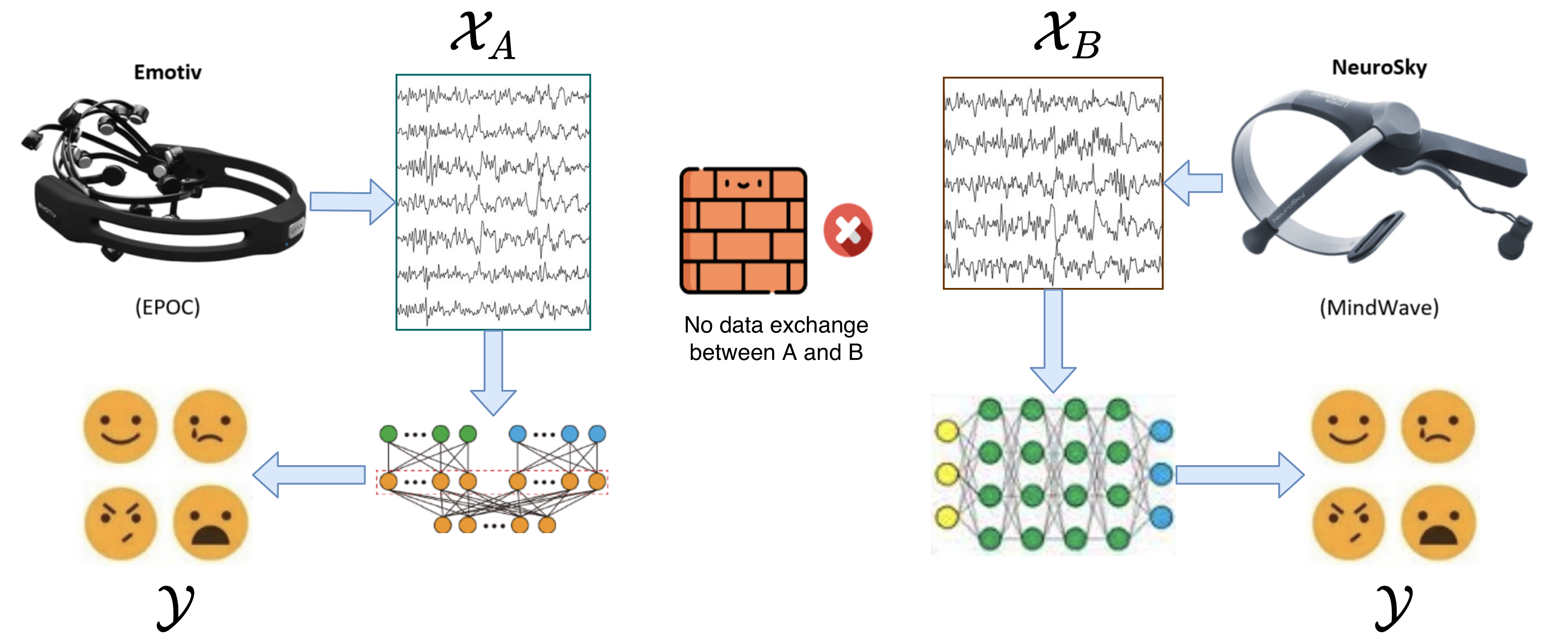} 
\centering
\caption{An example of label-sharing Hetero-FTL. Two EEG devices collect signals from different positions on head and aim to do the same emotion prediction task. }
\label{labelSharing_demo}
\end{figure*}

\subsubsubsection{a) Application examples.} Electroencephalographic (EEG) signals are collected from multiple electrodes on the head and are privacy-sensitive. As shown in Figure~\ref{labelSharing_demo}, two companies with different devices that collect EEG signals from different positions want to collaboratively train emotion prediction models without privacy violations. Their features are heterogeneous in terms of both channel number and electrode position, and there is no common user to align the data. Therefore, the two companies can leverage label-sharing Hetero-FTL to map their heterogeneous features into a common latent subspace according to label distribution; then, they can train a shared emotion prediction model.

\subsubsubsection{b) Existing studies.}
There are mainly two types of transfer learning approaches used to tackle the label-sharing Hetero-FTL issue: 1) parameter sharing, 2) knowledge distillation. We denote that these strategies can be combined to boost performance~\cite{liang2020think,he2020groutknowledgetransfer}. 
Although there are only a few works explicitly study label-sharing Hetero-FTL, some recent studies on \textit{system heterogeneity} in FL could be adapted to label-sharing Hetero-FTL problems, as they propose to train a heterogeneous local model architecture for each party~\cite{cao2022cofed,hin2021fedhe,liang2020think,he2020groutknowledgetransfer}. 
We elaborate these approaches in Section~\ref{sec-TL_HeteroFL}.

\subsubsection{Summary}
We summarize the utility-privacy-efficiency trade-off of the reviewed studies in data space heterogeneous federated transfer learning (Hetero-FTL) in Table~\ref{dataSpace_FL_summary}. We also list some related works proposed for Non-FL settings that can be modified to tackle data space heterogeneity. 
It can be observed that each aspect of Hetero-FTL is still under-explored. Most existing studies emphasize model performance (utility) and efficiency more than privacy, as they prefer to train plain-text neural networks without a rigorous privacy guarantee. Many of them were initially proposed to tackle system heterogeneity. 
Therefore, it is desirable to explore tailored Hetero-FTL algorithms with stringent privacy protection.

\begin{table*}[ht!]
\vspace{8mm}
\centering
\footnotesize
\begin{tabular}{c c c | c c c c c } 
 \toprule
   &  & & Utility & \multicolumn{1}{c}{Privacy}  & \multicolumn{3}{c}{Efficiency} \\
   \cmidrule{6-8} 
   \multirow{2}{*}{\begin{tabular}{@{}c@{}}Transfer \\ Method \end{tabular}} 
   & Setting
   & Existing Study 
   & \multirow{2}{*}{\begin{tabular}{@{}c@{}} model \\type \end{tabular}} 
   & \multirow{2}{*}{\begin{tabular}{@{}c@{}}privacy \\ mechanism\end{tabular}}
   & \multirow{2}{*}{\begin{tabular}{@{}c@{}}share\\labels \end{tabular}} 
   & \multirow{2}{*}{\begin{tabular}{@{}c@{}}asynchronous  \end{tabular}} 
   & \multirow{2}{*}{\begin{tabular}{@{}c@{}} aggregator   \end{tabular}} \\
   & & & & & & \\
 \midrule
 \multirow{2}{*}{\begin{tabular}{@{}c@{}}Data\\Augmentation\end{tabular}} 
 & FS & Gao et al.~\cite{9005992} & LM & HE,MPC & \texttimes & \texttimes & \texttimes \\
 \cmidrule{2-8} 
 & IS & Kang et al.~\cite{kang2022fedcvt} & NN & PT & \checkmark & \texttimes & \texttimes \\
  \midrule

 \multirow{9}{*}{\begin{tabular}{@{}c@{}}Parameter\\Sharing\end{tabular}} 
 &    & Liu et al.~\cite{YangLiu2019Secure} & NN & HE,MPC & \texttimes & \texttimes & \texttimes \\
 & IS & Sharma et al.~\cite{Sharma2019SecureAE} & NN & MPC & \texttimes & \texttimes & \texttimes \\
 &    & He et al.~\cite{he2022hybrid} & NN & PT & \texttimes & \texttimes & \checkmark \\
 &    & * Hou et al.~\cite{hou2021prediction} & LM & PT & - & - & - \\
 \cmidrule{2-8} 
 &    & Ju et al.~\cite{Ju2020FederatedTL} & NN & PT & \texttimes & \checkmark & \texttimes \\
 & LS & Gao et al.~\cite{Gao2019HHHFLHH} & NN & PT & \texttimes & \checkmark & \texttimes \\
 &    & Liang et al.~\cite{liang2020think} & NN & PT & \texttimes & \checkmark & \checkmark \\
 \cmidrule{2-8} 
 & \multirow{2}{*}{FS} & * Zhu et al.~\cite{zhu2011heterogeneous} & LM & PT & - & - & - \\
 &    & * Ye et al.~\cite{ye2018rectify} & LM & PT & - & - & - \\
 \midrule
 \multirow{7}{*}{\begin{tabular}{@{}c@{}}Knowledge\\Distillation\end{tabular}} 
 &    & Ren et al.~\cite{ren2022improving} & NN & PT & \texttimes & \texttimes & \checkmark \\
 &    & Wang et al.~\cite{wangvertical} & NN & PT & \texttimes & \texttimes & \checkmark \\
 &    & Li et a.~\cite{li2022semi} & NN & PT & \texttimes & \texttimes & \checkmark \\
 & IS & * Zhang et al.~\cite{zhang2018deep} & NN & PT & - & - & - \\
 &    & * Peng et al.~\cite{peng2021hierarchical} & NN & PT & - & - & - \\
 &    & * Tian et al.~\cite{tian2019contrastive} & NN & PT & - & - & - \\
 &    & * Guo et al.~\cite{guo2020online} & NN & PT & - & - & - \\
 \midrule
 \multirow{6}{*}{\begin{tabular}{@{}c@{}}Matrix\\Factorization\end{tabular}} 
 & IS,FS & Chai et al.~\cite{chai2022federated} & MF & RM,MPC & \checkmark & \texttimes & \checkmark \\
 \cmidrule{2-8} 
 &    & Chai et al.~\cite{chai2019secure} & MF & HE & \texttimes & \checkmark & \checkmark \\
 &    & Ammad et al.~\cite{ammad2019federated} & MF & PT & \texttimes & \checkmark & \texttimes \\
 & FS & Chen et al.~\cite{chen2020practical} & MF & MPC,DP & \checkmark & \checkmark & \texttimes \\
 &    & HegedHus et al.~\cite{hegedHus2019decentralized} & MF & PT  & \texttimes & \texttimes & \checkmark \\
 &    & Duriakova et al.~\cite{duriakova2019pdmfrec} & MF & PT  & \texttimes & \texttimes & \checkmark \\
 \bottomrule
\end{tabular}
\vspace{2mm}
\caption{Comparison of existing works for data space heterogeneous FL. 
\textbf{IS} denotes Instance-Sharing. \textbf{FS} denotes Feature-Sharing. \textbf{LS} denotes Label-Sharing. 
* denotes the approach is proposed for non-FL setting and can be modified to tackle this issue. 
LM denotes Linear Models.  NN denotes Neural Networks. 
HE denotes Homomorphic Encryption. MPC denotes Secure Multiparty Computation. DP denotes Differential Privacy. RM denotes Random Matrix. PT denotes Plain-text. - denotes not applicable. 
}
\vspace{-3mm}
\label{dataSpace_FL_summary}
\end{table*}

Table~\ref{data_hetero_ftl_transfer_summary} summarizes the transfer learning strategies adopted by Hetero-FTL researches. 
We identify four transfer learning strategies to address critical Hetero-FTL challenges, including data augmentation, parameter sharing, knowledge distillation, and matrix factorization. 
We find representation learning and knowledge distillation are popular strategies to transfer knowledge among parties and are widely used in all types of Hetero-FTL problems. 
For neural network models, each party builds a local model to map their feature space into a common latent subspace. 
For linear models, learning a linear mapping can map the feature representation between parties or to a common subspace via concept factorization. 
Knowledge distillation is still under-explored, especially in feature-sharing Hetero-FTL. Some of the listed KD approaches were proposed for system-heterogeneous FL and are not tailored to Hetero-FTL. Therefore, it is desirable to develop KD-based approaches to tackle Hetero-FTL problems. 
Collaborative filtering approaches such as matrix factorization and factorization machines are also explored for federated recommendation applications, as the federated recommendation is a feature-sharing Hetero-FTL problem by nature. 
We will introduce the studies using these transfer learning approaches to tackle data space heterogeneity in Section~\ref{sec-TL_HeteroFL}.

\begin{table*}[ht]

\begin{frame}
\centering
\footnotesize
\begin{tikzpicture}
\node (table) {
\begin{tabular}{c c | c  c c } 
 \toprule
  Perspective & Transfer method & Instance-sharing & Feature-sharing & Label-sharing \\
   \midrule
   Data-based &  Data Augmentation &  Kang et al.~\cite{kang2022fedcvt} & Gao et al.~\cite{9005992} & - \\
   \midrule
  \multirow{5}{*}{Architecture-based} & \multirow{3}{*}{\begin{tabular}{@{}c@{}}Parameter \\ Sharing\end{tabular}} 
    & \multirow{3}{*}{\begin{tabular}{@{}c@{}} Liu et al.~\cite{YangLiu2019Secure} \\ Sharma et al.~\cite{Sharma2019SecureAE} \end{tabular}}        
    & \multirow{3}{*}{\begin{tabular}{@{}c@{}} - \end{tabular}}   
    & \multirow{3}{*}{\begin{tabular}{@{}c@{}} Liang et al.~\cite{liang2020think} \\  Gao et al.~\cite{Gao2019HHHFLHH} \\ Ju et al.~\cite{Ju2020FederatedTL} \end{tabular}}  \\
  & & & &   \\
  & & & &  \\
   \cmidrule{2-5} 
  & \multirow{3}{*}{\begin{tabular}{@{}c@{}}Knowledge \\ Distillation\end{tabular}} 
  & \multirow{3}{*}{\begin{tabular}{@{}c@{}} Li et al.~\cite{li2022semi} \\ Wang et al.~\cite{wangvertical} \\ Ren et al.~\cite{ren2022improving} \end{tabular}}
   & \multirow{3}{*}{\begin{tabular}{@{}c@{}} - \end{tabular}} & \multirow{3}{*}{\begin{tabular}{@{}c@{}} - \end{tabular}} \\
  &  &  & &  \\
  &  &  & &  \\
  \midrule
  \multirow{5}{*}{Model-based} 
  & \multirow{5}{*}{\begin{tabular}{@{}c@{}}Matrix \\ Factorization\end{tabular}} 
  & \multirow{5}{*}{\begin{tabular}{@{}c@{}}Chai et al.~\cite{chai2022federated}\end{tabular}} 
  &  \multirow{5}{*}{\begin{tabular}{@{}c@{}}Chai et al.~\cite{chai2019secure}\\Ammad et al.~\cite{ammad2019federated}\\ Chen et al.~\cite{chen2020practical} \\ HegedHus et al.~\cite{hegedHus2019decentralized} \\ Duriakova et al.~\cite{duriakova2019pdmfrec} \end{tabular}}  
  & \multirow{5}{*}{\begin{tabular}{@{}c@{}}-\end{tabular}} \\
  &  & &  & \\
  &  & &  & \\
  &  & &  & \\
  &  & &  & \\

 \bottomrule
\end{tabular}
};
\draw [red, thick,rounded corners]
  ($(-1.3,1.96)$)
  rectangle 
  ($(6.9, 1.58)$);

\draw [red, thick,rounded corners]
  ($(2,0.5)$)
  rectangle 
  ($(4,1.4)$);

\draw [red, thick,rounded corners]
  ($(2,0.3)$)
  rectangle 
  ($(4,-0.6)$);

\draw [red, thick,rounded corners]
  ($(4.9,0.3)$)
  rectangle 
  ($(6.9,-0.6)$);

\draw [red, thick,rounded corners]
  ($(4.9,-1.1)$)
  rectangle 
  ($(6.9,-2.1)$);

\end{tikzpicture}
\end{frame}

\caption{Summary of transfer learning strategies adopted by Hetero-FTL research. Red boxes denote under-explored promising directions.}
\label{data_hetero_ftl_transfer_summary}
\end{table*}

\section{Statistical Heterogeneity in HFL}\label{sec-statistical_FL}

In this section, we investigate statistical heterogeneity in data space homogeneous (a.k.a. \textit{horizontal}) FL, where all parties have data from the same feature space. 
Data collected from the same data space can be imbalanced and non-IID (independent and identically) distributed in the FL system. 
For example, autonomous vehicles run in rural and urban environments. Such variance leads to the marginal distributions $P(\mathcal{X})$ of collected data differ. Moreover, different drivers in the same environments show various driving style preferences, leading to varying conditional distributions $P(\mathcal{Y}|\mathcal{X})$ of human driving behaviors. In addition, class imbalance and data size imbalance across parties may also vary significantly. 

The non-IID data among parties leads to poor global model convergence and a lack of solution personalization. These issues deteriorate the performance of the global model on each participant and could also disincentives the participants from participating in the FL. 
To tackle the non-IID issue, transfer learning approaches have attracted tremendous research interest in recent years. 
There are mainly two strategies to tackle statistical heterogeneity in FL. 1) Train a global model with good generalizability, then each party fine-tunes the pre-trained global model on its local dataset. 2) Directly train personalized models for all parties.

In statistical heterogeneous FL, the data spaces $\mathcal{D}_i$ in each party $P_i$ are the same in terms of both feature space $\mathcal{X}$ and label space $\mathcal{Y}$. However, the marginal distribution or conditional distribution may differ among different parties.
Since it was proposed, FL is meant to study the case where multiple clients holding homogeneous data aim to train a single global model collaboratively under specific privacy constraints. Therefore, most existing researches focus on data space homogeneous FL.
Research on data space homogeneous FL can be categorized into two cases: horizontal FL (HFL) and homogeneous federated transfer learning (Homo-FTL). 
The major difference is that, in HFL, there is one learning objective, and parties collaboratively train a global model on all data. HFL tackles the statistical heterogeneity by training \textit{a single global model}. 
However, in Homo-FTL, each party optimizes its objective function and builds its model, and the statistical heterogeneity is tackled by training \textit{personalized models} for all parties. 

We discuss the two cases in the following. 
Interested readers may refer to~\cite{Kairouz2019Survey105,9084352} for more details on homogeneous FL.

\subsection{Single Global Model}
HFL handles multi-client data that share the same set of features but are different in samples and is also referred to as \textit{federated optimization}. 
The pioneering works~\cite{McMahan2016ModelAvg,HBrendanMcMahan2016,DBLP:journals/corr/KonecnyMRR16} in FL mainly focus on this setting, where participants collaboratively train a global model based on their data with the help of a coordinator for model aggregation. 
The goal of HFL is to train a high-quality centralized global model. The objective can be formulated as:
\begin{equation*}
\min_{w\in \mathbb{R}^d} \left\{ f(w) := \frac{1}{n}\sum_{i=1}^{N}f_i(w) \right\} ,
\end{equation*}
where $w$ is the parameters of the federated model, and $f_i : \mathbb{R}^d \to \mathbb{R}$ is the training objective function that maps the model parameter set $w \in \mathbb{R}^d$ to a real valued training loss with respect to the training data $\mathcal{D}_i$ of party $P_i$.

To improve communication efficiency, Reisizadeh et al. ~\cite{pmlr-v108-reisizadeh20a} propose to use periodic averaging and quantization to reduce data transmission. 
Wang et al. ~\cite{luping2019cmfl} propose CMFL to preclude the irrelevant updated by clients. 
Konecny et al. ~\cite{Jakub2016CommEff} propose structured updates and sketched updates, which learn updates from a restricted space and compress the learned updates, respectively. 
For privacy preservation, differential privacy has been intensively studied in FL~\cite{liu2021flame,jiahao2021dpcecl,DBLP:journals/corr/abs-1712-07557} to protect client-side privacy. 
Homomorphic encryption and secure multiparty computation are also adopted for secure model training and aggregation~\cite{zhang2020batchcrypt,xu2019verifynet,truex2019hybrid}. 
Moreover, data in HFL can be 1) non-IID, 2) unbalanced, and 3) massively distributed, which eliminates the model performance. The non-IID issue is common as each client corresponds to a particular user, a location, or a time window. For example, in the virtual keyboard application, users who live in the UK and the USA speak English differently; thus, the conditional probability varies.  
To tackle the non-IID problem, many approaches have been proposed to mitigate the clients~\cite{Nagalapatti2021GoG}. 
For example, Jeong et al.~\cite{jeong2018communication} proposed federated distillation and federated augmentation for communication efficient FL on non-IID data. A generative model is trained to augment each party's local data towards yielding an IID dataset. 
Lin et. al~\cite{lin2020ensemble} introduced ensemble distillation for robust model aggregation on non-IID data. 

We elaborate different approaches to train a singel global model on Non-IID data in Secion~\ref{sec-TL_HeteroFL}.

\subsection{Personalized Local Models}
Different from HFL, which learns a single model for all parties even if the data is non-IID across different parties, Homo-FTL aims to transfer knowledge from the other parties to some or each party. 
Homo-FTL optimizes multiple objectives to achieve the best performance for each client with non-IID data. Therefore, Homo-FTL is highly related to multi-task learning, personalization, and meta-learning. A variety of multi-model approaches are proposed to tackle the non-IID data problem. 
We formulate the Homo-FTL as:
\begin{equation*} 
\min_{W \in \mathbb{R}^{d \times N} }  \left\{ \mathcal{G}(W) := \sum_{i=1}^{N} f_i(w_i) \right\},
\end{equation*} 
where $W = [w_1, ..., w_N]$ is a $d \times N$-dimensional matrix that collects $w_1, ..., w_N$ as its columns. 

Recently, personalized FL has attracted great attention. Fallah et al. ~\cite{NEURIPS2020_24389bfe} propose a model-agnostic meta-learning approach to train an initial shared model for each party to fine-tune with its local data. 
Huang et al. ~\cite{huang2021pcsfl} employ federated attentive message passing to facilitate similar clients to collaborate more and enable each client to own a local, personalized model. 
In some cases, different groups of parties have their own objectives, and federated clustering approaches can estimate the cluster identities of the parties and train models for the party clusters~\cite{ghosh2020efficient,donahue2021modelsharing}. 

We elaborate different approaches to train personalized local models on Non-IID data in Section~\ref{sec-TL_HeteroFL}.

\subsection{Summary}
As shown in Table~\ref{statisitical_FL_summary}, we summarize the utility-privacy-efficiency trade-off factors of the reviewed works on system heterogeneous federated transfer learning. It can be observed that existing studies highly focus on improving model performance by training plain-text neural networks and fails to protect data privacy by leveraging privacy-preserving techniques.

\begin{table*}[ht!]
\centering
\footnotesize
\begin{tabular}{ c c | c c c c c c } 
 \toprule
 \multirow{3}{*}{\begin{tabular}{@{}c@{}} Objective \end{tabular}} 
 & \multirow{3}{*}{\begin{tabular}{@{}c@{}} Transfer \\ Method \end{tabular}}
 & & Utility & \multicolumn{1}{c}{Privacy}  & \multicolumn{2}{c}{Efficiency} \\
 \cmidrule{6-7} 
 &  & Existing Study & \multirow{2}{*}{\begin{tabular}{@{}c@{}} model \\type \end{tabular}} & \multirow{2}{*}{\begin{tabular}{@{}c@{}}privacy \\ mechanism\end{tabular}} & \multirow{2}{*}{\begin{tabular}{@{}c@{}}asynchronous \end{tabular}} &  \multirow{2}{*}{\begin{tabular}{@{}c@{}} aggregator \end{tabular}} \\
 & & & & & \\
 \midrule
  \multirow{11}{*}{\begin{tabular}{@{}c@{}} Single \\ global \\ model \end{tabular}}
  & \multirow{4}{*}{\begin{tabular}{@{}c@{}} Data\\Augmentation \end{tabular}} 
 	& Zhao et al.~\cite{zhao2018fednoniiddata} & NN & PT & \texttimes & \checkmark \\
	& & Fu et al.~\cite{wu2020fedhome} & NN & PT & \texttimes & \checkmark \\
	& & Duan et al.~\cite{duan2020self} & NN & PT & \checkmark & \checkmark \\
	& & Jeong et al.~\cite{jeong2018communication} & NN & PT & \texttimes & \checkmark \\
 \cmidrule{2-7}
  &  \multirow{7}{*}{\begin{tabular}{@{}c@{}} Domain\\Adaptation \end{tabular}} 
	& Peng et al.~\cite{peng2019federated} & NN & PT  & \texttimes & \checkmark \\
	& & Sun et al.~\cite{sun2016return} & NN & PT  & \texttimes & \checkmark \\
	& & Hanzely et al.~\cite{hanzely2020federated} & NN & PT  & \texttimes & \checkmark \\
	& & Deng et al.~\cite{deng2020adaptive} & NN & PT  & \checkmark & \checkmark \\
	& & Smith et al.~\cite{Smith2017MT} & NN & PT  & \texttimes & \checkmark \\
	& & Huang et al.~\cite{huang2021pcsfl} & NN & PT  & \texttimes & \checkmark \\
	& & Shoham et al.~\cite{shoham2019overcoming} & NN & PT  & \checkmark & \checkmark \\
 \midrule
  \multirow{17}{*}{\begin{tabular}{@{}c@{}} Personalized \\ local \\ models \end{tabular}}
  & \multirow{4}{*}{\begin{tabular}{@{}c@{}} Parameter\\Sharing \end{tabular}} 
	& Singhal et al.~\cite{singhal2021federated} & NN & PT & \texttimes & \checkmark \\
	& & Bui et al.~\cite{bui2019federated} & NN & PT & \texttimes & \checkmark \\
	& & Liang et al.~\cite{liang2020think} & NN & PT & \checkmark & \checkmark \\
	& & Arivazhagan et al.~\cite{arivazhagan2019federated} & NN & PT & \texttimes & \checkmark \\
 \cmidrule{2-7}
  &  \multirow{11}{*}{\begin{tabular}{@{}c@{}} Knowledge\\Distillation \end{tabular}} 
	& Li et al.~\cite{li2019fedmd} & NN & PT & \texttimes & \texttimes \\
	& & Chang et al.~\cite{chang2019cronus} & NN & PT & \texttimes & \checkmark \\
	& & Zhang et al.~\cite{zhang2021parameterized} & NN & PT & \texttimes & \checkmark \\
	& & Gong et al.~\cite{gong2022preserving} & NN & PT & \checkmark & \texttimes \\
	& & Li et al.~\cite{li2021decentralized} & NN & PT & \texttimes & \checkmark \\
	& & Yang et al.~\cite{yang2021mutualpfl} & NN & PT & \checkmark & \checkmark \\
	& & Matsuda et al.~\cite{matsuda2022fedme} & NN & PT & \checkmark & \checkmark \\
	& & Li et al.~\cite{li2021fedh2l} & NN & PT & \texttimes & \checkmark \\
	& & Zhu et al.~\cite{zhu2018knowledge} & NN & PT  & \texttimes & \checkmark \\
	& & Afonin et al.~\cite{afonin2021towards} & NN & PT & \texttimes & \checkmark \\
	& & Zhu et al.~\cite{pmlr-v139-zhu21b} & NN & PT  & \texttimes & \checkmark \\
 \cmidrule{2-7}
  &  \multirow{2}{*}{\begin{tabular}{@{}c@{}} Matrix\\Factorization \end{tabular}} 
	& \multirow{2}{*}{\begin{tabular}{@{}c@{}} Chai et al.~\cite{chai2022federated} \end{tabular}}  & \multirow{2}{*}{\begin{tabular}{@{}c@{}} MF \end{tabular}} & \multirow{2}{*}{\begin{tabular}{@{}c@{}} RM  \end{tabular}} & \multirow{2}{*}{\begin{tabular}{@{}c@{}} \texttimes \end{tabular}} & \multirow{2}{*}{\begin{tabular}{@{}c@{}} \checkmark \end{tabular}}  \\

	 & & & &  & \\

 \bottomrule
\end{tabular}
\vspace{2mm}
\caption{Comparison of existing works for statistical heterogeneous FL. 
NN denotes Neural Networks. MF denotes matrix factorization. 
RM denotes Random Matrix. PT denotes Plain-text. }
\vspace{-3mm}
\label{statisitical_FL_summary}
\end{table*}

\section{System Heterogeneity}\label{sec-system_FL}

In FL systems, heterogeneity that happens at the system level can also exert a significant influence on the FL training process. 
Devices may differ in hardware (CPU, GPU, memory), network connectivity (WiFi, 4G, 5G), and power supply (battery mode), which leads to various computation, communication, and storage capabilities. 
Such heterogeneity of resource-constrained devices brings multiple challenges to FL~\cite{xu2021asynchronous}: 
\begin{itemize}
	\item Unreliable network connectivity. Network connectivity widely varies from one client to another. Most often, clients are under slow, limited, expensive, and unavailable connections, which significantly reduce the number of available ones at once. In addition, among the available clients, many might not be able to participate in each learning round due to their different computation capabilities.
	\item Low round efficiency. The faster devices have to wait for those straggler devices in each communication round for aggregation. 
	\item Low resource utilization. The competent devices may not be utilized thoroughly by the classical FL algorithms. 
\end{itemize}

To tackle the above challenges, there are mainly two types of strategies in system heterogeneous FL: 
\begin{itemize}
\item One is to train a global model with the best model performance under the various resource constraints. Many works formulate the problem as resource allocation problems or multi-objective optimization problems~\cite{luo2021cost,kim2021autofl}. 
Many studies have been focusing on resource allocation to make the optimal decisions about the chosen clients, learning hyper-parameters, quantity and length of communication rounds, and aggregation strategies.
\item The other is to train a tailored local model for each party with highest local model performance while optimizing the resource consumption. Many contributions has been made to train heterogeneous model architectures~\cite{diao2021heterofl} or quantization~\cite{ozkara2021quped} for different edge devices.
\end{itemize}

In this section, we investigate the approaches to mitigate system heterogeneity, including client selection, adaptive aggregation, joint optimization, and model compression. 


\subsection{Single Global Model}

\subsubsection{Client Selection}

Instead of allowing all clients to contribute equally in federated training, many client selection algorithms are proposed to select robust and powerful clients to improve the training efficiency.

FedCS is proposed to mitigate the device heterogeneity~\cite{nishio2018client}. In FedCS, a server first gather information from a set of randomly selected clients for network states and computing resources. Then, the server selects the maximum possible number of clients to update models within a prespecified time interval. 
To select the maximum possible number of clients, a greedy algorithm is proposed, where the clients taking the least time for local training are selected interatively. 
Yoshida et al.~\cite{yoshida2019hybrid} further propose Hybrid-FL to handle non-IID data distributions among clients. During the client selection phase, the devices are selected to form an approximately IID dataset. 
Yang et al.~\cite{TimothyYang2018} apply FL to Google keyboard query suggestions where smartphones are only available for training when they are in charging mode, connected to WiFi, and screened off. 
Chen et al.~\cite{chen2021towards} propose a heuristic greedy node selection strategy that iteratively selects heterogeneous IoT nodes to participate
in global learning aggregation based on their local computing and communication resources.
Zhu et al.~\cite{zhou2021tea} limit the number of devices training at the same time in the AFL network. A limit-size cache with a weighted averaging mechanism is introduced onto the server to reduce the impact of model staleness.
Hao et al.~\cite{hao2020time} designed a prioritized node selecting function according to the computing power and accuracy change of local models on each node. Other unselected nodes continue the iterations locally at the same time. 
Similarly, Imteaj et al.~\cite{imteaj2020fedar} proposed to use a trust score to each node based on its activities.

\subsubsection{Asynchronous Aggregation}

The efficiency of classical FL algorithms is susceptible to the straggling devices, as the server waits for all devices to complete local update before aggregation. Therefore, asynchronization is highlighted by many recent studies to boost the scalability and efficiency. 
In asynchronous FL, a client can update the global model whenever it sends a local update to the server. 

Sprague et al.~\cite{sprague2018asynchronous} point out that training FedAvg in an asynchronous manner is robust to clients joining halfway during training. However, when the training data is non-IID among devices, the convergence rate of the asynchronous approach is slower than FedAvg. 
The FedAsync algorithm is proposed to improve convergence~\cite{Xie2019Async}. FedAsync adaptively weight newly received client updates according to staleness. An outdated update from a straggler is weighted less, as it should be received in previous rounds. 

The asynchronous FL algorithms can be harmed by the straggler devices, which update low quality models. Semi-asynchronous algorithms are proposed~\cite{xu2021asynchronous}. Semi-asynchronous approaches combines synchronous and asynchronous FL. 
Hao et al.~\cite{hao2020time} a priority function is introduced to accurately select nodes with
large amounts of data or high computation power. Meanwhile, local models
on unselected nodes will be cached for a specific number of iterations before
being submitted to the aggregation server. Besides, a restriction on the local
training round number is set to prevent specific nodes from being unselected
for a lengthy period of time, leading the global model to overfit certain nodes.
The effectiveness of the scheme is evidenced by experiments conducted on
iid and non-iid datasets.

Wu et al.~\cite{Wu2021safa} propose SAFA, a semi-asynchronous FL protocol, to address the problems in FL such as low round efficiency and poor convergence rate in extreme conditions.
To boost efficiency and improve the quality of the global model, SAFA designs a cache-based lag-tolerant mechanism for model distribution, client selection and global aggregation. 
SAFA classifies clients into three types: up-to-date, deprecated, and tolerable. The first two types of clients update models in a synchronous manner, and the last type of client update models in an asynchronous manner. 
Thereby, SAFA achieves a tradeoff between faster convergence and lower communication cost.


\subsubsection{Model Compression}
Model compression techniques are explored to reduce the communication cost. 
Low-rank matrix factorization with heterogeneous model sizes is also explored to train heterogeneous models for devices with different hardware resources. 
FedHM~\cite{yao2021fedhm} is a heterogeneous federated model compression framework that distributes the heterogeneous low-rank models to clients and then aggregating them into a full-rank model. In FedHM, each party low-rank factorize the parameter matrix of the neural network with different sizes, according to its hardware resources. 
In each communication round, the server receives the factorized models and transforms the factorized low-rank models back to full-rank models. Then, the server conducts aggregation using weighted averaging.

Caldas et al.~\cite{Caldas2018exband} propose to minimize the communication cost by reducing the size of the models transmitted between the server and the clients. First, the server generates a smaller model with fewer parameters via federated dropout technique. 
Then, the server performs lossy compression on the resulting model and sends to the clients. 
Then, the clients decompress and train the local models.
Finally, the local models are lossily compressed and sent to the server, for decompression and aggregation.

\subsection{Personalized Local Models}

\subsubsection{Parameter Sharing}

HeteroFL~\cite{diao2021heterofl} proposes to train heterogeneous neural networks with various width by averaging the overlapped sub-matrices of all parameter matrices of all parameter matrices across various models. 
The number of parameters in smaller models are reduced in width by cropping each parameter matrix of the largest model into sub-matrix with various reduction radio.
With this construction, HeteroFL adaptively allocates subsets of global model parameters to each party according to its corresponding capabilities. 
To perform aggregation, each parameter is averaged from the parties whose local models contain that parameter. 
However, this approach is not beneficial for retaining large models' knowledge in smaller models because the width pruning operation breaks the original model structure. 
Moreover, even the same sub-matrix may have different behaviors in the small and large models due to the model structure differences, which may lead to a sub-optimal performance due to the mismatch of feature spaces. 
In addition, simply sharing parts of parameters across different models cannot effectively transfer useful knowledge encoded by strong models to other weaker models. 
To tackle the above issue, Liu et al.~\cite{Liu2022heterofl} further propose InclusiveFL, in which the shallow bottom model layers in the largest model are shared with other smaller local models, to eliminate the mismatch between small and large local models. 

Split-Mix~\cite{hong2022efficient} proposes to train multiple models of different sizes and adversarial robustness levels tailored to the budget of each device, to address the under-training issue in large models. 
During training, strong parties sample more base models for local training and weak parties sample less base models. In each communication round, the server aggregates each base model using the updates from the corresponding parties. 
In inference phase, the trained local models can be aggregated on-demand according to inference requirements. 
Munir et al.~\cite{munir2021fedprune} proposed FedPrune to achieve inclusive FL. It prunes the global model for weak parties based
on their capabilities.

FedMask~\cite{li2021fedmask} is proposed to achieve communication and computation efficient FL. 
In FedMask, each party learns a heterogeneous and structured sparse neural network. To achieve this, each party learns a sparse binary mask while freezing the parameters of local model. Only the binary masks are communicated between devices and the server. 
The model aggregation is conducted by averaging on the unmasked parameters for each local model. 
The authors of FedMask also propose a similar approach named Hermes~\cite{li2021hermes} that trains personalized and sparse neural networks by applying the structured pruning. Only the updates of these sub-networks are transmitted. To aggregate models, the server performs the average on only overlapped parameters across each sub-network. To transfer knowledge among heterogeneous models, the lower layers are retained and aggregated and the higher layers are pruned by channel. 
Lu et al.~\cite{lu2021heterogeneous} propose to use model mapping to fuse heterogeneous models in FL.

\subsubsection{Clustered Training}

Hierarchical knowledge transfer are widely explored in heterogeneous networks where the devices are group according to location~\cite{abad2020hierarchical}, communication delay~\cite{chai2020tifl}, or capacity~\cite{Liu2022heterofl}. Therefore, the knowledge is first transferred inside a group via homogeneous FL, then the group-wise models are aggregated for higher-level knowledge transfer. 
Abad et al.~\cite{abad2020hierarchical} studied a heterogeneous cellular network where the mobile devices are grouped by small base stations in different cells. The mobile devices are trained in a hierarchical way. Specifically, the devices in each cell first conduct FL by base stations. Then, the models trained within each cell are periodically aggregated via a macro base station to mutually transfer knowledge among the base stations. 
Similarly, Chai et al.~\cite{chai2020tifl} proposed a tire-based client selection approach that groups clients according to communication delay. The devices with similar training speed collaboratively conduct FL to train a model. The model trained by each group are then aggregated by a server. Considering the devices in internet of things (IoT) networks are composed of resource-limited devices and powerful devices, these approaches can be adopted to IoT applications.

\subsection{Summary}
We summarize the trade-off factors of the existing works on system heterogeneous FL in Table~\ref{system_FL_summary}. 
It can be observed that most existing studies in system heterogeneous FL focus on improving model performance by training plain-text neural networks and merely leverage privacy-preserving techniques such as HE and DP for privacy protection.

\begin{table*}[ht!]
\centering
\footnotesize
\begin{tabular}{ c c | c c c c c c } 
 \toprule
 \multirow{3}{*}{\begin{tabular}{@{}c@{}} Objective \end{tabular}} 
 & \multirow{3}{*}{\begin{tabular}{@{}c@{}} Transfer \\ Method \end{tabular}} 
 & & Utility & \multicolumn{1}{c}{Privacy}  & \multicolumn{2}{c}{Efficiency} \\
 \cmidrule{6-7} 
 &  & Existing Study & \multirow{2}{*}{\begin{tabular}{@{}c@{}} model \\type \end{tabular}} & \multirow{2}{*}{\begin{tabular}{@{}c@{}}privacy \\ mechanism\end{tabular}} & \multirow{2}{*}{\begin{tabular}{@{}c@{}}asynchronous \end{tabular}} &  \multirow{2}{*}{\begin{tabular}{@{}c@{}} aggregator \end{tabular}} \\
 & & & & & \\
 \midrule
  \multirow{14}{*}{\begin{tabular}{@{}c@{}} Single \\ global \\ model \end{tabular}}
  & \multirow{6}{*}{\begin{tabular}{@{}c@{}} Client\\Selection \end{tabular}} 
  & Nishio et al.~\cite{nishio2018client} & NN & PT  & \texttimes & \checkmark \\
	& & Yoshida et al.~\cite{yoshida2019hybrid} & NN & PT  & \checkmark & \checkmark \\
	& & Yang et al.~\cite{TimothyYang2018} & NN & PT  & \checkmark & \checkmark \\
	& & Chen et al.~\cite{chen2021towards} & NN & PT  & \texttimes & \checkmark \\
	& & Zhu et al.~\cite{zhou2021tea} & NN & PT  & \checkmark & \checkmark \\
	& & Similarly, Imteaj et al.~\cite{imteaj2020fedar} & NN & PT  & \checkmark & \checkmark \\
 \cmidrule{2-7} 
  &  \multirow{4}{*}{\begin{tabular}{@{}c@{}} Asynchronous \\ Aggregation \end{tabular}} 
 & Sprague et ap.~\cite{sprague2018asynchronous} & NN & PT  & \checkmark & \checkmark \\
 & & Xu et al.~\cite{xu2021asynchronous} & NN & PT  & \checkmark & \checkmark \\
 & & Hao et al.~\cite{hao2020time} & NN & PT  & \checkmark & \checkmark \\
 & & Wu et al.~\cite{Wu2021safa} & NN & PT  & \checkmark & \checkmark \\
 \cmidrule{2-7} 
 & \multirow{2}{*}{\begin{tabular}{@{}c@{}} Model \\Compression \end{tabular}} 
 & Yao et al.~\cite{yao2021fedhm} & NN & PT  & \texttimes & \checkmark \\
 & & Caldas et al.~\cite{Caldas2018exband} & NN & PT  & \texttimes & \checkmark \\
 \cmidrule{2-7} 
  &  \multirow{2}{*}{\begin{tabular}{@{}c@{}} Domain\\Adaptation \end{tabular}} 
  & Smith et al.~\cite{Smith2017MT} & NN & PT  & \texttimes & \checkmark \\
	& & Marfoq et al.~\cite{marfoq2021federated} & NN & PT & \texttimes & \checkmark \\
 \midrule
  \multirow{14}{*}{\begin{tabular}{@{}c@{}} Personalized \\ local \\ models \end{tabular}}
  & \multirow{7}{*}{\begin{tabular}{@{}c@{}} Parameter\\Sharing \end{tabular}} 
	& Diao et al.~\cite{diao2021heterofl} & NN & PT  & \texttimes & \checkmark \\
	& & Liu et al.~\cite{Liu2022heterofl} & NN & PT  & \texttimes & \checkmark \\
	& & Hong et al.~\cite{hong2022efficient} & NN & PT  & \checkmark & \checkmark \\
	& & Munir et al.~\cite{munir2021fedprune} & NN & PT  & \texttimes & \checkmark \\
	& & Li et al.~\cite{li2021fedmask} & NN & PT  & \checkmark & \checkmark \\
	& & Li et al.~\cite{li2021hermes} & NN & PT & \texttimes & \checkmark \\
	& & Lu et al.~\cite{lu2021heterogeneous} & NN & PT & \texttimes & \checkmark \\
 \cmidrule{2-7} 
  & \multirow{5}{*}{\begin{tabular}{@{}c@{}} Knowledge\\Distillation \end{tabular}} 
	& Zhang et al.~\cite{zhang2021fedzkt} & NN & PT & \texttimes & \checkmark \\
	& & Liu et al.~\cite{Liu2022heterofl} & NN & PT & \checkmark & \checkmark \\
	& & Ozkara et al.~\cite{ozkara2021quped} & NN & PT  & \texttimes & \checkmark \\
	& & He et al.~\cite{he2020groutknowledgetransfer} & NN & PT  & \texttimes & \checkmark \\
	& & Bistritz et al.~\cite{bistritz2020distributed} & NN & PT  & \texttimes & \checkmark \\
 \cmidrule{2-7} 
  & \multirow{2}{*}{\begin{tabular}{@{}c@{}} Matrix\\Factorization \end{tabular}} 
	& \multirow{2}{*}{\begin{tabular}{@{}c@{}} Yao et al.~\cite{yao2021fedhm} \end{tabular}}  & \multirow{2}{*}{\begin{tabular}{@{}c@{}} NN \end{tabular}}  & \multirow{2}{*}{\begin{tabular}{@{}c@{}} PT \end{tabular}} & \multirow{2}{*}{\begin{tabular}{@{}c@{}} \texttimes \end{tabular}}  & \multirow{2}{*}{\begin{tabular}{@{}c@{}} \checkmark \end{tabular}}  \\
	& & & & & & \\
 \bottomrule
\end{tabular}
\vspace{2mm}
\caption{Comparison of existing works for system heterogeneous FL. 
LM denotes Linear Models. FM denotes Factorization Machines. NN denotes Neural Networks. DT denotes Decision Trees. 
HE denotes Homomorphic Encryption. MPC denotes Secure Multiparty Computation. DP denotes Differential Privacy. PT denotes Plain-text.}
\vspace{-3mm}
\label{system_FL_summary}
\end{table*}

\section{Model Heterogeneity}\label{sec-model_FL}

Different parties may hold heterogeneous features and various device resources in FL. Therefore, it is desirable to train heterogeneous models for participants accordingly. 
However, averaging-based model aggregation algorithms~\cite{McMahan2016ModelAvg,Kamp2018MA} are not valid in such settings due to the infeasibility of parameter matching among local models. 

The requirement of \textit{model heterogeneity} in FL comes from three aspects: data space, statistical, and system heterogeneity. 
1) For data space heterogeneous FL where the feature spaces differ among different parties, each party must train a heterogeneous local model taking data from different feature spaces as input for label prediction. 
For example, in healthcare applications, a hospital has medical image data such as CT and MRI, and another has tabular medical record data. They may cooperate to do VFL by training a CNN model for medical images and a decision tree model for tabular data. 
2) For statistical heterogeneous FL, parties can train heterogeneous local models to tackle non-IID and imbalanced data. 
3) For system-heterogeneous FL, the participating devices or edges may have different constraints regarding performance, computation, and communication. The participants desire to tailor their own model architectures. 
For example, we can train human activity recognition models on smartphones and wearable devices. Smartphones have powerful computation resources. However, wearable devices collect more accurate movement data but have fewer computation resources. 
Therefore, we can train larger models for smartphones and lightweight models for wearable devices. 
In summary, training heterogeneous local and global models is a critical technique to tackle the three heterogeneity issues in FL. 

There are many approaches proposed to train heterogeneous models. When different parties have heterogeneous features such as VFL, it is natural to build a specific model for each feature space. The different models can be integrated into one model or aggregated for prediction. 
As a representative type of model compression, knowledge distillation~\cite{kdHinton2015,NIPS2014_ea8fcd92} adopts a "teacher-student" learning paradigm to learn a small model from a large teacher model effectively. 
Knowledge distillation relaxes the stringent requirement of homogeneous local models as it takes the logits as representations for knowledge transfer, providing an approach to building FL systems with heterogeneous model architectures. 

In this section, we will introduce some approaches to training heterogeneous model architectures and discuss how to apply the heterogeneous model to tackle each type of heterogeneity in FL.

\subsection{For Data Space Heterogeneity}

\subsubsection{Representation Learning}
In data space heterogeneous FL, parameter sharing-based approaches are widely studied to learn latent representations from different feature spaces based on aligned samples~\cite{YangLiu2019Secure,kang2022fedcvt,li2022semi,ren2022improving}. 
The representation learning approaches build a party-specific model for each party. 
First, each party trains a local model on their local features to extract latent representations. 
Then, a top global model is trained based on the feature representations extracted by each local model for prediction. 

Liu et al.~\cite{YangLiu2019Secure} proposed a secure federated transfer learning (SFTL) approach to tackle the instance-sharing Hetero-FTL problem. SFTL first maps the feature spaces from two parties to a common latent space and then trains a classifier for prediction. 
To fully use the local information, the feature representations extracted by local models can be concatenated as the input of the task-specific top model. However, it requires the samples have full features for training and inference, which may not be feasible for real-world multi-party VFL applications. 
Moreover, the proposed SFTL can only transfer knowledge from one source party to one target party. The knowledge from both parties can not be transferred mutually. The number of participants can neither scale up. 

FedSVD~\cite{chai2022federated} is a federate singular vector decomposition (SVD) approach that can conduct SVD for horizontally and vertically partitioned data among multiple parties. FedSVD employs a random orthogonal linear transformation to map the private data into a random linear space. Then, the factorization server does standard SVD computation to decompose the masked data. Finally, the decomposed matrices are sent back to all clients to remove the random transformation.  
The decomposed matrices contain a shared matrix and a private matrix. For VFL, the shared matrix contains latent representations for heterogeneous features of each sample. While the private matrix is the personalized local model that maps heterogeneous feature spaces to the common latent space. 

\subsubsection{Tree-based Models}
Decision tree models are widely used to tackle data space heterogeneity in VFL. 
The decision trees trained among all the parties are naturally heterogeneous in architecture. 
Liu et al.~\cite{Liu_2020fedforest} propose a Federated Forest based on bagging and the CART tree. Concerning its own features, each party preserves the partitioned information. Performance-wise, the proposed approach is on par with the centralized version.
In order to accelerate homomorphic encryption, Fu et al.~\cite{fu2021vf2boost} propose a concurrent training protocol and tailored cryptographic operations, which resulted in the creation of the effective vertical federated GBDT technique known as VF2Boost.
Federated gradient boosting approaches generate a boosting model concurrently with the base learner, which can be laborious and communication-inefficient. As a result, Han et al.~\cite{han2022fedgbf} suggest combining boosting and bagging to create many decision trees concurrently.
By canceling decision routes and modifying a practical, secure inference approach, Chen et al.~\cite{Fed-EINI9671749} investigate the interpretability of decision tree ensembles in VFL during the inference phase.

\subsubsection{Data Augmentation}
Kang et al.~\cite{kang2022fedcvt} propose FedCVT to tackle the two-party instance-sharing Hetero-FTL via representation learning. 
The main idea is to predict the missing features of the other party, given the local features as input. Two parties train a feature prediction model based on their aligned samples and mutually predict the missing feature space and the pseudo-labels. 
Then, FedCVT picks the pseudo-labeled samples with high probability as training data. 
FedCVT trains three models on the augmented samples—two local classifiers for local prediction on local features and one federated classifier for federated prediction. 
Gao et al.~\cite{9005992} tackles the feature-sharing Hetero-FTL problem by using feature mapping to convert heterogeneous feature spaces to homogeneous feature space.

\subsubsection{Knowledge Distillation}
Knowledge distillation enables knowledge transfer across heterogeneous feature spaces from different parties, assuming there exist some aligned instances. 
Participants can hold heterogeneous local models in terms of architecture and parameters in federated knowledge distillation frameworks.  

Ren et al.~\cite{ren2022improving} proposed to transfer knowledge from privileged features in passive parties to the active party via knowledge distillation. Furthermore, an oblivious transfer is leveraged to protect data ID privacy during training and inference. 
Paillier homomorphic encryption is used to train federated models in a privacy-preserving manner. Oblivious Transfer (OT) is used to align samples without privacy leakage. 

Wang et al.~\cite{wangvertical} propose to transfer knowledge via representation distillation to tackle the multi-party instance-sharing Hetero-FTL problem. 
Specifically, they adopt FedSVD for representation learning. Then, they train an auto-encoder based on representation distillation to transfer knowledge from the full feature space of aligned samples to the feature space of the active party. 
Given the extracted task-independent feature representations, the active party can train a task-specific model with the concatenated representations as input. 
However, this approach can only transfer knowledge in a pair-wise manner. When the number of the passive party grows, the computation complexity grows linearly.

\subsection{For Statistical Heterogeneity}

Zhu et al.~\cite{pmlr-v139-zhu21b} proposed a data-free federated distillation algorithm by training a feature extractor and a feature representation generator. This proposed approach employs a feature extractor to project data into a latent space. 
Lin et al.~\cite{lin2020ensemble} proposed to use ensemble distillation for robust model fusion on heterogeneous local models. To transfer knowledge, an unlabeled dataset or a generator is used to sample data for all participants to compute logits and distill knowledge.  
Lan et al.~\cite{zhu2018knowledge} propose a multi-branch design that all local models share a common feature extractor, and each has a unique unshared branch. A gate is used for the weighted aggregation of the logits of all the branches. Although this approach is proposed for centralized learning, it can be modified to fit model heterogeneous FL problems. 

Li et al.~\cite{li2021decentralized} propose a peer-to-peer decentralized, FL approach via mutual knowledge distillation. Specifically, in each communication round, a group of parties is selected to conduct FL, and then the trained models are sent to the second group of parties. The parties in the second group compute two soft logits computed by their local models and the received models. These two logits are aggregated as teacher logits to train both the local and the received models. 
Yang et al.~\cite{yang2021mutualpfl} propose personalized federated mutual learning (PFML) to use the non-IID characteristics to generate tailored models for each party. Specifically, the PFML approach integrates mutual learning into the local update process in each party to improve the performance of both the global model and the personalized local models. The convergence rate is also accelerated by mutual distillation. 
FedH2L~\cite{li2021fedh2l} is proposed to address statistical and model heterogeneity via mutual distillation simultaneously. It exchanges only posteriors on a shared seed set between participants in a decentralized manner.

\subsection{For System Heterogeneity}
\subsubsection{Representation Learning}
HeteroFL~\cite{diao2021heterofl} proposes to train heterogeneous neural networks with various widths by averaging the overlapped sub-matrices of all parameter matrices across various models. 
The number of parameters in smaller models is reduced in width by cropping each parameter matrix of the largest model into a sub-matrix with various reduction radios.

FedMask~\cite{li2021fedmask} is proposed to achieve communication and computation efficient FL. 
In FedMask, each party learns a heterogeneous and structured sparse neural network. To achieve this, each party learns a sparse binary mask while freezing the parameters of the local model. Only the binary masks are communicated between devices and the server. 
The model aggregation is conducted by averaging the unmasked parameters for each local model. 
The authors of FedMask also propose a similar approach named Hermes~\cite{li2021hermes} that trains personalized and sparse neural networks by applying structured pruning. Only the updates of these sub-networks are transmitted. To aggregate models, the server performs the average on only overlapped parameters across each sub-network. The lower layers are retained and aggregated to transfer knowledge among heterogeneous models, and the higher layers are pruned by channel. 
Lu et al.~\cite{lu2021heterogeneous} propose to use model mapping to fuse heterogeneous models in FL. 

\subsubsection{Knowledge Distillation}
QuPeD, which Ozkara et al.~\cite{ozkara2021quped} present, allows clients to train tailored compressed models with various quantization settings and model dimensions. QuPeD takes into account that clients with different resources have different requirements for inference efficiency and test accuracy.
The proposed technique first trains quantized models using a relaxed optimization problem that also optimizes quantization values. Then, by transferring information from the global model to quantized local models, knowledge distillation is used to train compressed local models.


A client-inclusive FL approach called InclusiveFL~\cite{Liu2022heterofl} is presented. 
It gives clients with varied computing capacity access to models of various sizes.
First, because the lower layers are crucial for maintaining the information in downstream tasks, the shallow bottom model layers in the biggest model are shared with other smaller local models in order to reduce the mismatch between small and large local models.
The parameters of common bottom levels are then updated using a layer-wise heterogeneous aggregation approach.
In order to better transfer information from large models with powerful customers to small models with weak clients, a momentum knowledge distillation approach is finally presented.
Specifically, to transfer knowledge from a large model to small models, the top encoder layer in small models is trained by momentum distillation to imitate the behavior of the top encoder layers in a larger model.

\section{Apply Transfer Learning to Tackling Statistical, Data Space, System and Model Heterogeneity}\label{sec-TL_HeteroFL}


Federated learning faces heterogeneity in terms of data space, statistics, and system. 
In this section, we discuss different transfer learning approaches that can be adopted to address the different heterogeneity in FL.

\subsection{Knowledge distillation}\label{noniid_KD}
This section provides a comprehensive survey of knowledge distillation and investigates how knowledge distillation can be used to tackle the model heterogeneity issue in FL. 
We first introduce the definition of knowledge distillation. Then, we review existing works that leverage knowledge distillation to tackle the model heterogeneity problem in FL. Interested readers may refer to~\cite{Gou2021kdsurvey} for a more detailed introduction. 

\subsubsection{Distillation Approaches}
Knowledge distillation was initially proposed to compress an ensemble of neural networks~\cite{kdHinton2015}. 
Three forms of knowledge can be leveraged to transfer knowledge from the teacher model to the student model: response-based knowledge, feature-based knowledge, and relation-based knowledge. 
Both response-based and feature-based knowledge leverage the output of specific layers to transfer knowledge from teacher to student. 
Relation-based knowledge takes the instance-wise or layer-wise distance as knowledge to train the student model. 
We elaborate these three forms of knowledge in Appendix~\ref{appendix_knowledge_distillation}.

Knowledge distillation for heterogeneous FL falls into the \textit{group-based online distillation} scheme, which is an end-to-end training scheme.
The idea of group-based online distillation is to train multiple student local models simultaneously by learning from ground-truth labels and distilling from their aggregated soft targets, a specific form of aggregation of intermediate peer predictions. 
In federated online distillation, the teacher model is the aggregation of the local models, and each local model is a student model~\cite{guo2020online,zhang2018deep,zhu2018knowledge}. 
Therefore, the teacher model is updated simultaneously with the student models, and the whole knowledge distillation framework is trained in one phase. The online distillation approaches proposed for non-federated settings are promising to be adopted to tackle heterogeneous FL problems. 

Data-free knowledge distillation is also explored in heterogeneous FL to relax the requirement of a public dataset for knowledge transfer. 
Nayak et al.~\cite{nayak2019zero} first proposed zero-shot knowledge distillation by synthesizing the data impressions from the teacher model and using them as surrogates for knowledge distillation. Micaelli and Storkey~\cite{micaelli2019zero} achieved data-free distillation by training an adversarial generator to search for samples on which the student poorly matches the teacher and then using them to teach the student model.

\subsubsection{Existing Studies}

\subsubsubsection{a) Data space heterogeneity.}
Knowledge distillation can be used to tackle instance-sharing Hetero-FTL problems, where the participants have different feature spaces and some aligned instance IDs. In such settings, the knowledge can be transferred by the logits of the aligned instances output by the heterogeneous local models. 

Only a few recently proposed works use knowledge distillation to tackle the instance-sharing Hetero-FTL problem. 
To enable federated inference on non-overlapping data in VFL, Ren et al.~\cite{ren2022improving} proposed to distill knowledge from privileged features that are only available in training phase in passive parties to the active party. 
The active party trains a local model using the VFL model trained on aligned samples as the teacher model. Thereby, the collaboratively trained local model can predict un-aligned samples with higher accuracy than the locally trained model.
Furthermore, Paillier's HE is leveraged to achieve privacy-preserving model training. Oblivious transfer is adopted to protect sample ID privacy in training and inference phases. 
Wang et al.~\cite{wangvertical} representation distillation to propose a task-independent knowledge transfer approach. 
First, all parties collaboratively learn latent representations for aligned samples via FedSVD~\cite{chai2022federated}. 
FedSVD maps heterogeneous features from all parties to a latent space. 
Secondly, the active party trains an auto-encoder on the feature representations, to distill knowledge from aligned samples to un-aligned samples. 
Finally, the active party conducts task-specific training and inference on the enriched representations and local features. 
However, this approach transfers knowledge from each passive party in a pair-wise manner, and the dimension of input features of the task-specific model grows linearly as the passive party number grows. 

Li et al.~\cite{li2022semi} propose SplitKD to first train a teacher VFL model on aligned samples, then teach a student local model for the active party via knowledge distillation. 
The teacher VFL model consists of a top model for classification and personalized bottom models for each party. 
In the distillation phase, the active party's local model is initialized by the pre-trained bottom model and finetuned by distilling knowledge from the VFL model. 

Online mutual knowledge distillation methods~\cite{guo2020online,zhang2018deep} simplify the training process into one stage, where the soft predictions of each model are aggregated into one teacher prediction. The teacher prediction updates all local models together with the ground truth of private labeled data. 
Taking a two-party instance-sharing Hetero-FTL problem as an example, two parties first locally train heterogeneous student models with their private data in different feature spaces. 
Then, they can use mutual knowledge distillation to update two local models over the aligned instances collaboratively. 
In mutual knowledge distillation~\cite{zhang2018deep}, two student models learn from each other by taking the logits from the other model as soft targets.  

Guo et al.~\cite{guo2020online} consider a centralized setting with homogeneous data, treat all the DNNs as student models, and collaboratively update them in a single stage. Although this approach is proposed for centralized learning, it can be modified to tackle data space heterogeneity. 
Since each party trains a compact local model, the active party can conduct inference by ensemble aggregation with an arbitrary number of aligned parties. The active party can use its local model for inference to predict unaligned test data. On the other hand, when the test data is aligned across several parties, all parties with aligned features collaboratively predict with the ensemble of local models.
Zhu et al.~\cite{pmlr-v139-zhu21b} propose a data-free federated distillation algorithm that employs a feature extractor to project data into a latent space.
Although being proposed for HFL, this approach can be adopted to tackle system or data space heterogeneity by training heterogeneous feature extractors for different parties. 
Similarly, He et al.~\cite{he2020groutknowledgetransfer} consider a centralized setting and propose to train multiple feature extractors and a classifier via knowledge distillation. The local feature extractor models are unique for each heterogeneous feature space and map input data to a common latent subspace. The local feature representations and logits are transmitted to a central server to train a global classifier. The predictions of the global teacher classifier are then sent back to each party for knowledge transfer.

\textit{Cross-modal distillation} is another line of related studies in data space heterogeneous FL. 
Although existing works are proposed for centralized learning, these works can inspire us to propose approaches for instance-sharing Hetero-FTL.
Peng et al.~\cite{peng2021hierarchical} address the visual-textual cross-modal learning problem by introducing a life-long knowledge distillation approach. A hierarchical recurrent network is constructed to leverage information from both semantic and attention levels through adaptive network expansion. 
Tian et al.~\cite{tian2019contrastive} propose to use a contrastive loss to transfer pair-wise information across different modalities. 
Existing cross-modal distillation methods rely on aligned or paired instances between two modalities~\cite{gupta2016cross,tian2019contrastive,thoker2019cross}. Cross-modal distillation is restricted by the lack of aligned samples across different modalities.

\subsubsubsection{b) Statistical heterogeneity.}
In general, there are four main types for distillation-based FL architectures: 1) distillation of knowledge to each client to learn stronger personalized models; 2) distillation of knowledge to the server to learn stronger server models, 3) bidirectional distillation to both the clients and the FL server; and 4) distillation amongst clients.

FedMD~\cite{li2019fedmd} is a distillation-based FL framework that leverages knowledge distillation to enable parties to train unique local models on their private local data.
The main idea is to transfer knowledge by aggregating logits of local models computed on a public dataset shared by the parties. 
For every communication round, each party trains its model based on the teacher logits computed on the public dataset as well as the logits computed by its local model on its private dataset. 
Thereby, each party learns a personalized model by transferring knowledge from other parties. 
Chen et al.~\cite{chen2020online} propose online distillation with diverse peers by incorporating an attention-based mechanism to generate a distinct set of weights for each local model. 
Each party assigns individual weights to all the local models during aggregation to derive their own target distributions. The first stage conducts group-based distillation among all local models for knowledge transfer with diverse target distributions. 
Then, the second distillation stage transfers the knowledge from all local models to a global model. 

Chang et al.~\cite{chang2019cronus} propose a two-stage robust knowledge distillation approach to tackle heterogeneous distributions. In the first cold-start stage, each party trains a local model on private data. In the second stage, the clients transfer knowledge by sharing the logits of public data. The server aggregates the logits via the robust mean estimation algorithm RobustFilter~\cite{diakonikolas2017being}. Then the clients update local models using private data and soft-labeled public data. 
Zhang et al.~\cite{zhang2021parameterized} introduce a knowledge coefficient matrix to train personalized local models via knowledge distillation. The proposed KT-pFL updates the personalized soft logits of each party by a linear combination of all local soft logits using a knowledge coefficient matrix, which can adaptively enhance knowledge transfer among parties with similar data distribution. The knowledge coefficient matrix and the model are alternatively updated in each round via gradient descent. 
Gong et al.~\cite{gong2022preserving} propose a one-shot offline knowledge distillation approach named FedKD, using unlabeled public data. A quantized and noisy ensemble of local predictions from fully trained local models is proposed for privacy guarantees without sacrificing accuracy.



Lan et al.~\cite{zhu2018knowledge} propose a multi-branch design that all local models share a common feature extractor, and each has a unique unshared branch. A gate is used for the weighted aggregation of the logits of all the branches. Although this approach is proposed for centralized learning, it can be modified to fit model heterogeneous FL problems. 
Afonin and Karimireddy~\cite{afonin2021towards} propose federated kernel ridge regression to achieve model-agnostic FL. 
FedGKT~\cite{he2020groutknowledgetransfer} is also proposed to improve model personalization performance via group knowledge transfer. It leverages alternating minimization to train small local models and a large global model via bidirectional knowledge distillation.

Lin et al.~\cite{lin2020ensemble} proposed to use ensemble distillation for robust model fusion on heterogeneous local models. To transfer knowledge, an unlabeled dataset or a generator is used to sample data for all participants to compute logits and distill knowledge.  
It assumes a setting in which the edge clients require different model architectures due to diverse computational capabilities. 
The server constructs p distinct prototype models, each representing clients with identical model architectures (e.g., ResNet and MobileNet). For each communication round, FedAvg is first performed among clients from the same prototype group to initialize a student model. Cross-architecture learning is then performed via ensemble distillation, in which the client (teacher) model parameters are evaluated on an unlabelled public dataset to generate logit outputs that are used to train each student model in the server.

To eliminate the dependence on a public dataset in FL, Zhu et al.~\cite{pmlr-v139-zhu21b} proposed data-free federated knowledge distillation by training a feature representation generator. 
The server trains the generator to aggregate the distribution of class representations from all participants. Then, the generator is sent to participants whose knowledge is distilled to local models. 
This proposed approach employs a feature extractor to project data into a latent space. 
Due to system heterogeneity or data space heterogeneity, the feature extractor can be heterogeneous for each party. 
However, it remains a challenge to generate high-quality and diverse data for data-free knowledge distillation to improve the model generalizability.

\textit{Mutual Distillation.}\label{KDMutualDistillation}
Deep Mutual Learning~\cite{zhang2018deep} has been explored to train personalized local models to tackle statistical heterogeneity in a decentralized manner. It allows two models to learn from each other from their predicted logits during the training process. 
Recently, many works have explored mutual knowledge distillation in FL to train heterogeneous local models to address the non-IID issue~\cite{li2021decentralized,yang2021mutualpfl,matsuda2022fedme,li2021fedh2l}.
The key idea of mutual federated distillation is to take the local model of another party or group as a teacher model and use the predicted posteriors to each their own local model. The local models are exchanged among parties for mutual knowledge transfer. It enables each party to train a heterogeneous model in a decentralized manner. 

Li et al.~\cite{li2021decentralized} propose a peer-to-peer decentralized, FL approach via mutual knowledge distillation. Specifically, in each communication round, a group of parties is selected to conduct FL, and then the trained models are sent to the second group of parties. The parties in the second group compute two soft logits by their local models and the received models. These two logits are aggregated as teacher logits to train both the local and the received models. 

Yang et al.~\cite{yang2021mutualpfl} propose personalized federated mutual learning (PFML) to use the non-IID characteristics to generate tailored models for each party. Specifically, the PFML approach integrates mutual learning into the local update process in each party to improve the performance of both the global model and the personalized local models. The convergence rate is also accelerated by mutual distillation. 

Koji et al.~\cite{matsuda2022fedme} propose FL via Model exchange (FedMe), which personalizes models with automatic model architecture tuning during the learning process. 
In FedMe, clients exchange their models for model architecture tuning and model training. 
First, to optimize the model architectures for local data, clients tune their own personalized models by comparing them to exchanged models and picking the one that yields the best performance. 
Second, clients train both personalized and exchanged models by using deep mutual learning~\cite{zhang2018deep}, despite different model architectures across the clients.
Similarly, FedH2L~\cite{li2021fedh2l} is proposed to address statistical and model heterogeneity via mutual distillation simultaneously. It exchanges only posteriors on a shared seed set between participants in a decentralized manner. 

To summarize, mutual knowledge distillation makes FL communication efficient and model agnostic and can effectively transfer knowledge among heterogeneous data silos.

\subsubsubsection{c) System heterogeneity.}\label{KDForSystemHeterogeneity}
Parameter-sharing approaches~\cite{FelixSattler2019NonIID} require all local models have the same model architecture. 
However, in a realistic business setting, participants may desire to design their unique model architecture due to system heterogeneity reasons such as computation or communication capacity. 
This kind of model heterogeneity is caused by system heterogeneity and poses new challenges to traditional FL. 

FedZKT~\cite{zhang2021fedzkt} is a zero-shot knowledge transfer algorithm that allows devices to determine the on-device models based on their local resources independently. The server receives heterogeneous local models and adversarially trains generator with the by aggregating the soft predictions of the local models. 
The distilled central knowledge is then sent back as the corresponding on-device model parameters, which can be easily absorbed on the device side. 
The generator is trained to maximize the discrepancy between the global model prediction and the prediction of the ensemble of local models.
FedZKT uses a bidirectional knowledge transfer. To train the global model, the global model is trained using the averaged soft targets of all local models. 
Then, the global model is used as the teacher model to transfer knowledge to each local model. 

Chan and Ngai~\cite{hin2021fedhe} introduce knowledge distillation to train heterogeneous models in the clients. The logits of each category are transmitted and aggregated in the server. Then, the clients match the average logits with the same label to the instances. Therefore, the knowledge is transferred among parties via class-level logit aggregation.

He et al.~\cite{he2020groutknowledgetransfer} propose group knowledge transfer as a way to enhance the performance of model customization for resource-constrained edge devices. Through a bidirectional distillation method, it trains tiny edge models on edge and a big server model on the server using alternating minimization. The big server model employs the KL-divergence loss to reduce the discrepancy between the ground truth and soft labels predicted by the local models after receiving extracted features from the local models as inputs. The server model then incorporates the information passed from the local models. Similarly, each local model uses its own private dataset and the projected soft labels downloaded from the server to calculate the KL-divergence loss. This streamlines the dissemination of information from the server model to the local models.

Bistritz et al.~\cite{bistritz2020distributed} propose a distillation distributed algorithm for on-device learning and is architecture independent. It assumes that all edge devices are connected to just a small number of nearby devices in an IoT edge FL environment. Only devices that are linked can exchange data. 
The learning technique is semi-supervised; local training is done on private data, while federated training is done on a publicly available dataset that is not labeled. Each client broadcasts its soft decisions to its neighbors throughout each communication cycle while simultaneously receiving their broadcasts. 
Each client then uses a consensus technique to update its soft decisions depending on those of its neighbors. The client's model weights are then updated using the revised soft choices by regularizing its local loss. This procedure facilitates model learning via knowledge transfer amongst neighboring clients in a network.

To learn tailored model architectures for different devices, knowledge distillation is widely adopted for knowledge transfer among heterogeneous models. 
Ozkara et al.~\cite{ozkara2021quped} propose quantized personalization via distillation (QuPeD) that clients with diverse resources have various requirements on inference efficiency and test accuracy and allow clients to learn personalized compressed models with diverse quantization parameters and model dimensions. 
The proposed algorithm first trains quantized models through a relaxed optimization problem, where quantization values are also optimized. Then, knowledge distillation is introduced to train compressed local models by transferring knowledge from the global model to quantized local models. 

Recently, a client-inclusive FL method InclusiveFL~\cite{Liu2022heterofl} is proposed. It assigns models of different sizes to clients with different computation resources. 
First, to eliminate the mismatch between small and large local models, the shallow bottom model layers in the largest model are shared with other smaller local models, as the lower layers are most critical for retaining the knowledge in downstream tasks. 
Then, a layer-wise heterogeneous aggregation method is proposed to update the parameters of shared bottom layers. 
Finally, a momentum knowledge distillation method is proposed to better transfer knowledge in big models on powerful clients to the small models on weak clients. 
Specifically, to transfer knowledge from a large model to small models, the top encoder layer in small models is trained by momentum distillation to imitate the behavior of the top encoder layers in a larger model. 

Both QuPeD and InclusiveFL adopt a hierarchical FL architecture. 
During FL, the devices are first grouped according to capacity. Devices in each group conduct homogeneous FL to train a model. Then, models with different sizes are aggregated for knowledge transfer.

\subsection{Data Augmentation}

\subsubsection{Data space heterogeneity}
In feature-sharing Hetero-FTL problems, participants have some common features but no aligned instance IDs, as shown in Figure~\ref{featureSharing}.
Gao et al.~\cite{9005992} first studied the problem of privacy-preserving feature-sharing Hetero-FTL and proposed to use feature mapping to convert heterogeneous feature spaces to homogeneous feature spaces. 
The data are augmented by predicting its missing features.
Participants first learn instance weights over the shared feature space for domain adaptation. For heterogeneous mutual transfer learning, each party trains a linear model locally to map the shared features to the local ones. Then, the other party fulfills the missing feature space via secure computation. Once the whole feature space is fulfilled, the feature-sharing Hetero-FTL problem is converted into a HFL problem. 
However, this approach only applies to linear models. 

To address the instance-sharing Hetero-FTL problem, 
Kang et al.~\cite{kang2022fedcvt} propose FedCVT to estimate the representations of missing features via semi-supervised learning. 
Two parties jointly predict the missing features and pseudo-labels by training a feature prediction model on the aligned samples.
FedCVT then selects the highly likely pseudo-labeled samples as training data. 
FedCVT uses the imputed (augmented) samples to jointly train three models, including one federated classifier for federated prediction and two local classifiers for local prediction on local features.

Yitao et al.~\cite{Yitao2022multiview} propose FedMC to align similar unaligned data via data collaboration. 
FedMC first learns a latent feature space on aligned and labeled samples. Then, it measures the distance between each pair of unaligned samples from the active party and passive party in this latent space. 
Then, FedMC aligns two samples that are similar in the latent space in a pair and adds the aligned samples to the training set.
Finally, FedMC trains a joint federated model on the expanded training set. 

\subsubsection{Statistical Heterogeneity}
A data sharing technique that gives each client a modest quantity of global data that is balanced by classes, was proposed by Zhao et al.~\cite{zhao2018fednoniiddata}. Their research demonstrates that even a tiny amount of data can increase considerable accuracy. 
The FedHome approach~\cite{wu2020fedhome} trains a generative convolutional auto-encoder model. Following the FL process, each client further customizes a locally enhanced class-balanced dataset. This dataset is created by applying a synthetic data generation on the encoder network's low-dimensional features using local data.
A self-balancing FL framework named Astraea~\cite{duan2020self} uses Z-score-based data augmentation and down-sampling of local data to address the class imbalance. The server needs statistical data on the local data distributions of the clients (e.g., class sizes, mean, and standard deviation values).
FAug, a federated augmentation strategy proposed by Jeong et al.~\cite{jeong2018communication}, includes training a generative adversarial network model in the server. 
To train the GAN model, certain data samples from the minority classes are submitted to the server. Each client receives a copy of the trained GAN model, which it uses to generate extra data to supplement its own local data and create an IID dataset.

\subsection{Parameter Sharing}

\subsubsection{Data space heterogeneity}

The parameter sharing approaches use representation learning to map heterogeneous features into a common latent space. Then, a classifier is trained to map the latent feature representations to the label space.

\subsubsubsection{a) Instance-sharing Hetero-FTL.}
We first introduce representation learning methods proposed for instance-sharing Hetero-FTL problems, in which participants have some aligned instance IDs but have no feature in common, as shown in Figure~\ref{instanceSharing}.
Liu et al.~\cite{YangLiu2019Secure} first study the problem of secure federated transfer learning in the instance-sharing Hetero-FTL setting. They assume two parties holding heterogeneous features have limited aligned instances. They first train two neural networks to map two feature spaces into a common latent subspace to transfer knowledge from a source party to a target party. Then, the two parties conduct HFL to train a neural network model to map the latent subspace to the label space. Additive HE and secret sharing are incorporated into the training algorithm to protect data privacy. The secure approach provides the same level of accuracy as the plain-text method. 
As a further improvement, Sharma et al.~\cite{Sharma2019SecureAE} leverage a more efficient secure computation framework named SPDZ~\cite{cryptoeprint:2011:535} for efficient training. 
Kang et al.~\cite{kang2022fedcvt} propose a semi-supervised learning approach that uses limited aligned instances to boost the model performance of each party. This approach estimates representations for missing features, predicts pseudo-labels for unlabeled samples to expand the training set, and trains three classifiers jointly based on different views of the expanded training set to improve the VFL model's performance. Neither raw data nor model parameters are transmitted between parties to protect data privacy. 
He et al.~\cite{he2022hybrid} propose FedHSSL, a federated hybrid semi-supervised learning framework to fully use all available data (including un-labeled and un-aligned) data for training. FedHSSL exploits generic feature representations shared among different feature spaces to boost the
performance of the joint model via partial model aggregation. In FedHSSL, each party has a two-tower structured model. One sub-model is trained on aligned data for supervised representation learning; the other is trained on un-labeled data via self-supervised learning. 

Instead of learning feature representations in a common latent subspace, it is also possible to directly learn a mapping from one feature space to another. 
Hou et al.~\cite{hou2021prediction} consider a centralized setting and study the problem of learning with feature evolution in a data stream, where all old features in the data stream vanish unpredictably, and new features emerge simultaneously. 
During the feature evolution, there is an overlapping period in which, for each data sample, the old features gradually vanish, and new features appear. 
The authors first use least squares to learn a mapping from the new feature space to the old one. Therefore, the knowledge of the previous model can be transferred to the new feature space. 
Ensemble learning is leveraged to use both the old and new features without manually deciding which base models should be incorporated. 
Instead of transferring knowledge to a new feature space in~\cite{hou2021prediction}, instance-sharing Hetero-FTL requires knowledge transfer between different feature spaces. 

\subsubsubsection{b) Feature-sharing Hetero-FTL.}
Several transfer learning approaches proposed for non-FL settings can be adopted to tackle the feature-sharing Hetero-FTL problem. 
Zhu et al.~\cite{zhu2011heterogeneous} study heterogeneous transfer learning in image classification by leveraging documents and tags as auxiliary data. The tags are shared features that transfer knowledge between image and text space via collective matrix factorization. 
As a related non-FL approach, ReForm is proposed by Ye et al.~\cite{ye2018rectify} to reuse a previous model by rectifying via heterogeneous predictor mapping. 
Each party locally trains a mapping from shared features to local features as feature meta representation (meta-feature) of the local features. Then, a mapping between the meta-features is trained to convert the model parameters of one local feature space to the model parameters of the other local feature space. 
However, this approach only applies to scenarios with a few overlapped features and makes multiple assumptions about the data distribution.

\subsubsubsection{c) Label-sharing Hetero-FTL.}
In label-sharing Hetero-FTL, participants have neither feature in common nor aligned instance ID, and they only have the same label space, as shown in figure~\ref{labelSharing}. 
When the heterogeneous feature spaces from different parties are similar, all parties can leverage representation learning to train local models to map their local feature space to a common latent subspace. Then, a shared model can be collaboratively trained to map the feature representations in the latent subspace to the label space. 
Ju et al.~\cite{Ju2020FederatedTL}, and Gao et al.~\cite{Gao2019HHHFLHH} propose a neural network-based framework to tackle heterogeneous electroencephalographic (EEG) signal classification. Clients use different devices to collect EEG signals from heterogeneous feature spaces in their settings. Moreover, they aim to train federated models to predict the same task. 
To transfer knowledge among clients, they train a neural network model for each client to map heterogeneous features to a low-dimensional symmetric positive definite manifold. To align all heterogeneous samples in each class in the reduced manifold, they adapt maximum mean discrepancy (MMD)~\cite{gretton2006kernel} as a constraint. 
Then, they train a neural network via federated aggregation to map the manifold to the label space. 
Gao et al.~\cite{Gao2019HHHFLHH} further consider that different groups of clients share the same feature space, and all groups of clients collaboratively map their heterogeneous features into a common manifold by training a unique manifold reduction layer in each group. Then, all clients conduct HFL to train a federated layer for prediction. 
Liang et al.~\cite{liang2020think} harnessed representation learning to handle heterogeneous data and efficient global model updates. The feature representations can be learned via local supervised learning, unsupervised learning, self-supervised learning, or fair representation learning.

\subsubsection{Statistical heterogeneity}

Parameter sharing approaches decouple the local model parameters from the global model parameters, and local models share some parameters with the global model. Local parameters are trained on the parties, and only the global parameters are sent to the server. 
The parameter sharing approaches can conduct either 1) representation learning by sharing the bottom layers or 2) multi-task learning by sharing the top layers with the global model. 
In representation learning approaches, all parties learn personalized bottom layers as feature extractors and collaboratively train a global model for prediction. 
Singhal et al.~\cite{singhal2021federated} proposed federated reconstruction based on model-agnostic meta-learning for partially local FL suitable for training and inference at scale. The model parameters are partitioned into global and local parameters. For each communication round, each party receives the global model and reconstructs its local model, and locally updates the received global model. Then, the server aggregates updates to only the global model across parties. 
Bui et al.~\cite{bui2019federated} train a document classification model using a bidirectional LSTM architecture by treating user embeddings as the private model parameters and character embeddings as the global model parameters. 
Liang et al.~\cite{liang2020think} propose to combine local representation learning and global FL. 
Moreover, the approaches based on personalized bottom layers and global top layers can be easily extended to data space heterogeneous FL when the sample IDs are aligned. 
For sharing top layers for multi-task learning, Arivazhagan et al.~\cite{arivazhagan2019federated} proposed a setting where all parties share a global feature extractor, and each party trains a personalized task-specific model.

\subsubsection{System heterogeneity}

The parameters of the lower layers are most critical for retaining the knowledge in downstream tasks. Therefore, sharing the lower layers among heterogeneous models can effectively transfer knowledge from large models to small models. 

HeteroFL~\cite{diao2021heterofl} proposes to train heterogeneous neural networks with various width by averaging the overlapped sub-matrices of all parameter matrices of all parameter matrices across various models. 
The number of parameters in smaller models are reduced in width by cropping each parameter matrix of the largest model into sub-matrix with various reduction radio.
With this construction, HeteroFL adaptively allocates subsets of global model parameters to each party according to its corresponding capabilities. 
To perform aggregation, each parameter is averaged from the parties whose local models contain that parameter. 
However, this approach is not beneficial for retaining large models' knowledge in smaller models because the width pruning operation breaks the original model structure. 
Moreover, even the same sub-matrix may have different behaviors in the small and large models due to the model structure differences, which may lead to a sub-optimal performance due to the mismatch of feature spaces. 
In addition, simply sharing parts of parameters across different models cannot effectively transfer useful knowledge encoded by strong models to other weaker models. 
To tackle the above issue, Liu et al.~\cite{Liu2022heterofl} further propose InclusiveFL, in which the shallow bottom model layers in the largest model are shared with other smaller local models, to eliminate the mismatch between small and large local models. 

Split-Mix~\cite{hong2022efficient} proposes to train multiple models of different sizes and adversarial robustness levels tailored to the budget of each device, to address the under-training issue in large models. 
During training, strong parties sample more base models for local training and weak parties sample less base models. In each communication round, the server aggregates each base model using the updates from the corresponding parties. 
In inference phase, the trained local models can be aggregated on-demand according to inference requirements. 
Munir et al.~\cite{munir2021fedprune} proposed FedPrune to achieve inclusive FL. It prunes the global model for weak parties based
on their capabilities.

FedMask~\cite{li2021fedmask} is proposed to achieve communication and computation efficient FL. 
In FedMask, each party learns a heterogeneous and structured sparse neural network. To achieve this, each party learns a sparse binary mask while freezing the parameters of local model. Only the binary masks are communicated between devices and the server. 
The model aggregation is conducted by averaging on the unmasked parameters for each local model. 
The authors of FedMask also propose a similar approach named Hermes~\cite{li2021hermes} that trains personalized and sparse neural networks by applying the structured pruning. Only the updates of these sub-networks are transmitted. To aggregate models, the server performs the average on only overlapped parameters across each sub-network. To transfer knowledge among heterogeneous models, the lower layers are retained and aggregated and the higher layers are pruned by channel. 
Lu et al.~\cite{lu2021heterogeneous} propose to use model mapping to fuse heterogeneous models in FL.

\subsection{Domain Adaptation}
In order to improve personalization, domain adaptation seeks to lessen the domain disparity between a source domain and a target domain.
Three steps are often included in the training process: 1) global models using FL, 2) local models utilizing local data to change the global model, and 3) customized models using transform learning to improve the local model.

\subsubsection{Statistical heterogeneity}
A federated adversarial domain adaptation (FADA) algorithm is proposed by Peng et al.~\cite{peng2019federated}, which leverages adversarial adaptation to tackle the unsupervised federated domain adaptation problem. It aims to transfer knowledge from multiple source domains to one target domain. 
In FADA, the gradients of all local models are aggregated with a dynamic attention mechanism to update the global model. 
To transfer knowledge from among local models, FADA learns to extract domain-invariant features using adversarial domain alignment and a feature disentangler. 
To enable domain adaptation, an alignment layer, such as the correlation alignment (CORAL) layer~\cite{sun2016return} is often added before the softmax layer for adaptation of the second-order statistics of the source and target domains. 

\subsubsubsection{a) Model Interpolation}
Some work~\cite{hanzely2020federated,deng2020adaptive} propose to train a mixture of global and local models to balance generalization and personalization. 
Hanzely et al.~\cite{hanzely2020federated} use a penalty parameter to discourage the local models from being too dissimilar from the averaged global model. 
The degree of knowledge transferred among parties can be controlled by adjusting the penalty parameter. 
Similarly, Deng et al.~\cite{deng2020adaptive} proposed the APFL algorithm to find the optimal combination of global and local models by introducing a mixing parameter for each party. With the mixing parameter, the parties can adaptively transfer knowledge from the global model during the FL process. 

\subsubsubsection{b) Multi-Task Learning}
Multi-task learning can train a model that jointly performs several related tasks, thus widely used in FL to tackle statistical heterogeneity. Multi-task learning considers pairwise party relationships. 
Smith et al.~\cite{Smith2017MT} introduce federated multi-task learning and propose the MOCHA algorithm, which learns a separate model for each party. MOCHA leverages a primal-dual formulation for federated model optimization. Thus it is only applicable to convex models.  
Huang et al.~\cite{huang2021pcsfl} proposed FedAMP that leverages an attention-based mechanism to conduct pairwise collaboration for parties with similar data distributions. 
FedCurv~\cite{shoham2019overcoming} estimates parameter importance via the Fisher information matrix and uses penalization steps to avoid catastrophic forgetting when moving across learning tasks.

\subsection{Matrix Factorization}

\subsubsection{Data space heterogeneity}

Most studies on federated recommendations use collaborative filtering to address the feature-sharing Hetero-FTL problem, where participants have some features in common but have no aligned instance ID, as shown in Figure~\ref{featureSharing}. The matrix is low-rank factorized into user embeddings and item embeddings. 
Chai et al.~\cite{chai2019secure} proposed a secure federated matrix factorization (FedMF) method. In FedMF, each user holds user profiles (features) as private features, and the item profiles are shared among all users. 
In matrix factorization, the item features are trained jointly with the user features over private user-item interaction data. 
To conduct secure FedMF, the item features are encrypted by homomorphic encryption (HE) and transmitted to the recommendation system for aggregation. After aggregation, the item features are sent back to all users for local training. 
Ammad et al.~\cite{ammad2019federated} proposed the federated collaborative filtering (FCF) method, which is similar to FedMF.
Chen et al.~\cite{chen2020practical} studied the problem of privacy-preserving click-through-rate prediction in the recommendation and proposed PrivRec by adopting a factorization machine (FM)~\cite{rendle2010factorization} that consists of a linear model and the feature interaction model. 
In PrivRec, the clients locally store private user-item interaction data, and the recommender system server has public point-of-interest data about the items. 
For the linear model part, the parameters corresponding to user features remain on the client-side and are specific to each client. 
The linear model parameters corresponding to item features are the same across all clients. 
To protect user privacy, secret-sharing and local differential privacy are used to train the linear model. 
To train the feature interaction model, PrivRec sends all user features and item features to the recommender for aggregation. 
However, Gao et al.~\cite{gao2020FedFM} proved that the aggregation strategy used in federated FM might cause model leakage and user information leakage. 
Moreover, user-item interaction data is highly sparse, and secret-sharing converts sparse data into a dense matrix, making the existing accelerating approaches in plain-text data infeasible. 
To further improve the privacy against the model inversion attack and the data reconstruction attack in federated FM, Gao et al.~\cite{gao2020FedFM} proposed FedFM by combining the secret-sharing and HE for secure feature interaction. The sparsity of the interaction data can be utilized over HE encrypted data. It makes the training of the secure federated FM much faster.

\subsubsection{Statistical heterogeneity}
FedSVD~\cite{chai2022federated} is a federate singular vector decomposition (SVD) approach that can conduct SVD for horizontally and vertically partitioned data among multiple parties. FedSVD employees a random orthogonal linear transformation to map the private data into a random linear space. Then, the factorization server does standard SVD computation to decompose the masked data. Finally, the decomposed matrices are sent back to all clients to remove the random transformation.  
The decomposed matrices contains a shared matrix and a private matrix. 
For HFL, the shared matrix is the shared global model, while the private matrix is the 
For VFL, the shared matrix contains latent representations for heterogeneous features of each sample. While the private matrix is the personalized local models that maps heterogeneous feature spaces to the common latent space. 

\subsubsection{System heterogeneity}
Low-rank matrix factorization with heterogeneous model sizes is also explored to train heterogeneous models for devices with different hardware resources. 
FedHM~\cite{yao2021fedhm} is a heterogeneous federated model compression framework that distributes the heterogeneous low-rank models to clients and then aggregating them into a full-rank model. In FedHM, each party low-rank factorize the parameter matrix of the neural network with different sizes, according to its hardware resources. 
In each communication round, the server receives the factorized models and transforms the factorized low-rank models back to full-rank models. Then, the server conducts aggregation using weighted averaging.

\begin{table*}[!htb]
\vspace{8mm}
\centering
\footnotesize
\begin{tabular}{c | c  c } 
 \toprule
   Transfer Method & Advantages & Disadvantages \\
   \midrule
   \multirow{2}{*}{\begin{tabular}{@{}c@{}} Data \\ Augmentation \end{tabular}} & \multirow{2}{*}{\begin{tabular}{@{}c@{}} Easy to implement, can be built \\ on the general FL training procedure. \end{tabular}} & \multirow{2}{*}{\begin{tabular}{@{}c@{}} High risk of privacy leakage. \\ May require a public dataset. \end{tabular}} \\
   & & \\

   \midrule
   \multirow{2}{*}{\begin{tabular}{@{}c@{}} Parameter \\ Sharing \end{tabular}} & \multirow{2}{*}{\begin{tabular}{@{}c@{}} Simple formulation. \\ Easy to design heterogeneous models. \end{tabular}} &  \multirow{2}{*}{\begin{tabular}{@{}c@{}} Difficult to determine the optimal \\ privatization strategy.  \end{tabular}} \\
    & & \\

   \midrule
   \multirow{3}{*}{\begin{tabular}{@{}c@{}} Knowledge \\ Distillation \end{tabular}} & \multirow{3}{*}{\begin{tabular}{@{}c@{}} Easy to support model and system \\ heterogeneity. Communication efficient. \\ Low risk of privacy leakage. \end{tabular}} & \multirow{3}{*}{\begin{tabular}{@{}c@{}} May require a public proxy dataset. \end{tabular}}\\
   & & \\
   & & \\

   \midrule
   \multirow{2}{*}{\begin{tabular}{@{}c@{}} Domain \\ Adaptation \end{tabular}} & \multirow{2}{*}{\begin{tabular}{@{}c@{}} Reduce the domain discrepancy between \\ global and local models. \end{tabular}} & \multirow{2}{*}{\begin{tabular}{@{}c@{}} Single global model setup. \end{tabular}}\\
    & & \\

   \midrule
   \multirow{2}{*}{\begin{tabular}{@{}c@{}} Matrix \\ Factorization \end{tabular}} & \multirow{2}{*}{\begin{tabular}{@{}c@{}}  Easy to implement. \\ Training is efficient. \end{tabular}} & \multirow{2}{*}{\begin{tabular}{@{}c@{}} Cannot generate high-order representations. \end{tabular}} \\
    & & \\

 \bottomrule
\end{tabular}
\vspace{2mm}
\caption{Summary of transfer learning techniques to tackle heterogeneity in FL.}
\vspace{-3mm}
\label{compare_TL_method}
\end{table*}

\subsection{Summary}

In this section, we have discussed five transfer learning strategies to address the heterogeneity challenges in FL, including data augmentation, parameter sharing, knowledge distillation, and matrix factorization. As shown in Table~\ref{compare_TL_method}, we now summarize and compare the transfer learning techniques in terms of their advantages and disadvantages.

Table~\ref{fl_for_heteroFL} summarizes the transfer learning approaches adopted to tackle three types of heterogeneity in FL. We highlight the under-explored areas in red rectangles. 
The five transfer learning approaches can be grouped into three perspectives. 
The data-based strategy includes the data augmentation approaches. The architecture-based strategies include parameter sharing, knowledge distillation, and domain adaptation approaches. 
The model-based strategy includes matrix factorization approaches. 

We find parameter sharing (representation learning) and knowledge distillation are the most widely used approaches to transfer knowledge among parties and are widely used to address all types of heterogeneity. 
Knowledge distillation is still under-explored in Hetero-FTL. 
Some of the listed KD approaches were proposed for system-heterogeneous FL and are not tailored to Hetero-FTL. Therefore, it is desirable to develop KD-based approaches to tackle Hetero-FTL problems. 
Matrix factorization approaches are also explored for federated recommendation applications, as the federated recommendation is a feature-sharing Hetero-FTL problem by nature. 
However, matrix factorization approaches are still under-explored to tackle statistical and system heterogeneity. 

\begin{table*}[ht!]

\begin{frame}
\centering
\footnotesize
\begin{tikzpicture}
\node (table) {
\begin{tabular}{c c | c  c c } 
 \toprule
  Perspective & Transfer Method & \multicolumn{3}{c}{Heterogeneity Type} \\
     \cmidrule{3-5} 

  & & Data Space & Statistical  &  System \\
   \midrule
   Data-based &  Data Augmentation 
   & ~\cite{9005992,kang2022fedcvt} 
   & \cite{zhao2018fednoniiddata,wu2020fedhome,duan2020self,jeong2018communication} & - \\
   \midrule
  \multirow{5}{*}{\begin{tabular}{@{}c@{}}Architecture-\\ based\end{tabular}} & \multirow{2}{*}{\begin{tabular}{@{}c@{}}Parameter \\ Sharing\end{tabular}} 
  &  \multirow{2}{*}{\begin{tabular}{@{}c@{}}~\cite{YangLiu2019Secure,Sharma2019SecureAE,Ju2020FederatedTL,Gao2019HHHFLHH} \\ \cite{he2022hybrid,liang2020think} \end{tabular}} 
  & \multirow{2}{*}{\begin{tabular}{@{}c@{}}\cite{singhal2021federated,bui2019federated,liang2020think,arivazhagan2019federated} \end{tabular}} 
  & \multirow{2}{*}{\begin{tabular}{@{}c@{}}\cite{diao2021heterofl,Liu2022heterofl,hong2022efficient,munir2021fedprune} \\ \cite{li2021fedmask,li2021hermes,lu2021heterogeneous} \end{tabular}}  \\
  & &  & & \\
   \cmidrule{2-5} 
  & \multirow{3}{*}{\begin{tabular}{@{}c@{}}Knowledge \\ Distillation\end{tabular}} 
  & \multirow{3}{*}{\begin{tabular}{@{}c@{}} ~\cite{li2022semi,ren2022improving,wangvertical} \end{tabular}}
  & \multirow{3}{*}{\begin{tabular}{@{}c@{}} \cite{li2019fedmd,chang2019cronus,zhang2021parameterized,gong2022preserving} \\ \cite{li2021decentralized,yang2021mutualpfl,matsuda2022fedme,li2021fedh2l} \\ \cite{zhu2018knowledge,afonin2021towards,pmlr-v139-zhu21b} \end{tabular}}  
  & \multirow{3}{*}{\begin{tabular}{@{}c@{}}  \cite{zhang2021fedzkt,ozkara2021quped,Liu2022heterofl} \\ \cite{he2020groutknowledgetransfer,bistritz2020distributed,hin2021fedhe} \end{tabular}} \\
   & & & & \\
   & & & & \\
   \cmidrule{2-5} 
   & \multirow{2}{*}{\begin{tabular}{@{}c@{}}Domain \\ Adaptation\end{tabular}} & - 
   & \multirow{2}{*}{\begin{tabular}{@{}c@{}} ~\cite{peng2019federated,sun2016return,hanzely2020federated,deng2020adaptive} \\ \cite{Smith2017MT,huang2021pcsfl,shoham2019overcoming}  \end{tabular}} 
   & \multirow{2}{*}{\begin{tabular}{@{}c@{}}\cite{Smith2017MT,marfoq2021federated}  \end{tabular}} \\
    & & & & \\
  \midrule
  \multirow{2}{*}{Model-based} & \multirow{2}{*}{\begin{tabular}{@{}c@{}}Matrix \\ Factorization\end{tabular}}
  & \multirow{2}{*}{\begin{tabular}{@{}c@{}} ~\cite{chai2019secure,ammad2019federated,chen2020practical,chai2022federated} \\ \cite{hegedHus2019decentralized,duriakova2019pdmfrec} \end{tabular}}
  & \multirow{2}{*}{\begin{tabular}{@{}c@{}}  \cite{chai2022federated} \end{tabular}} 
  & \multirow{2}{*}{\begin{tabular}{@{}c@{}}  \cite{yao2021fedhm}  \end{tabular}}\\
       & & & & \\
 \bottomrule
\end{tabular}
};
\draw [red,thick,rounded corners]
  ($(-2.2,0)$)
  rectangle 
  ($(0.85, -0.75)$);

\draw [red,thick,rounded corners]
  ($(-2.2,1.13)$)
  rectangle 
  ($(0.85, 1.46)$);

\draw [red,thick,rounded corners]
  ($(-2.2,1.0)$)
  rectangle 
  ($(0.85, 0.3)$);

\draw [red,thick,rounded corners]
  ($(4.7,1.13)$)
  rectangle 
  ($(7.2, 1.46)$);

\end{tikzpicture}
\end{frame}

\caption{Summary of existing works using transfer learning to tackle different types of heterogeneity in FL. Red boxes denote under-explored promising directions.}
\label{fl_for_heteroFL}
\end{table*}


\section{Applications}\label{sec-applications}

This section investigates the applications of heterogeneous FL in the recommendation, advertisement, financial risk control, and healthcare. As we have discussed in Section~\ref{sec-TL_HeteroFL}, the data space heterogeneous FL is the most promising research direction and is attracting increasing research interests. This section mainly focuses on the promising application scenarios of data space heterogeneous FL. 


\subsection{Recommendation, Advertisement and Search}
The federated recommender system is one of the hottest research topics among the applications of FL, where the data distribution is heterogeneous by nature. 
Federated recommender systems can be categorized into three types regarding the data partition among the participants~\cite{Gao2020PrivacyTA}.
In vertical federated recommender systems, all participants hold heterogeneous model parameters, which are required for model inference. 
In federated transfer recommender systems, the user-item interaction matrix is generally partitioned by users or items. Participants share some parameters, while each has its own model parameters. 
When all participants share the same user-item interaction matrix, though rarely happens in reality, HFL can be conducted to learn the user-item profiles. 
We denote that some works categorize federated recommender systems into horizontal and vertical regarding the partitioning of user-item interaction matrix by users and by items, respectively~\cite{QiangYang2019,shmueli2017secure}.

Both cross-device and cross-silo settings have been explored for different scenarios recently. 
In \textit{cross-device settings}, users' mobile devices are taken as clients. In contrast, the recommender holds a server for federated model aggregation. 
Most existing researches are based on collaborative filtering~\cite{Ammaduddin2019FederatedCF,ammad2019federated,chai2019secure}. 
Generally, each mobile device represents a user and trains its local user profile on its local data. The item profiles are updated by each user and sent to the server.
After receiving the item profiles from numerous mobile devices, the recommender server conducts model aggregation to update the global item profiles. Therefore, the model aggregation is 
Chai \textit{et al.}~\cite{chai2019secure} proposed a federated matrix factorization (FedMF) method and Ammad \textit{et al.}~\cite{ammad2019federated} proposed a federated collaborative filtering (FCF) method, which are similar to each other. In FCF and FedMF, each user client computes gradients and updates the model. Chen \textit{et al.}~\cite{chen2020practical} adopted factorization machine in FL for point-of-interest recommendation (PriRec). However, it breaches privacy by revealing feature interaction weights to participants. 
Chen~\textit{et al.}~\cite{Chen2020RobustFR} proposed a robust federated recommender system to handle personalization and sparse user feedback against Byzantine attacks.

In \textit{cross-silo settings}, recommenders from diverse industries may adopt \textit{federated transfer learning} to bridge knowledge from different domains. The user-item data distribution can be vertical, where each collaborating vendor offers a different subset of items to the same underlying population of users. In such a case, the knowledge is transferred by collaboratively training the user profile. 
Otherwise, the data distribution is horizontal with different subsets of users.
Shmueli and Tassa ~\cite{shmueli2017secure} designed a secure distributed item-based CF approach based on HE. A mediator is introduced to perform intermediate secure computations yielding reduced communication costs. 
Polat and Du~\cite{Polat2008PrivacypreservingTN} focus on the case of two vendors with binary user-item data. They show how to offer recommendations in different distribution scenarios without deep privacy violations. Yakut and Polat~\cite{Yakut2012ArbitrarilyDD} studied item-based predictions on two e-commerce sites with arbitrarily distributed data. The privacy is obtained by adding fake ratings for obfuscation. 

Alternatively, one recommender can conduct \textit{VFL} together with an auxiliary data provider to use the auxiliary features (e.g., demographic information) for performance improvement. This setting is asymmetric, where the auxiliary data provider does not benefit at all from the collaboration. For example, Jeckmans~\cite{Jeckmans2012PrivacypreservingCF} adopted Prillier's HE scheme to leverage data from one company to improve the recommendation performance of the other company. 
Gao~\textit{et al.}~\cite{Gao2020PrivacyTA} investigated the plain-text vertical federated matrix factorization based on the model presented in~\cite{koren2009matrix} and found that the major privacy risk is ID leakage. 

There exist various challenges in federated recommender systems. First, data is sparse in recommender systems compared to other applications. Many existing privacy-preserving approaches adopt HE and SMC to protect data privacy by encrypting the user-item interaction matrix~\cite{chai2019secure,ammad2019federated}. 
However, since the user-item interaction matrix is sparse, the information is encoded in the entry index such as (user\_id, item\_id). 
To prevent privacy leakage, existing approaches encrypt all entries and update the full encrypted matrix for model training and aggregation, which causes impractically low efficiency.
Therefore, developing sparse data-friendly algorithms in privacy-preserving federated recommendation settings is desirable. 

VFL trains model over the aligned data. However, the in-aligned labeled data is not considered by existing VFL approaches. The vertical federated model can not evaluate the un-aligned instances.
In the inference phase, the recommender must first align the instance and decide to use the vertical federated model or the local model. 
Such a two-step process is time-consuming and fails to transfer knowledge between the vertical federated model and the local model. 
Therefore, finding a privacy-preserving FL approach is essential to enable both parties to build a high-performance model that leverages all heterogeneous features and labeled instances for model training.


\subsection{Finantial Risk Control}
Financial risks such as credit card transaction fraud cost billions yearly. Financial risk control suffers from dataset insufficiency, skewed distribution, and limitation of detection time. Moreover, due to data privacy and security, different financial institutions are not allowed to share their datasets. Researchers explored using cross-silo FL to tackle such challenges. 
Zheng~\textit{et al.}~\cite{ijcai2020_642} proposed a federated meta-learning model for fraud detection, where a shared whole novel meta-learning-based classifier is constructed by FedAvg. Yang~\textit{et al.}~\cite{10.1007/978-3-030-23551-2_2} designed a federated fraud detection model based on HFL with a data balance approach. The challenges, risks, and incentives of the federated credit system are investigated in~\cite{Papadopoulos2020}. 
Although most papers focus on horizontal settings, we denote that there are also strong motivations for companies with heterogeneous data to collaborate. For example, Tongdun Co., Ltd. explored federated load fraud detection between a consumer finance corporation and a public telecommunications operator, each with heterogeneous features. 

\subsection{Healthcare}
Medical healthcare is one of the major application directions of FL. Most existing works consider horizontally partitioned medical data among different hospitals~\cite{brisimi2018federated,xu2020federated,ju2020privacy}. For example, Brisimi~\textit{et al.} developed a decentralized support vector machine to predict hospitalizations for patients with heart diseases. 
Some works investigate FL with heterogeneous data distribution. Heterogeneous federated transfer learning (Hetero-FTL)~\cite{9005992} is proposed to transfer knowledge between participants sharing common features and is studied in in-hospital mortality prediction.
Hetero-FTL is suitable for healthcare applications where data deficiency and privacy are critical challenges. 

In healthcare, equipment differences may also cause data heterogeneity. For example, Gao~\textit{et al.}~\cite{Gao2019HHHFLHH} noticed that different electroencephalographic (EEG) devices might differ in electrode number, position, and sampling rate. Therefore, EEG data collected by different devices may not be directly used in FL. A hierarchical heterogeneous horizontal federated learning (HHHFL) approach is proposed to address the data heterogeneity challenge in EEG data. A federated transfer learning approach is designed later to improve the knowledge transfer performance in EEG data~\cite{Ju2020FederatedTL}.
Data heterogeneity broadly exist in healthcare applications due to the divergence of medical device and the multi-modality of medical records. Therefore, more efforts should be devoted to solving the challenge of heterogeneous FL. 


\section{Promising Future Research Directions}\label{sec-challenges}

This section discusses the significant challenges and promising future directions when building heterogeneous FL systems.
Based on existing heterogeneous FL literature review, we envision promising future research directions toward new heterogeneous FL framework design, realistic bench-marking, and trustworthy heterogeneous FL approaches.

\subsection{Framework Design}


\subsubsection{Model Aggregation}
In heterogeneous FL settings, averaging-based model parameter aggregation techniques could become infeasible as the model architecture varies for heterogeneous feature spaces. 
Recently, representation learning has been used to learn common feature representations from different feature spaces~\cite{YangLiu2019Secure,Sharma2019SecureAE,Gao2019HHHFLHH}. This line of study aims to convert Hetero-FTL into horizontal homogeneous FL by training a feature space-specific sub-model for each party. In this way, averaging-based model aggregation can be applied to train a global feature classifier to map the intermediate features to the label space. 
For the multi-active party settings, many studies explore to transfer knowledge in label space via knowledge distillation (KD)~\cite{hin2021fedhe,cao2022cofed}. However, these studies are proposed to train on homogeneous data in statistically or systematically heterogeneous settings and can not be directly applied to Hetero-FTL problems. 
There is a lack of studies leveraging KD to tackle the data space heterogeneity in VFL or Hetero-FTL. 
Specialized aggregation procedures for Hetero-FTL still need to be extensively explored.

\subsubsection{Adaptability}
In heterogeneous FL, either instances or features are required to be aligned among parties for knowledge transfer in collaborative training. 
However, the training and test data may not be aligned during federated model training and inference. Moreover, as the number of participants increases in VFL and instance-sharing Hetero-FTL, the number of aligned instances drops drastically. 
Some existing approaches to handling missing features can be adapted into heterogeneous FL. 
For example, in vertical decision tree models, the un-aligned missing features can be handled as missing values in centralized decision trees~\cite{fu2021vf2boost}. 
In multi-active party settings, each party train a local base learner, and the global model is the ensemble of all base learners. 
If an instance can not be aligned among all parties, the models of the un-aligned parties can be omitted during training or inference. 
However, few tailored approaches in heterogeneous FL handle the adaptability to instance or feature un-alignment. 
Therefore, it is desirable to develop FL architectures that can train and predict partially aligned data with a part of parties. 

\subsubsection{Efficiency}

Communication and computation efficiency are significant problems that affect the performance of FL. 
In heterogeneous FL, participants need to transfer intermediate results, local model parameters, or local model predictions to each other or a server. 
System redundancy, privacy mechanism, and performance requirements can affect communication and computation efficiency in terms of data volume or time delay.
Some works eliminate the communication redundancy by proposing tailored encryption and secure computation protocol~\cite{fu2021vf2boost} or asynchronous model aggregation scheme. In contrast, some other work focuses on the efficiency-performance-privacy trade-off triangle and proposes to provide weaker privacy protection and lower model performance for lower communication costs. For example, the parties may conduct more local iterations before communication-costly model aggregation. 
Although there have been several works studying asynchronous VFL algorithms~\cite{hu2019fdml,gu2020privacy}, it is still an open problem to design asynchronous algorithms for solving real-world VFL tasks with only one or partial parties holding labels.

The approach that makes full use of data sparsity is also essential to propose efficient privacy-preserving heterogeneous FL. 
We denote that existing sparse data-friendly algorithms in plain-text centralized learning can not be directly applied to privacy-preserving FL settings. For example, in heterogeneous FL, the private data may be protected by HE or secret-shared, making it infeasible to transfer the index of the non-zero value in training data~\cite{gao2020FedFM}. Therefore, sparse data-friendly heterogeneous FL approaches remain to be explored.

In VFL, a great number of features partitioned in multiple parties can decrease the computation and communication efficiency in training and inference processes. However, most features could be redundant for prediction. Privacy-preserving federated feature selection is significant in improving the efficiency of heterogeneous FL.
Federated evolutionary computation can be developed to tackle the feature selection problem in VFL problems~\cite{feng2020multiparticipant,bakopoulou2019federated,moran2017centralized}.

\subsection{Trustworthy Heterogeneous FL}

\subsubsection{Privacy}
Privacy preservation in heterogeneous FL is a key consideration due to increasingly stringent legal requirements. 
Many researches in VFL adapted HE and secure computation techniques to gradient boost decision tree models~\cite{fu2021vf2boost,Liu_2020fedforest,wu2020privacy} and linear models~\cite{zhang2021secure,xu2021fedv} and provide security against semi-honest adversaries. 
Only a few studies~\cite{9005992,YangLiu2019Secure,Sharma2019SecureAE,chen2020practical} provide stringent privacy guarantee under semi-honest or malicious threat models in Hetero-FTL. 
The deep learning-based majority in Hetero-FTL prefer to take model performance and communication efficiency as the primary concern and sacrifice data privacy by training the Hetero-FTL model in plain-text without obfuscation such as DP. 
Some researchers map heterogeneous feature space into a common latent feature space via different constrain metrics such as maximum mean discrepancy (MMD)~\cite{gretton2006kernel}, and  Earth mover’s distance (EMD)~\cite{zhao2018fednoniiddata}.
However, these metrics can only be calculated with access to raw data.
The problem of heterogeneous federated deep model training in a privacy-preserving manner remains open.

\subsubsection{Robustness}
FL provides better privacy protection compared to traditional centralized training as it eliminates raw data transmission among parties. However, it is demonstrated that the model integrity can be attacked by malicious adversaries, who deviate from the training scheme and intentionally break the model integrity. 
Compared to the studies on attacks on homogeneous FL, there are only a few studies on the robustness against malicious attacks, including poisoning attacks and adversarial attacks in VFL~\cite{liu2021defendlabelvfl,fu2022labelinferencevfl}. 
Yuan et al.~\cite{yuan2022byzantine} studied Byzantine-resilient vertical federated logistic regression via dual sub-gradient method. 
To the best of our knowledge, there is no study on robustness against attacks in Hetero-FTL. 
Therefore, it is essential to study attack methods to heterogeneous FL and develop defensive strategies to counteract these attacks to ensure the robustness of the heterogeneous FL systems. 
Especially with more complex protocols and architectures developed for Hetero-FTL, more work is needed to study related forms of attacks and defenses to enable robust Hetero-FTL approaches to emerge.

\subsubsection{Explainability}
Recently, the explainability of FL models is starting to gain traction as practical FL applications face pressure from government agencies and the general public for the underlying reason to get the prediction, especially for the high-risk application scenarios. 
However, explainable heterogeneous FL has not been explored in the literature. 
The explainability of FL models may also be associated with potential privacy risks from inadvertent information leakage, as some gradient-based explanation methods might cause privacy leakage~\cite{DBLP:journals/corr/abs-1907-00164}. 
To the best of our knowledge, there is a lack of study exploring the explainability in heterogeneous FL. 
It is an important future research direction to balance the trade-off between explainability, performance, and privacy in heterogeneous FL.






            %
\section{Conclusion}\label{sec-conclusion}


In this survey, we provide an overview of privacy-preserving heterogeneous federated learning. 
We propose a unique taxonomy of heterogeneous federated learning for data space, statistical, and system heterogeneity. 
We categorized each type of heterogeneity according to the problem setting and learning approaches and highlighted key ideas and methodologies for those heterogeneous federated learning approaches. 
We find that the data space heterogeneous federated learning, especially the Hetero-FTL is still under-explored and is a promising research direction that attracts increasing research interests. 
Then, we discuss the applications of privacy-preserving heterogeneous federated learning in different areas. 
Finally, we discuss several challenges and promising future research opportunities. 
We believe that the discussions in this survey based on our proposed heterogeneous federated learning will serve as a valuable roadmap for aspiring researchers and practitioners to enter the field of heterogeneous federated learning and contribute to its long-term development.

            %

\appendices

\section{Privacy Preservation Techniques}\label{appendix_priv_pres_tech}

We introduce four types of privacy preservation approaches, namely 1) secure multi-party computation (MPC), 2) homomorphic encryption (HE), 3) differential privacy (DP), and 4) trusted execution environment (TEE).

\subsubsection{Secure Multi-Party Computation} \label{PPML_MPC}
Secure Multi-Party Computation (MPC) is a sub-field of cryptography initially proposed and generalized by Andrew Yao~\cite{yao1982protocols63} in 1986. 
MPC aims to create methods for parties to jointly compute a function over their private inputs without revealing any information about the inputs. 
It is possible to compute any functionality without revealing any information other than the output.

MPC allows to compute functions of private input values so that each party learns only the corresponding function output value, but not input values from other parties. 
For example, given a secret value $x$ that is split into $n$ shares so that a party $P_i$ only knows $x_i$, all parties can collaboratively compute
\begin{equation*}
    y_1,...,y_n = f(x_1, ..., x_n).
\end{equation*}
Party $P_i$ learns nothing beyond the output value $y_i$ corresponding to its own input $x_i$.

In general, MPC can be implemented  through three different frameworks: 1) Secret Sharing (SS)~\cite{shamir1979share66, Rabin:1989:VSS:73007.73014}, 2) Oblivious Transfer (OT)~\cite{GMW1987, cryptoeprint:2016:505}, and 3) Threshold Homomorphic Encryption (THE)~\cite{10.1007/3-540-44987-6_18, 10.1007/978-3-540-45146-4_15}. 
From a certain point of view, both oblivious transfer protocols and threshold homomorphic encryption schemes use the idea of secret sharing. This might be why secret sharing is widely regarded as the core of MPC. In the rest of this section, we will introduce secret sharing and oblivious transfer. 

\subsubsubsection{a) Secret sharing} (SS) hides a secret value by splitting it into random parts and distributing these parts (a.k.a. shares) to different parties, so that each party has only one share and thus only one piece of the secret~\cite{Beimel2011,shamir1979share66}. 
Every secret sharing protocol consists of:
\begin{itemize}
    \item A sharing procedure $(x_1, x_2, ..., x_n) = \text{share}(x)$.
    \item A reconstruction procedure. For any $i_1, i_2, ..., i_m$, we have $x = \text{reconstruction}(x_{i_1}, x_{i_2}, ..., x_{i_m})$.
    \item A security condition. For any $x, x'$ and any $i_1, i_2, ..., i_{m-1}$, $(x_{i_1}, x_{i_2}, ..., x_{i_{m-1}})$ and $(x'_{i_1}, x'_{i_2}, ..., x'_{i_{m-1}})$ are distributed identically. 
\end{itemize}

Depending on the specific secret sharing schemes used, all or a known threshold of shares are needed to reconstruct the original secret value~\cite{Tutdere2015,shamir1979share66}.  
For example, Shamir's Secret Sharing is constructed based on polynomial equations and provides information-theoretical security, and it is also efficient using matrix calculation speedup~\cite{shamir1979share66}. 

For arithmetic secret sharing, consider that a party $P_i$ wants to share a secret $S$ among $n$ parties $\{P_i\}_{i=1}^n$ in a finite field $F_q$. To share $S$, the party $P_i$ randomly samples $n-1$ values $\{s_i\}_{i=1}^{n-1}$ from $\mathbb{Z}_q$ and set $s_n = S - \sum_{i=1}^{n-1} s_i \; mod \; q$. Then, $P_i$ distributes $s_k$ to party $P_k$, for $k \neq i$. 
We denote the shared $S$ as $\langle S \rangle = \{ s_i\}_{i=1}^n$.

To conduct secure addition, each party locally carries out the arithmetic addition operation one two secret shares. 
The secure multiplication can be performed via Beaver triples~\cite{beaver1991efficient}, which can be generated in an offline phase.


\subsubsubsection{b) Oblivious transfer} (OT) is a two-party computation protocol proposed by Rabin in 1981~\cite{OTeprint-2005-12523}. Here, we define 1-out-of-n OT: 
\begin{definition}
    1-out-of-n Oblivious Transfer: Suppose Party A has a list $(x_1, ..., x_n)$ as the input, Party B has $i \in {1, ..., n}$ as the input. 1-out-of-n Oblivious Transfer is an MPC protocol where A learns $F_A = 0$ and B learns $F_B = x_i$, that is,
    \begin{enumerate}
        \item A learns nothing about $i$.
        \item B learns nothing else but $x_i$.
    \end{enumerate}
\end{definition}
When $n=2$, we get 1-out-of-2 OT which has the following property: 1-out-of-2 OT is universal for 2-party MPC~\cite{10.1007/978-3-540-85174-5_32}. That is, one can conduct any 2-party MPC given a 1-out-of-2 OT protocol. 

There are many constructions of OT such as Bellare-Micali's~\cite{10.1007/0-387-34805-0_48}, Naor-Pinka's~\cite{Naor2001EfficientOT} and Hazay-Lindell's~\cite{Hazay2010EfficientST} approaches. 
%
%
The construction of OT requires public-key type of assumptions (e.g., discrete log assumption)~\cite{Impagliazzo:1989:LPC:73007.73012}, which is computational costly. 
For efficiency, Beaver~\cite{Bea96OTextension} proposed to \textit{extend} OT by first generating a few \textit{seed} OTs based on public-key cryptography, which can then be extended to any number of OTs using symmetric-key cryptosystems only. OT extension is now widely applied in MPC protocols \cite{mohassel2017secureml, demmler2015aby, cryptoeprint:2016:505} to improve efficiency.

Yao's Garbled Circuit (GC)~\cite{yao1982protocols63} is an OT-based secure 2-party computation protocol that can evaluate any function.
The key idea of Yao's GC is to decompose the computational circuits (we can use circuits (AND, OR, XOR) to compute any arithmetic operation) into generation and evaluation stages. 

\subsubsection{Homomorphic Encryption}\label{PPML_HE}
Homomorphic encryption (HE) is a cryptographic scheme that enables homomorphic operations on encrypted data without decryption. 
The concept of HE was initially introduced by Rivest et al.~\cite{rivest1978data32} in 1978 to perform computation over ciphertext.
Since then, researchers has made may attempts to design such homomorphic schemes. 

An HE scheme $\mathcal{H}$ is an encryption scheme that allows certain algebraic operations to be carried out on the encrypted content, by applying an efficient operation to the corresponding ciphertext without knowing the decryption key. 

\begin{definition} \textbf{(Homomorphic Encryption)}
    An HE scheme $\mathcal{H}$ consists of the following poly-time algorithms:
    \begin{equation}
        \mathcal{H} = \{KeyGen, Enc, Dec, Eval\}
    \end{equation}
    where 
    \begin{itemize}
        \item[-] $KeyGen$: Key generation. 
        A cryptographic generator $g$ is taken as the input. 
        For asymmetric HE, a pair of keys $\{pk, sk\} = KeyGen(g)$ are generated, where $pk$ is the public key for encryption and $sk$ is the secret key for decryption. 
        For symmetric HE, only a secret key $sk = KeyGen(g)$ is generated. 
        \item[-] $Enc$: Encryption. 
        For asymmetric HE, an encryption scheme takes the public key $pk$ and the plaintext $m$ as the input, and generates the ciphertext $c = Enc_{pk}(m)$ as the output. 
        For symmetric HE, an HE scheme takes the secret key $sk$ and the plaintext $m$, and generates ciphertext $c = Enc_{sk}(m)$. 
        \item[-] $Dec$: Decryption. 
        Given the secret key $sk$ and the ciphertext $c$, the corresponding plaintext $m = Dec_{sk}(c)$. 
        \item[-] $Eval$: Evaluation. 
        The function $Eval$ takes the ciphertexts $c, c'$ as the input, and outputs a ciphertext corresponding to a functioned plaintext. 
    \end{itemize}
\end{definition}

Let $Enc_{pk}(\cdot)$ denote the encryption function with $pk$ as the encryption key. Let $\mathcal{M}$ denote the plaintext space and $\mathcal{C}$ denote the ciphertext space. A secure cryptosystem is called \emph{homomorphic} if it satisfies the following condition: $\forall m_1, m_2 \in \mathcal{M}, \;\; Enc_{pk}(m_1 \odot_{\mathcal{M}} m_2) \leftarrow Enc_{pk}(m_1)\odot_{\mathcal{C}} Enc_{pk}(m_2)$
for some operators $\odot_{\mathcal{M}}$ in $\mathcal{M}$ and $\odot_{\mathcal{C}}$ in $\mathcal{C}$, where $\leftarrow$ indicates the left-hand side term is equal to or can be directly computed from the right-hand side term without any intermediate decryption.

There are three types of HE schemes: 
\begin{itemize}
\item Partially Homomorphic Encryption (PHE) supports the evaluation of functionality with one type algebraic operation (e.g. addition). 
\item Somewhat Homomorphic Encryption (SHE) schemes can evaluate two types of algebraic operations, but only for a subset of functionality. 
\item Fully Homomorphic Encryption (FHE) supports the evaluation of arbitrary functionality composed of multiple types of algebraic operations. 
\end{itemize}
In general, for HE schemes, the computational complexity increases as the functionality grows. 
Interested readers can refer to~\cite{armknecht2015guide} and~\cite{Acar:2018:SHE:3236632.3214303} for more details regarding different classes of HE schemes.

The Goldwasser–Micali cryptosystem~\cite{goldwasser1982probabilistic} is a PHE scheme that enables addition modulo 2 operation over ciphertext. 
The Paillier cryptosystem~\cite{Paillier99public-keycryptosystems} is a PHE scheme that allows additive operations over ciphertext. It has been widely applied to various PPML researches.  
Gentry~\cite{gentry2009fully} proposed the first FHE scheme that supports both additive and multiplicative operations for unlimited number of times. 
Recently, Cheon, Kim, Kim and Song (CKKS)~\cite{cheon2017homomorphic} proposed an approximate FHE scheme that supports a special kind of fixed-point arithmetic. 
The CKKS scheme is the most efficient approach for evaluating polynomial approximations, and is the preferred approach for implementing PPML applications.

\subsubsection{Differential Privacy}\label{PPML_DP}
Differential Privacy (DP)~\cite{dwork2006calibrating} was initially developed to facilitate secure analysis over privacy-sensitive data. 
The key idea of DP is to protect membership-privacy by confusing the adversary when it tries to query individual information so that adversaries cannot distinguish individual-level sensitivity from the query results. 

DP provides an information-theoretic security guarantee that the output of a function is insensitive to any particular record in the dataset. 
Therefore, DP can be used to resist the membership inference attack. 
The $(\epsilon, \delta)$-differential privacy is defined as follows: 
\begin{definition}\textbf{($\epsilon, \delta$)-differential privacy.}
A randomized mechanism $\mathcal{M}$ preserves ($\epsilon, \delta$)-differential privacy if given any two datasets $D$ and $D'$ differing by only one record, and for all $S \subset Range(\mathcal{M})$,
\begin{equation*}
    Pr[\mathcal{M}(d)\in S] \leq \text{exp}(\epsilon)Pr[\mathcal{M}(D') \in S]  + \delta
\end{equation*}
where $\epsilon$ is the privacy budget and $\delta$ is the failure probability.
\end{definition}

When $\delta = 0$, we achieve a stronger notion named $\epsilon$-differential privacy.
The \emph{privacy loss} is defined as $ln \frac{Pr[\mathcal{M}(D)\in S]}{Pr[\mathcal{M}(D') \in S]}$ , where $ln$ denotes natural logarithm.

The approaches to design DP algorithms can be categorized into two types. 
One is the addition of noise according to the sensitivity of a function~\cite{dwork2006calibrating}. 
The other is choosing noise according to an exponential distribution on discrete values~\cite{mcsherry2007mechanism}. 

The sensitivity of a real-valued function expresses the maximum possible change in its value due to the addition or removal of a single sample:
\begin{definition} \textbf{(Sensitivity)}
    For two datasets $D$ and $D'$ differing by only one record, and a function $\mathcal{M}: \mathcal{D} \rightarrow \mathcal{R}^d$ over an arbitrary domain, the sensitivity of $\mathcal{M}$ is the maximum change in the output of $\mathcal{M}$ over all possible inputs:
    \begin{equation}
        \Delta \mathcal{M} = \max\limits_{D, D'} \; \lVert {\mathcal{M}(D) - \mathcal{M}(D')} \rVert
    \end{equation}
    where $\lVert \cdot \rVert$ is a norm of the vector. The $l_1$-sensitivity or the $l_2$-sensitivity is defined when the $l_1$-norm or $l_2$-norm is applied, respectively.
\end{definition}

We denote the Laplace distribution with parameter $b$ as $Lap(b)$. $Lap(b)$ has a probability density function $P(z|b) = \frac{1}{2b}exp(-|z|/b)$. 
Given a function $\mathcal{M}$ with sensitivity $\Delta \mathcal{M}$, the addition of noise drawn from a calibrated Laplace distribution $Lap(\Delta \mathcal{M}/\epsilon)$ maintains $\epsilon$-differential privacy~\cite{dwork2006calibrating}:
\begin{theorem}\textbf{(Laplace mechanism).}
    Given a function $\mathcal{M}: \mathcal{D} \rightarrow \mathcal{R}^d$ over an arbitrary domain $D$, for any input $X$, the function:
    \begin{equation}
        \mathcal{M}(X) + Lap(\frac{\Delta \mathcal{M}}{\epsilon})^d
    \end{equation}
    provides $\epsilon$-differential privacy.
\end{theorem}
The $\epsilon$-differential privacy can also be achieved by adding independently generated Laplace noise from distribution $Lap(\Delta \mathcal{M}/\epsilon)$ to each of the $d$ output terms.
Moreover, changing Laplace noise to Gaussian or binomial noise, and scale to the $l_2$-sensitivity of the function, sometimes yields better accuracy but only ensures weaker DP~\cite{dwork2006our, dwork2004privacy}.

The \emph{exponential mechanism}~\cite{mcsherry2007mechanism} can achieve DP on discrete values. 
The exponential mechanism is given a quality function $q$ that scores outcomes of a calculation, where higher scores are better. 
For a given database and $\epsilon$ parameter, the quality function induces a probability distribution over the output domain, from which the exponential mechanism samples the outcome. This probability distribution favors high-scoring outcomes, while ensuring $\epsilon$-differential privacy.
\begin{theorem}\textbf{(Exponential mechanism).}
    Let $q:(\mathcal{D}^n \times \mathcal{R}) \rightarrow \mathbb{R}$ be a quality function, which given a dataset $d\in \mathcal{D}^n$, assigns a score to each outcome $r\in \mathcal{R}$. For any two datasets $D$ and $D'$ differing by only one record, let $S(q) = \max\limits_{r,D,D'}\lVert q(D,r) - q(D', r) \rVert_1$. Let $\mathcal{M}$ be a mechanism for choosing an outcome $r\in \mathcal{R}$ given a dataset instance $d\in D^n$. Then, the mechanism $\mathcal{M}$, defined as
    \begin{equation}
        \mathcal{M}(d,q) = \left\{ \text{return $r$ with prob.} \propto \exp\left(\frac{\epsilon q(d,r)}{2S(q)}\right) \right\}
    \end{equation}
    provides $\epsilon$-differential privacy.
\end{theorem}

According to how and where the noise perturbation is applied, the DP algorithms can be categorized into four types as follows:
\begin{enumerate}
    \item \emph{Input perturbation}: The noise is added to the training data.
    \item \emph{Objective perturbation}: The noise is added to the objective function of the learning algorithms.
    \item \emph{Algorithm perturbation}: The noise is added to the intermediate values such as gradients in iterative algorithms. 
    \item \emph{Output perturbation}: The noise is added to the output parameters after training.
\end{enumerate}

Differential privacy only aims to protect membership-privacy that one adversary can not determine if a certain instance is inside the dataset. However, the statistics of the dataset can still be revealed, which could be sensitive in some cases, such as financial data, medical data and other commercial and health applications. 
Moreover, DP faces a trade-off between utility and privacy as it introduces noises to results. 
It is shown that current DP mechanisms in ML rarely provide practical privacy-utility trade-offs: settings that provide little accuracy loss can hardly provide strong privacy protection, while the settings that provide strong privacy result in large accuracy loss. 
Readers who are interested in differential privacy can refer to the tutorial given by~\cite{dwork2014algorithmic}.

\subsubsection{Trusted Execution Environment}
As an attempt to guarantee data privacy while compromising less efficiency and utility, the Trusted Execution Environment (TEE), a hardware-based private data analysis technique, attracts increasing attention in the PPML community~\cite{narra2019privacypreserving,CHEN202069teefl,chen2020training}. 
\begin{definition}
A TEE is a tamper-resistant processing environment that guarantees the authenticity of the code executed, the integrity of the runtime states, and the confidentiality of its code, data, and runtime states stored on persistent memory. 
It runs on a separation kernel which guarantees spatial and temporal separation, control of information flow and fault isolation.
\end{definition}
Representative implementations include Intel’s SGX technology, AMD’s SEV technology, and ARM’s TrustZone technology.
Fan and Hamed~\cite{CHEN202069teefl} explored the use of TrustZone in edge devices to conduct efficient and private FL. The local model training process is conducted in the TEE of client devices to prevent clients from inferring information from gradients present in the global model or conducting causative attacks. The server encrypts the model after model aggregation and sends it to the clients. The model is decrypted inside the TEE for training, in which the client can not learn information about the model. After local training, the trusted application encrypts the model before sending it outside the TrustZone and back to the server. Therefore, all clients collaboratively train the model without knowing its plaintext. 
Chen et al.~\cite{chen2020training} studied the privacy-preserving FL with TEE via SGX. Local model training and severe model aggregation are conducted in secure enclaves for privacy preservation and training process integrity. 
Apart from protecting the model against clients by conducting FL in TEEs of edge devices and encrypting models, clients can also encrypt and send private data to TEE in a server. The server then decrypts data inside TEE and trains a global model over data contributed by all the clients.

\section{Knowledge Distillation}~\label{appendix_knowledge_distillation}
Three forms of knowledge can be leveraged to transfer knowledge from the teacher model to the student model: response-based knowledge, feature-based knowledge, and relation-based knowledge. 
We introduce different categories of knowledge as follows.

\subsubsubsection{a) Response-based knowledge} refers to the logits $z$, which are the outputs of a deep model's last fully connected layer. The response-based knowledge distillation directly mimics the final prediction of the teacher model and has been widely used in Hetero-FTL. 
In response-based knowledge distillation, soft targets~\cite{kdHinton2015} are introduced to improve knowledge transfer as they contain the informative dark knowledge from the teacher model. 
Soft targets are the probabilities that the input belongs to the class and can be estimated by a softmax function as
\begin{equation*}
p(z_i, T) = \frac{exp(z_i/T)}{\sum_{j}exp(z_j)/T},
\end{equation*}
where $z_i$ is the logit for the $i$-th class, and $T$ is a temperature factor that controls the importance of each soft target. 
The distillation loss is the divergence of the soft targets of teacher and student, which can be formulated as
\begin{equation*}
\mathcal{L}_{Res} = T^2 \mathcal{L}_{div}(p(z_t, T), p(z_s, T)),
\end{equation*}
where $\mathcal{L}_{div}$ denotes that divergence loss of soft targets. Generally, $\mathcal{L}_{div}$ often employs Kullback-Leibler(KL) divergence loss. 
In vanilla knowledge distillation~\cite{kdHinton2015}, the student model is trained by the distillation loss and the cross entropy loss $\mathcal{L}_{CE}(y, p(z_s, T=1))$ between the ground truth labels and the soft logits of the student model. 

\subsubsubsection{b) Feature-based knowledge distillation} employs the output of intermediate layers to replace the logits used in response-based knowledge. The feature activations of the teacher and the student are directly matched to distillation loss. 
Therefore, the teacher and the student models first project the features to a common latent space. 
Then, a distillation loss is employed to minimize the distance of the two feature representations from the teacher and the student of the same instance in the latent space. Zhu et al.~\cite{pmlr-v139-zhu21b} propose to use a generative network to learn feature representations to transfer knowledge between parties in FL. 

\subsubsubsection{c) Relation-based knowledge} explores the distance between different instances or layers in the same model as knowledge to train the student model. 
The distillation loss is based on relations of feature representations~\cite{yim2017gift} or instance\cite{you2017learning,passalis2020probabilistic,park2019relational}. 
Yim et al.~\cite{yim2017gift} transfer knowledge to the student model via the Gram matrix between two layers. 
The mutual relations of instances are also leveraged as knowledge~\cite{you2017learning,park2019relational}. 
Park et al.~\cite{park2019relational} transfer knowledge from instance relations by proposing a relational knowledge distillation scheme. 
Passalis et al.~\cite{passalis2020probabilistic} use feature representations to model the instance-wise relations as probabilistic distribution.

%
\IEEEpeerreviewmaketitle

\ifCLASSOPTIONcaptionsoff
  \newpage
\fi



\bibliographystyle{IEEEtran}
\bibliography{ref}

\end{document}